\documentclass{article}
\usepackage{iclr2027_conference,times}
\usepackage{amsmath,amsfonts,bm}
\usepackage{xspace}

\newcommand{\peerreveal}{\textsc{PeerReveal}\xspace}
\newcommand{\peerrevealbase}{\textsc{PeerReveal-Base}\xspace}
\newcommand{\semdetect}{\textsc{Sem-Detect}\xspace}
\newcommand{\anchor}{\textsc{Anchor}\xspace}
\newcommand{\radar}{\textsc{Radar}\xspace}
\newcommand{\tf}{\textsc{Tf}\xspace}
\newcommand{\logrank}{\textsc{LogRank}\xspace}
\newcommand{\fastdetectgpt}{\textsc{Fast-DetectGPT}\xspace}
\newcommand{\binoculars}{\textsc{Binoculars}\xspace}
\newcommand{\mage}{\textsc{Mage}\xspace}
\newcommand{\editlens}{\textsc{EditLens}\xspace}
\newcommand{\roberta}{\textsc{RoBERTa}\xspace}
\xspaceaddexceptions{\&}

\def\eqref#1{equation~\ref{#1}}

\def\1{\bm{1}}

\DeclareMathAlphabet{\mathsfit}{\encodingdefault}{\sfdefault}{m}{sl}
\SetMathAlphabet{\mathsfit}{bold}{\encodingdefault}{\sfdefault}{bx}{n}

\usepackage{algorithm}
\usepackage{algpseudocode}
\usepackage{hyperref}
\hypersetup{hidelinks}
\usepackage{url}
\usepackage{booktabs}
\usepackage{capt-of}
\usepackage{placeins}
\usepackage{etoolbox}
\AtBeginEnvironment{table}{\setlength{\belowcaptionskip}{10pt}}
\usepackage{array}
\usepackage{amsmath}
\usepackage{amssymb}
\usepackage{graphicx}
\usepackage{multirow}
\usepackage{xcolor}
\newcommand{\rev}[1]{#1}
\newenvironment{revblock}{}{}
\usepackage{tikz}
\usetikzlibrary{calc,arrows.meta,positioning}
\usepackage{pgfplots}
\pgfplotsset{compat=1.18}

\title{PeerReveal: Learning to Generalize in\\AI Peer Review Detection}

\author{Zi Li$^{1,2}$ \quad Andr\'{e} V. Duarte$^{1,3}$ \quad Lei Li$^{1}$ \\
\normalfont $^{1}$Carnegie Mellon University, $^{2}$Nanjing University, $^{3}$Instituto Superior T\'{e}cnico/INESC-ID \\
\normalfont \texttt{zili@smail.nju.edu.cn, aduarte@andrew.cmu.edu, leili@cs.cmu.edu}
}

\iclrfinalcopy %
\renewcommand{\headrulewidth}{0pt} %
\begin{document}

\maketitle
\lhead{} %
\suppressfloats[t] %

\begin{abstract}
Can we distinguish fully AI-generated reviews from those using AI only to rewrite human judgement? Furthermore, can we do so reliably when reviews are generated from diverse AI models and workflows across scientific domains? To this end, we introduce \peerreveal, a new detector that grounds its decisions in learned relationships between a review's claims and reference claims drawn from AI-generated reviews of the same paper, complemented by review-level textual signals. We evaluate \peerreveal on PeerShift, a new, challenging test set of 5,430 reviews spanning multiple domains, including machine learning, life sciences, and social sciences, combining variation in disciplinary review conventions with diverse AI review-production mechanisms. With regular training, \peerreveal achieves 70.52\% pooled Macro-F1 on this benchmark, improving over the strongest baseline by 11.48 points. We develop an effective and practical adversarial training framework that further improves the detector's generalization using a small local generator optimized to produce challenging reviews, raising pooled Macro-F1 from 70.52\% to 82.40\%. These results show that robust AI-review detection benefits from richer review-level evidence and adversarial training on more challenging generation processes.
\end{abstract}

\section{Introduction}
\label{sec:introduction}

Peer review is central to scientific progress. As large language models become increasingly capable of reading papers and producing detailed evaluations, LLM assistance in scientific reviewing now ranges from polishing language grounded in human judgement to generating an assessment end to end~\citep{pmlr-v235-liang24b,Zhou_2026}. This range of practices makes review provenance increasingly difficult to infer: similar-looking reviews may reflect human judgements polished or expanded by an LLM, or critiques and judgements generated by the model itself.

\begin{figure}[t]
\centering
\includegraphics[width=.92\linewidth]{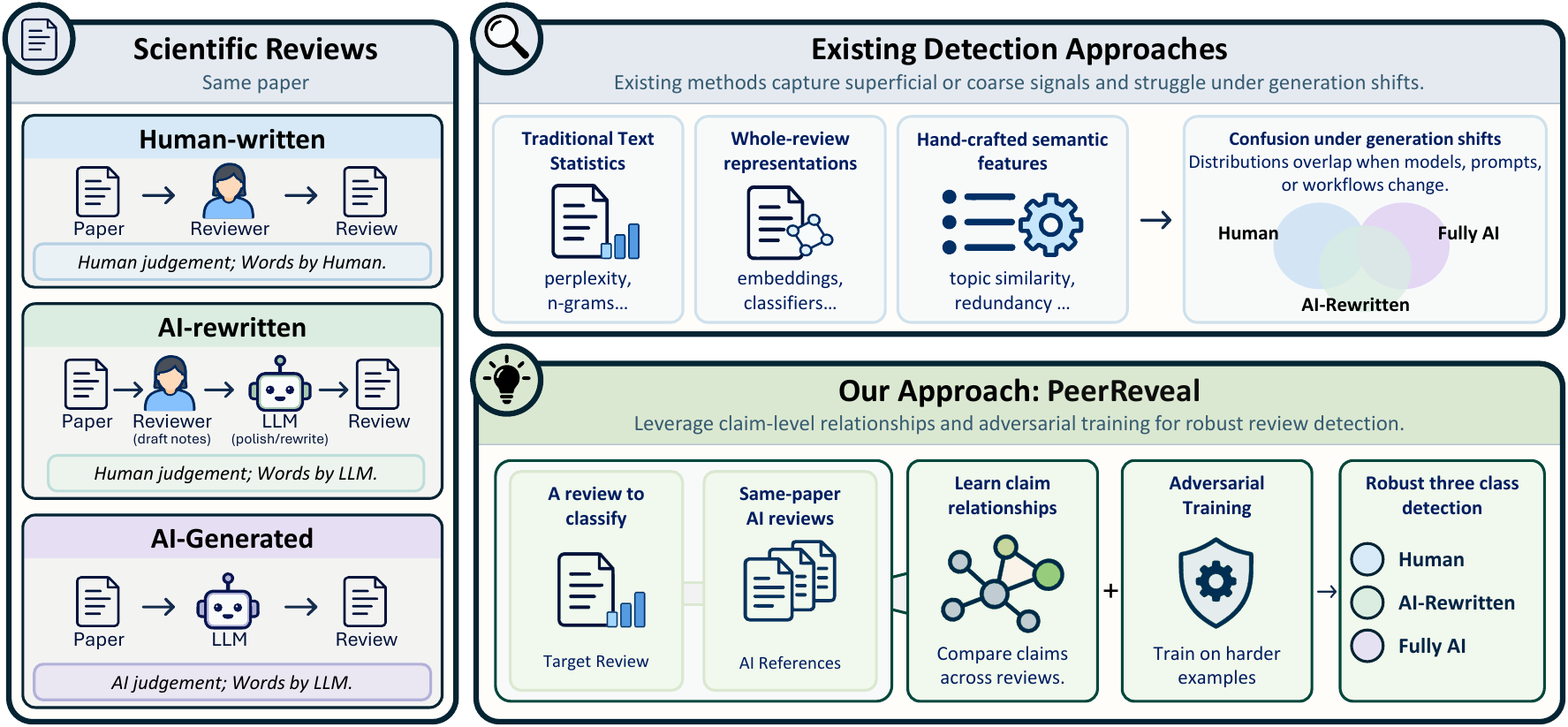}
\caption{\peerreveal learns claim relationships and uses adversarial training to distinguish human-written, AI-rewritten, and AI-generated reviews across models, workflows, and domains.}
\label{fig:overview}
\end{figure}

Existing scientific-review detectors already exploit manuscript context and review-specific signals~\citep{kumar-etal-2024-quis}, including comparisons with AI-generated reviews of the same paper as references, but two challenges remain. First, reference-based approaches usually compare reviews at the whole-review level or summarize claim-level comparisons through predefined features, rather than learning directly from interactions between individual target and reference claims~\citep{ICLR2026_64592277,duarte2026semdetect}. Second, existing evaluations are typically built around a small, fixed set of models and prompting procedures selected in advance, even though real-world review production may involve models and workflows that cannot be anticipated when a benchmark is constructed.

We address the first challenge with \peerreveal (Figure~\ref{fig:overview}), a blockwise-masked Transformer \citep{DBLP:conf/nips/VaswaniSPUJGKP17} that, through directed attention, learns directly from interactions between individual claims in the target review and in same-paper AI reference reviews. Using the ML-only regular training corpus of \semdetect~\citep{duarte2026semdetect}, \peerreveal improves Macro-F1 by 11.83 points on the in-distribution evaluation, 21.55 points with unseen writer models, and 18.84 points with unseen writer models and prompts, demonstrating its architectural benefit.

Although these results are strong under controlled writer-model and prompt shifts, they do not establish whether the detector remains reliable when review production departs from the conditions anticipated during benchmark construction. For that reason, we construct PeerShift, a broader, deliberately challenging test set, using an automatic search method that discovers diverse ways of generating AI reviews and rewriting human reviews while maintaining review quality. The resulting evaluation spans different writer models and comprises a balanced set of 5,430 reviews across machine learning, life sciences, and social sciences. On its ML subset (PeerShift-ML), \peerreveal trained only on the original ML data achieves 48.14\% Macro-F1, outperforming \semdetect by 13.74 points, but leaving a substantial generalization gap.

This remaining gap suggests that detector architecture alone is insufficient when training is confined to a fixed distribution. We therefore develop an adversarial generator--detector training framework in which a small local generator is optimized to produce difficult, high-quality reviews that expose weaknesses in the detector. On PeerShift, adversarial training improves pooled Macro-F1 from 70.52\% with regular training to 82.40\%, demonstrating an effective and practical approach to improving detector generalization with a small local generator.

Our contributions are:
\begin{enumerate}
    \item \textbf{A stronger peer-review authorship detector.} \peerreveal directly learns interactions between individual claims in target and same-paper AI reference reviews, substantially improving detection under both in-distribution and controlled distribution-shift settings.

    \item \textbf{Systematic construction of challenging peer-review evaluations.} We automatically search for diverse review-production mechanisms under quality constraints, yielding PeerShift, a cross-domain test set probing generalization beyond model and prompt shifts.

    \item \textbf{A practical adversarial training framework for improving generalization.} Our framework adversarially optimizes a small local generator to produce difficult training examples, providing an effective source of additional supervision and substantially improving detection under review-production conditions \rev{unseen in training}.
\end{enumerate}

\section{Related Work}
\label{sec:related-work}

Detecting AI-generated text has prompted research on watermarking, statistical detection, and supervised classification~\citep{wu-etal-2025-survey}. We first review general-purpose detectors and methods for scientific peer review, and then discuss generalization and adversarial training.

\paragraph{General-purpose AI-text detection.}
Watermarking embeds detectable signals during text generation, but requires control over the generation process~\citep{pmlr-v202-kirchenbauer23a}. Zero-shot detectors instead identify generated text from its statistical properties without training on labeled detection examples. DetectGPT~\citep{pmlr-v202-mitchell23a} and \fastdetectgpt~\citep{ICLR2024_6b8c6f84} use probability-curvature signals, while \binoculars compares predictions from two language models~\citep{pmlr-v235-hans24a}. Supervised detectors learn from labeled human and AI-generated text, as in the fine-tuned Longformer~\citep{beltagy2020longformerlongdocumenttransformer} detector released with \mage~\citep{li-etal-2024-mage}.

\paragraph{Scientific-review detection.}
Scientific-review detection can exploit both review-specific linguistic patterns and the relationship between a review and the manuscript it evaluates. \citet{pmlr-v235-liang24b} estimate AI use at the corpus level from word-frequency changes, while \citet{kumar-etal-2024-quis} use token-frequency patterns to classify individual reviews. \anchor generates an AI review of the same paper as a reference and compares it with the target review using whole-review embedding similarity~\citep{ICLR2026_64592277}. A complementary approach places hidden instructions in manuscripts to induce detectable patterns in AI-generated reviews~\citep{10.1371/journal.pone.0331871}.

AI-assisted editing further complicates authorship attribution, because human judgements may be expressed in AI-revised language. \editlens estimates the extent of AI editing in a text~\citep{ICLR2026_c7d936e6}. \rev{Recent benchmarks separate human-written, AI-polished, and AI-generated peer reviews~\citep{saha2026policies,sadeghian2026peerprism}.} \semdetect combines claim-level semantic statistics with textual features to distinguish human-written, AI-rewritten, and fully AI-generated reviews~\citep{duarte2026semdetect}. Building on this claim-level evidence, \peerreveal retains individual claim representations and learns their interactions instead of reducing their comparisons to predefined aggregate statistics.

\paragraph{Generalization and adversarial training.}
\mage evaluates detection across text domains and generator models~\citep{li-etal-2024-mage}, while RAID broadens evaluation to diverse models, domains, decoding settings, and adversarial modifications~\citep{dugan-etal-2024-raid}. \semdetect also evaluates transfer across writer models, prompting strategies, and review domains. We extend this evaluation setting by automatically searching for diverse ways to generate AI reviews and rewrite human reviews, constructing a broader test set across models and scientific domains.

DIPPER shows that paraphrasing can evade AI-text detectors while largely preserving meaning~\citep{NEURIPS2023_575c4500}. For robustness to paraphrasing, \radar trains a paraphraser to evade detection and a detector to recognize its outputs~\citep{NEURIPS2023_30e15e59}. Our framework uses detector feedback to optimize a small local review generator and improves three-class detection through training on difficult AI-generated and AI-rewritten reviews under review-quality constraints.

\section{PeerShift: A Challenging Peer-Review Test Set}
\label{sec:evaluation-construction}

A detector may perform well under fixed generation procedures yet remain unreliable under other plausible ways of generating AI reviews or rewriting human ones. In practice, the model and workflow used to produce a review are unknown.

We therefore construct PeerShift, a three-class peer-review authorship test set, through automatic search over review-production mechanisms that guide assessment formation or human-review reconstruction. The search selects complementary mechanisms producing challenging, high-quality reviews and fixes them, within each scientific domain, for generation across writer models.

\subsection{Review-Production Mechanisms}

A review-production mechanism specifies how a writer model forms a review: the writer model is the language model that produces the text, whereas the mechanism determines which evidence it considers and how it reasons toward a judgement. The search below varies the mechanism while holding the base review prompt fixed. We represent a mechanism as a program: an ordered sequence of compatible operations, each a reusable reviewing instruction from a shared library.

For fully AI-generated reviews, operations cover evidence selection, critique and judgement formation, and grounding repair. For example, a program combines operations to select a result, assess alternative explanations against evidence, and form a judgement reflecting uncertainty.

For AI-rewritten reviews, operations vary organization, emphasis, judgement placement, and argumentative presentation while preserving the source's reasons, judgement, and stance. Programs are compiled by appending their operation instructions to a fixed review prompt. For AI generation, the writer receives the manuscript; for rewriting, it receives source-order notes extracted from the human review rather than the review itself, and reconstructs a complete review from them.

\subsection{Quality-Constrained Mechanism Search}

We search for complementary mechanisms that increase detection difficulty and diversity while satisfying quality and validity constraints:
\begin{equation}
\mathcal{S}^{*}=\arg\max_{\mathcal{S}\subseteq\mathcal{Z},\,|\mathcal{S}|=K}\;
\sum_{z\in\mathcal{S}}\min_{e\in\mathcal{E}}\bigl[r_e(\varnothing)-r_e(z)\bigr]
\quad\text{s.t. }Q_z\geq\tau_Q,\; V_z\geq\tau_V\;\;\forall z\in\mathcal{S}.
\label{eq:mechanism-search-objective}
\end{equation}
Here $\mathcal{Z}$ is the program space over the operation library, $r_e(z)$ the recall of development detector $e\in\mathcal{E}$ on the target class (AI-generated or rewritten) of pilot reviews from program $z$, and $\varnothing$ the base prompt on the same papers, writer models, and ratings; the inner minimum $D_z$ is the difficulty of $z$. $U_z$ measures marginal diversity in programs and outputs relative to archive $\mathcal{A}$ and breaks remaining ties (Appendix~\ref{app:acceptance-constraints}). $Q_z$ measures review utility, while $V_z$ measures paper grounding for AI generation or source fidelity for rewriting; $\tau_Q$ and $\tau_V$ set minimum quality and validity.

Algorithm~\ref{alg:review-mechanism-search} summarizes the AI-generation search. The frontier $\mathcal{F}$ contains candidates for expansion, and $\mathcal{A}$ retains those passing the quality and validity checks. \textsc{Add} introduces an operation, \textsc{Replace} substitutes one, \textsc{Combine} merges compatible programs, and \textsc{Repair} corrects diagnosed validity failures. Candidates are evaluated through pilot reviews generated on a balanced development schedule $\mathcal{D}$ of papers, writer models, and ratings.

\begin{algorithm}[H]
\caption{Automatic quality-constrained mechanism search}
\label{alg:review-mechanism-search}
\small
\begin{algorithmic}[1]
\Require prompt $p_0$; operations $\mathcal{O}$; development schedule $\mathcal{D}$; detector panel $\mathcal{E}$; budget $B$; output size $K$
\State $\mathcal{F}\gets\{\varnothing\}$; $\mathcal{A}\gets\varnothing$ \Comment{base prompt, no additional operations}
\For{$b=1,\ldots,B$}
    \State $\mathcal{P}\gets\Call{Expand}{\mathcal{F},\mathcal{A},\mathcal{O};\ \textsc{Add},\textsc{Replace},\textsc{Combine},\textsc{Repair}}$
    \ForAll{complete programs $z\in\mathcal{P}$}
        \State $p_z\gets\Call{Compile}{p_0,z}$; $R_z\gets\Call{BalancedGenerate}{p_z,\mathcal{D}}$
        \State $(D_z,U_z,Q_z,V_z)\gets\Call{Evaluate}{R_z,\mathcal{E},\mathcal{A}}$; retain $z$ in $\mathcal{A}$ if $Q_z\geq\tau_Q$ and $V_z\geq\tau_V$
    \EndFor
    \State $\mathcal{F}\gets\Call{SelectFrontier}{\mathcal{A};\{D_z,U_z\}_{z\in\mathcal{A}}}$
\EndFor
\State $\mathcal{A}\gets\Call{Confirm}{\mathcal{A},\mathcal{D}}$ \Comment{larger balanced pilots}
\State $\mathcal{S}^{*}\gets\Call{SelectComplementary}{\mathcal{A},K}$;\ \Return $\Call{Freeze}{\mathcal{S}^{*}}$
\end{algorithmic}
\end{algorithm}

\subsection{Cross-Domain Test-Set Construction}

The search runs separately for each domain and selects four AI-generation and six rewrite mechanisms for ML and for Nature Communications, and three and five for PeerJ. We freeze them and generate reviews from independent manuscript--human-review corpora, requiring reasoned judgements, rating agreement, and the validity checks above. PeerShift contains 5,430 reviews: 1,830 in its ML subset (PeerShift-ML) and 1,800 each in Nature Communications and PeerJ Psychology and Social Sciences, with balanced classes in each domain. Appendix~\ref{app:mechanism-details} details the search controller and detectors, operations, frozen mechanisms and prompts, and dataset composition.

\section{Three-Class Detector Design and Adversarial Training}
\label{sec:detection-framework}

After defining the evaluation shifts, we turn to the trainable components of our approach. We first introduce \peerreveal's three-class detector architecture and then describe the adversarial training framework. Regular training produces \peerrevealbase; continuing training with regular and adversarial reviews produces \peerreveal.

\subsection{\peerreveal Three-Class Detector}
\label{sec:unified-detector}

\peerreveal jointly models claim-level semantic evidence and review-level textual signals. It retains the variable-length target and reference claim sequences rather than reducing their relationships to fixed aggregate statistics. Directed self-attention learns from interactions among individual claims and their comparisons to distinguish human-written, AI-rewritten, and fully AI-generated reviews.

\begin{figure}[htbp]
\centering
\includegraphics[width=0.9\linewidth]{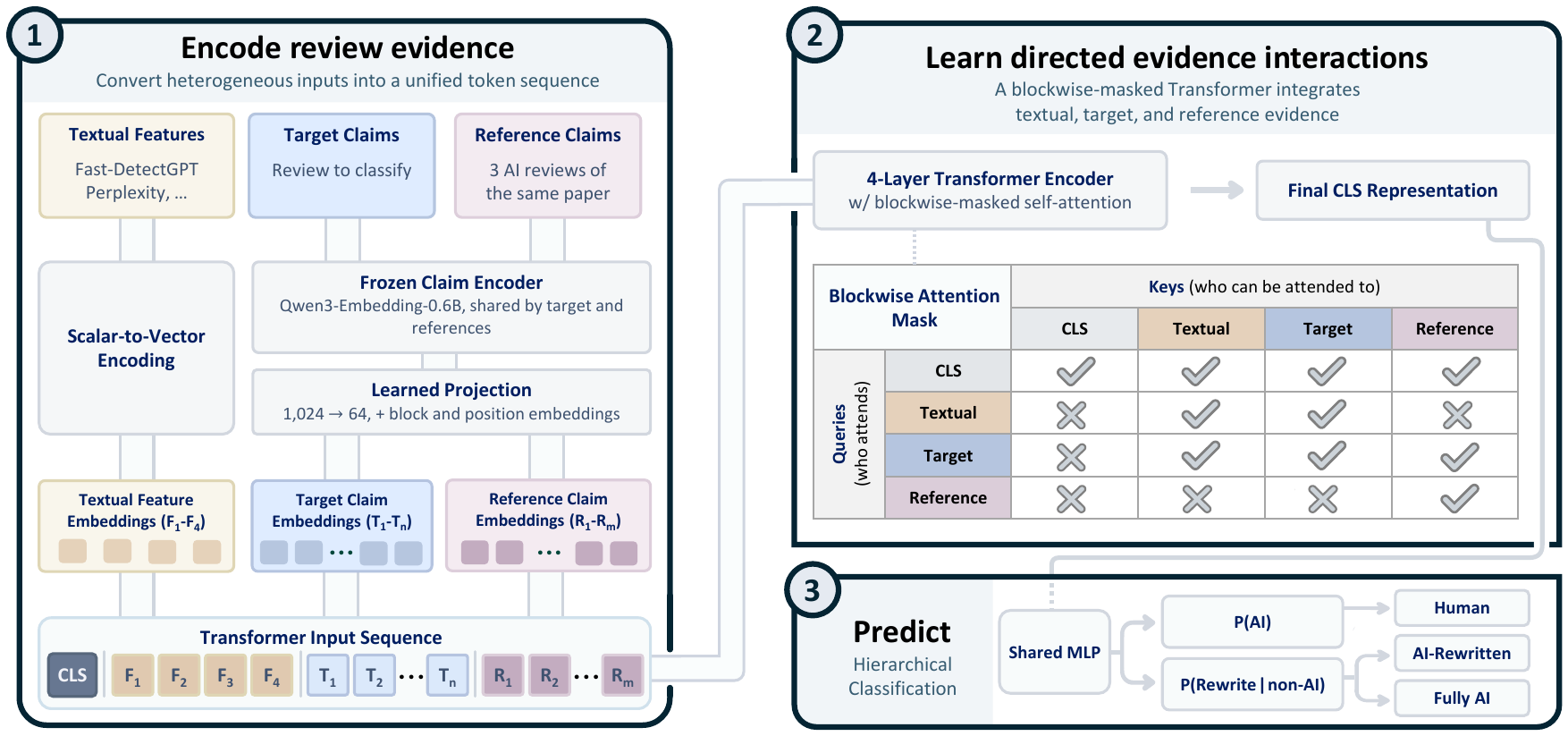}
\caption{\peerreveal detector. A frozen claim encoder and shared projection encode target and reference claims; four textual statistics are encoded as feature vectors. Directed blockwise attention integrates these inputs, and the final CLS representation predicts the three review sources.}
\label{fig:detector}
\end{figure}

\paragraph{Input construction.}

For each target review, we use three pre-generated AI reviews of the same manuscript and rating as references, following \semdetect~\citep{duarte2026semdetect}. We decompose each review into functionally labeled claims and retain evaluation, constructive-input, and clarification-dialogue claims, corresponding to judgements, actionable suggestions, and questions. Each retained claim is represented using a frozen claim encoder. We also retain \semdetect's four review-level textual features for the target: perplexity, entropy, top-200 token percentage, and the \fastdetectgpt~\citep{ICLR2024_6b8c6f84} score.

\paragraph{Claim and textual-feature representation.}

Let $T=(t_1,\ldots,t_n)$ be the retained target-claim sequence, and let $R=(r_1,\ldots,r_m)$ contain the retained claims from the three references. Claims are concatenated in the fixed retained-category order while preserving extractor order within each category. References are concatenated into one pooled block in the frozen reference-selection order.

Claim embeddings share a learned projection into the model space and receive target/reference block and positional embeddings. Let $\hat z_l$ be the standardized textual feature $l\in\{1,\ldots,4\}$. Its token $F_l$ combines a learned feature-identity vector $e_l^{\mathrm{id}}$, the value along a learned value vector $e_l^{\mathrm{value}}$, and the shared textual-block embedding $b_{\mathrm{text}}$: $F_l=e_l^{\mathrm{id}}+\hat z_l e_l^{\mathrm{value}}+b_{\mathrm{text}}$.
Denoting the resulting target and reference vectors by $T_i$ and $R_j$, respectively, the encoder input is
\begin{equation}
X=
[\mathrm{CLS},F_1,\ldots,F_4,T_1,\ldots,T_n,R_1,\ldots,R_m].
\label{eq:unified-input}
\end{equation}
Claim sequences are padded within each mini-batch, with padding excluded from attention.

\paragraph{Directed blockwise-masked Transformer.}

An encoder-only Transformer~\citep{DBLP:conf/nips/VaswaniSPUJGKP17} applies multi-head self-attention under one fixed directed block mask (Figure~\ref{fig:detector}).

The reference block attends only within the pooled references, so its representation is not conditioned on the target review. The target block is the central integration space: it reads target claims, textual features, and reference evidence. Textual-feature tokens attend within their block and to target claims, so that review-level features are interpreted with review content. The CLS token reads itself and all three input blocks to form the final review representation; no other block reads CLS.

\paragraph{Classification and training.}

The final normalized CLS representation is passed through a shared MLP and two classification heads. The first predicts $a=P(\mathrm{AI})$; the second predicts $b=P(\mathrm{Rewrite}\mid\mathrm{non\mbox{-}AI})$. The resulting three-class distribution is
\begin{equation}
P(\mathrm{AI},\mathrm{Rewrite},\mathrm{Human})=[a,(1-a)b,(1-a)(1-b)].
\label{eq:classification-head}
\end{equation}
Training combines binary cross-entropy for AI/non-AI with conditional rewrite/human cross-entropy on non-AI examples. Claim extraction and claim embedding remain fixed; the feature-token parameters, claim projection, block and position embeddings, Transformer, and heads are optimized jointly. Appendices~\ref{app:attention-ablation} and~\ref{app:detector-implementation} give attention-mask ablations and implementation details, respectively.

\subsection{Adversarial Training Framework}
\label{sec:adversarial-training}

To train \peerrevealbase, we construct a regular three-class training corpus across ML, Nature Communications, and PeerJ. Each domain supplies manuscript--human-review pairs; conventional prompts generate AI reviews and rewrites. Appendix~\ref{app:training-corpora} gives corpus sizes and generation settings.

\begin{figure}[t]
\centering
\includegraphics[width=0.94\linewidth]{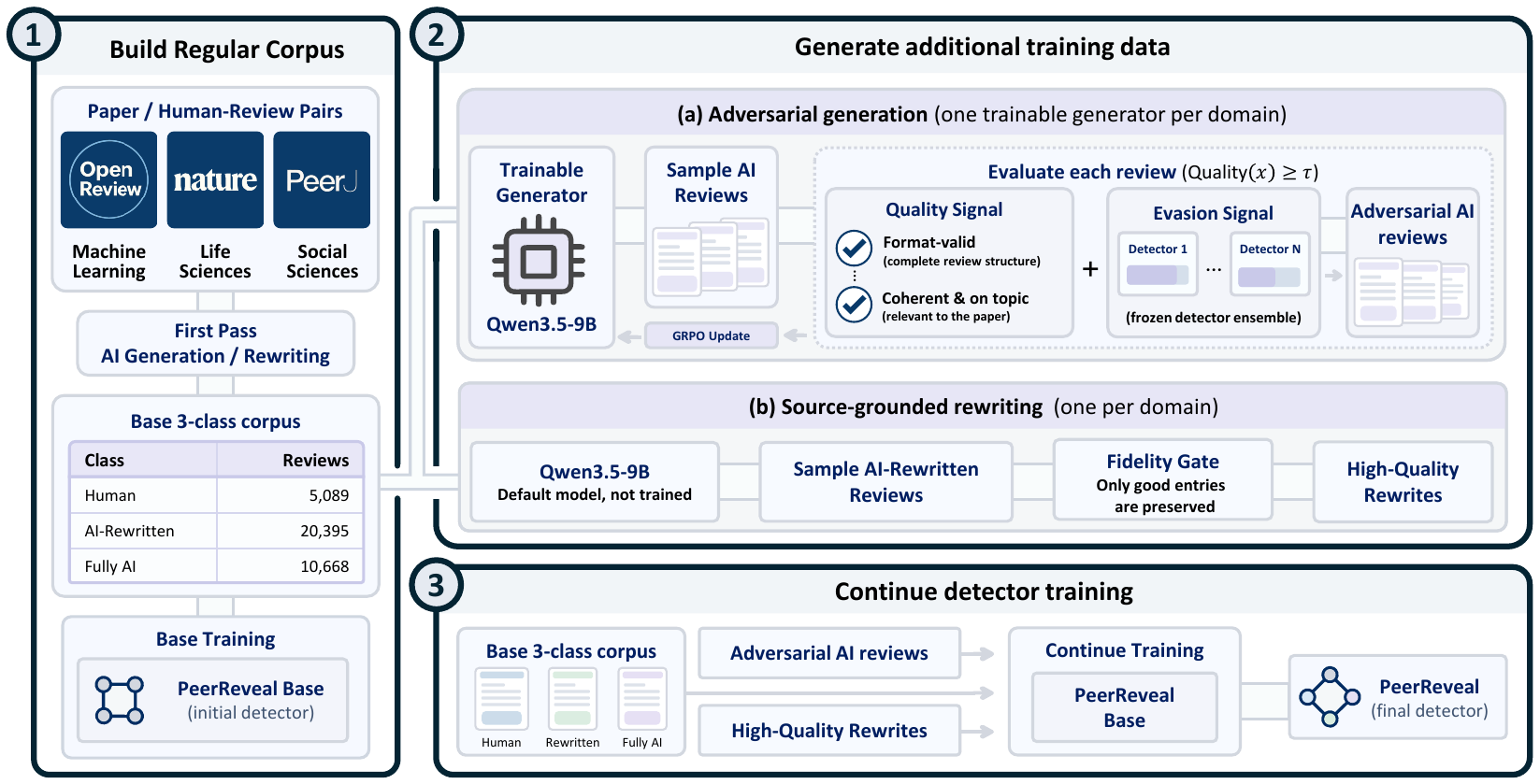}
\caption{Adversarial generator-detector training. Frozen detector and quality feedback optimize Qwen3.5-9B through GRPO. Three-domain regular training yields \peerrevealbase; adding adversarial AI reviews and source-grounded rewrites yields \peerreveal.}
\label{fig:adversarial-training-overview}
\end{figure}

We optimize local Qwen3.5-9B~\citep{qwen3.5} generators, one per domain, using review-quality feedback and feedback from a frozen detector ensemble $\mathcal{E}_{\mathrm{fixed}}$, comprising \semdetect, \peerreveal, \radar, and \anchor. Their adversarial AI reviews, together with source-grounded rewrites, augment the regular corpus to train \peerreveal. Figure~\ref{fig:adversarial-training-overview} summarizes this framework.

\subsection{Adversarial AI Review Generator}

\subsubsection{Strategy-Conditioned Exploration}

Given the same paper, requested rating, and strategy condition, Qwen3.5-9B independently generates four reviews, which form a comparison group for the GRPO~\citep{shao2024deepseekmathpushinglimitsmathematical} update.
The three conditions are \textbf{paper-only}, which reviews the complete manuscript directly; \textbf{human-frame}, which supplies sentence templates taken from real human reviews with their academic vocabulary blanked out and asks the writer to fill in the blanks from the manuscript; and \textbf{cached external reviews}, which supplies two same-paper reviews collected in advance and asks the writer to fuse them into one more human-sounding review.
Strategy conditioning diversifies the evidence attended to by the generator, the organization of the review, and the formation of critiques and judgements.

\subsubsection{Detector Feedback and GRPO Optimization}

Each sampled review is evaluated for format validity, review quality, and detector evasion. These evaluations provide the reward signals for GRPO optimization.
We combine evasion scores from multiple detectors to guide the generator toward reviews that are difficult across detection methods. Let $e_d(r_i)\in[0,1]$ denote detector $d$'s evasion score for review $r_i$, with larger values indicating greater evasion. The joint score is
\begin{equation}
E(r_i)=
\left(\sum_{d\in\mathcal E_{\mathrm{fixed}}}
w_d\sqrt{e_d(r_i)}\right)^2,
\qquad \sum_d w_d=1.
\label{eq:combined-evasion}
\end{equation}
The feedback combines the three-class detectors' human/rewrite evidence, \radar's human probability, and \anchor's reference-based decision. Weights are $.40$, $.30$, $.15$, and $.15$ for \peerreveal, \semdetect, \radar, and \anchor. Scorers stay frozen throughout optimization.

Within each group, reviews satisfying the format, quality, and length requirements receive group-relative evasion advantages. Advantages subtract the group mean evasion score of eligible reviews, and GRPO increases the relative likelihood of higher-scoring reviews. Appendices~\ref{app:detector-score-mappings} and~\ref{app:group-credit}--\ref{app:policy-optimization} detail calibration mappings, failure handling, and optimization.

\subsection{Source-Grounded Adversarial Rewrite Generation}

In the rewrite path, Qwen3.5-9B rewrites each human review given its paper abstract. It preserves the source judgement and reasons while exploring changes in expression, organization, and emphasis. Frozen detector scores guide candidate selection toward rewrites that evade correct classification.

Candidates enter detector training after passing a fidelity gate for judgement, reasons, and stance and a source-distance gate that removes near-copies. The paper abstract supplies grounding for references implicit in the source review. The AI-generation and rewrite paths provide complementary examples for three-class detector training.

\subsection{Adversarial Detector Update}

From \peerrevealbase, we continue training on regular reviews and adversarial AI and rewrite reviews from all domains to obtain \peerreveal, sampling by domain and class. Retaining regular data preserves supervision for all three classes while the new examples broaden the distribution.

\section{Experiments}
\label{sec:experiments}

\subsection{Implementation and Evaluation Setup}

\paragraph{Data and protocols.}
We run two comparisons. The ML-only comparison follows the regular training setting of \semdetect~\citep{duarte2026semdetect} and evaluates on four ML test sets: Original, which matches the training writer models and prompts; OOD-M, which uses unseen writer models; OOD-M+P, which uses unseen writer models and prompts; and PeerShift-ML. The main comparison trains on all three domains and evaluates on the full 5,430-review PeerShift test set from Section~\ref{sec:evaluation-construction}. Nature Communications supplies open peer-review records; PeerJ supplies matched initial manuscripts and first-round human reviews in Psychology and Social Sciences. The three-domain regular corpus contains 36,152 reviews and adversarial augmentation adds 8,464; training, development, and test papers are disjoint, and model selection uses development sets generated with the same mechanisms as PeerShift. Appendices~\ref{app:training-corpora} and~\ref{app:model-selection} detail the corpora, development sets, and training settings.

\paragraph{Metrics.}
For binary evaluation, fully AI-generated reviews are positive and the other two classes negative; we report AUROC and true-positive rate (TPR) at 1\% false-positive rate (FPR), both from continuous AI scores ($P(\mathrm{AI})$ for three-class detectors), and binary Macro-F1 at a decision threshold fixed before evaluation (Appendix~\ref{app:binary-results}). Three-class evaluation reports Macro-F1 and class recall. Pooled metrics combine all 5,430 predictions rather than averaging domain scores.

\subsection{Baselines}
We compare general-purpose detectors (\logrank~\citep{ippolito-etal-2020-automatic}, \fastdetectgpt~\citep{ICLR2024_6b8c6f84}, \binoculars~\citep{pmlr-v235-hans24a}, \mage~\citep{li-etal-2024-mage}, \radar~\citep{NEURIPS2023_30e15e59}, and \editlens~\citep{ICLR2026_c7d936e6}) and review-specific methods (\tf~\citep{kumar-etal-2024-quis}, \anchor~\citep{ICLR2026_64592277}, and \semdetect~\citep{duarte2026semdetect}). \anchor uses same-paper GPT-5.5 reference reviews. \semdetect is trained on the same regular corpus as \peerrevealbase and, for a matched comparison with \peerreveal, also on the augmented corpus.

We also fine-tune \roberta~\citep{liu2019robertarobustlyoptimizedbert} on full reviews with a three-class classification head, using the same regular and augmented training corpora.

\subsection{Research Questions}
Our experiments ask: (1)~how challenging PeerShift is for existing detectors (Section~\ref{sec:binary-results}); (2)~how well \peerreveal distinguishes three review sources across domains and improves with adversarial training (Section~\ref{sec:three-class-results}); (3)~how \peerreveal compares with \semdetect under its original training and test settings (Section~\ref{sec:generation-robustness}); and (4)~whether adversarial training enables small local 9B generators to evade multiple detectors (Section~\ref{sec:adversarial-results}).

\section{Results}
\label{sec:results}

\subsection{Existing Detectors Struggle on PeerShift}
\label{sec:binary-results}
Table~\ref{tab:binary-main} evaluates whether detectors recognize fully AI-generated judgement without flagging human judgement expressed with AI assistance, with human-written and AI-rewritten reviews as negatives. \peerreveal achieves the highest AUROC (98.02\%) and binary Macro-F1 (93.02\%), while adversarially trained \roberta leads in TPR at 1\% FPR (72.76\%; Figure~\ref{fig:roc-lowfpr}). For \peerreveal, PeerJ is the hardest domain at low FPR (TPR at 1\% FPR: 49.83\%, versus 77.54\% on ML and 50.50\% on Nature Communications); per-domain results and ROC curves are in Appendix~\ref{app:binary-results}.

\begin{table}[htbp]
\begin{minipage}[t]{0.555\linewidth}
\caption{Pooled binary detection on PeerShift (\%). Positive: AI-generated; negative: human-written and AI-rewritten. \semdetect and \roberta use matched regular/adversarial training corpora.}
\label{tab:binary-main}
\centering\small
\setlength{\tabcolsep}{2pt}
\begin{tabular}{lrrr}
\toprule
Detector & AUROC & TPR@1\% & Macro-F1\\
\midrule
\logrank & 35.39 & 0.88 & 40.03\\
\fastdetectgpt & 64.59 & 19.23 & 63.99\\
\binoculars & 44.24 & 0.00 & 49.39\\
\radar & 67.77 & 1.10 & 59.50\\
\anchor & 87.26 & 25.91 & 74.14\\
\tf & 55.52 & 6.19 & 50.98\\
\mage & 52.73 & 0.11 & 39.99\\
\editlens & 73.11 & 0.11 & 63.21\\
\midrule
\roberta (regular) & 77.59 & 49.72 & 67.66\\
\roberta (adversarial) & 89.10 & \textbf{72.76} & 83.76\\
\semdetect (regular) & 77.87 & 27.35 & 40.00\\
\semdetect (adversarial) & 87.13 & 18.73 & 74.59\\
\peerrevealbase & 92.32 & 34.86 & 82.80\\
\peerreveal & \textbf{98.02} & 60.77 & \textbf{93.02}\\
\bottomrule
\end{tabular}
\end{minipage}\hfill
\begin{minipage}[t]{0.415\linewidth}
\vspace{0pt}
\setlength{\belowcaptionskip}{0pt}
\centering
\includegraphics[width=\linewidth]{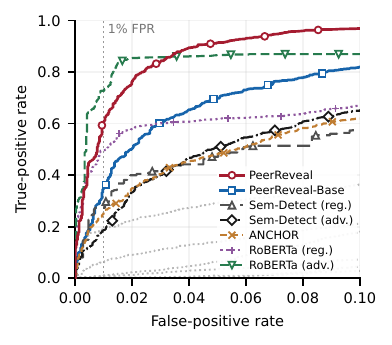}
\captionof{figure}{Pooled ROC curves on PeerShift (FPR 0--0.1). Weaker baselines are gray. Appendix~\ref{app:binary-results} gives full-range and per-domain ROC curves.}
\label{fig:roc-lowfpr}
\end{minipage}
\end{table}

\subsection{\peerreveal Improves Three-Class Detection, with Further Gains from Adversarial Training}
\label{sec:three-class-results}

Table~\ref{tab:three-domain-main} compares \semdetect, \roberta, and \peerreveal under matched regular and adversarial training. With regular training, \peerrevealbase reaches 70.52\% pooled Macro-F1, 11.48 points above \roberta; adversarial training with reviews from local 9B generators adds 11.88 points, bringing \peerreveal to 82.40\%, with the largest gain on ML (+21.22). \roberta also benefits from it, reaching 73.04\%, while \peerreveal remains 9.36 points ahead.

\begin{table}[htbp]
\begin{minipage}[t]{0.55\linewidth}
\caption{Three-class Macro-F1 (\%) with matched corpora; reg./adv.\ denote regular/augmented training.}
\label{tab:three-domain-main}
\centering\small
\setlength{\tabcolsep}{5pt}
\begin{tabular}{lrrrr}
\toprule
 & \multicolumn{4}{c}{PeerShift}\\
\cmidrule(lr){2-5}
Detector & ML & Nature & PeerJ & Pooled\\
\midrule
\roberta (reg.) & 57.01 & 58.54 & 61.07 & 59.04\\
\roberta (adv.) & 69.92 & 71.31 & 77.56 & 73.04\\
\semdetect (reg.) & 39.84 & 55.09 & 58.30 & 51.52\\
\semdetect (adv.) & 51.52 & 55.59 & 60.20 & 56.17\\
\midrule
\peerrevealbase & 58.73 & 73.93 & 77.67 & 70.52\\
\textbf{\peerreveal} & \textbf{79.95} & \textbf{81.98} & \textbf{85.18} & \textbf{82.40}\\
\bottomrule
\end{tabular}
\end{minipage}\hfill
\begin{minipage}[t]{0.42\linewidth}
\vspace{0pt}
\setlength{\belowcaptionskip}{0pt}
\centering
\includegraphics[width=\linewidth]{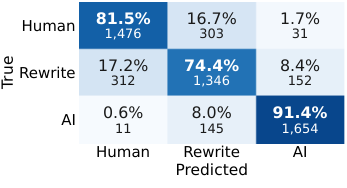}
\captionof{figure}{\peerreveal three-class confusion matrix on pooled PeerShift (\% by row). See Appendix~\ref{app:per-class-results} for class recalls.}
\label{fig:confusion-v1}
\end{minipage}
\end{table}

\paragraph{Class-wise performance.}

\peerreveal recalls 91.38\% of AI-generated, 74.36\% of rewritten, and 81.55\% of human reviews (Figure~\ref{fig:confusion-v1}). Human reviews are almost never labeled AI-generated (1.71\%); the errors lie on the human/rewrite boundary, where 16.74\% are predicted as rewrites. Appendix~\ref{app:examples} works through one paper's three reviews with every detector's output.

\subsection{\peerreveal Outperforms \semdetect under Its Original Training and Test Settings}
\label{sec:generation-robustness}

Under \semdetect's ML-only regular training (Table~\ref{tab:ml-only}), \peerrevealbase leads by 11.83 points on Original, 21.55 on OOD-M, and 18.84 on OOD-M+P. Its 48.14\% on PeerShift-ML marks the remaining generalization gap under broader review-generation conditions (Appendix~\ref{app:ml-only}).

\begin{table}[htbp]
\caption{Comparison under \semdetect's ML-only regular training: pooled Macro-F1 (\%).}
\label{tab:ml-only}
\centering\small
\begin{tabular}{lrrrr}
\toprule
Detector & Original & OOD-M & OOD-M+P & PeerShift-ML\\
\midrule
\semdetect & 84.14 & 71.59 & 69.18 & 34.40\\
\peerrevealbase & \textbf{95.97} & \textbf{93.14} & \textbf{88.01} & \textbf{48.14}\\
\midrule
Gain (pp) & +11.83 & +21.55 & +18.84 & +13.74\\
\bottomrule
\end{tabular}
\end{table}

\subsection{Adversarially Trained Local 9B Generators Evade Multiple Detectors}
\label{sec:adversarial-results}

Table~\ref{tab:generator-evasion} evaluates the local Qwen3.5-9B~\citep{qwen3.5} generators before and after adversarial training. Evasion is the fraction of quality-eligible reviews classified as non-AI on held-out papers (Appendix~\ref{app:generator-evaluation}). Training increases evasion for most detectors, including \editlens and \tf, neither used for feedback. \peerreveal is less evaded than \peerrevealbase at both thresholds; Appendix~\ref{app:generator-example} shows an adversarial review it detects. These local 9B generators enable effective and practical adversarial training (Table~\ref{tab:three-domain-main}), with gains that transfer to PeerShift's mechanisms and writers absent from generator training.

\begin{table}[htbp]
\caption{Pooled evasion (\%) before/after generator training at 1\% and 5\% human FPR, among quality-eligible reviews. $\dagger$ marks detector families used for generator-training feedback. We omit weaker detectors exceeding 91\% evasion before training at 1\% FPR; Appendix~\ref{app:generator-evaluation} gives all details.}
\label{tab:generator-evasion}
\centering\small
\begin{tabular}{lrrrr}
\toprule
Detector & \multicolumn{2}{c}{1\% FPR} & \multicolumn{2}{c}{5\% FPR}\\
 & Before & After & Before & After\\
\midrule
\radar{}$^\dagger$ & 98.80 & 100.00 & 85.29 & 78.55\\
\anchor{}$^\dagger$ & 76.28 & 99.67 & 41.74 & 95.05\\
\tf & 88.29 & 94.06 & 58.26 & 76.90\\
\editlens & 59.76 & 88.45 & 0.30 & 40.59\\
\roberta (regular) & 12.91 & 43.89 & 10.21 & 30.03\\
\roberta (adversarial) & 0.90 & 42.90 & 0.90 & 42.57\\
\midrule
\semdetect (regular)$^\dagger$ & 79.88 & 100.00 & 61.56 & 100.00\\
\peerrevealbase{}$^\dagger$ & 55.26 & 99.01 & 18.02 & 95.71\\
\peerreveal{}$^\dagger$ & 6.31 & 72.28 & 0.00 & 48.84\\
\bottomrule
\end{tabular}
\end{table}

\subsection{\texorpdfstring{\rev{\peerreveal Generalizes beyond PeerShift}}{PeerReveal Generalizes beyond PeerShift}}
\label{sec:beyond-peershift}
\begin{revblock}
We test all detectors, unchanged, on two recent peer-review benchmarks built by other groups, \citet{saha2026policies} and PeerPrism~\citep{sadeghian2026peerprism}. Both contain human-written, LLM-polished, and fully LLM-generated reviews from writer models absent from our training data (Appendix~\ref{app:external-benchmarks}). \peerreveal achieves the highest three-class Macro-F1 on both (Table~\ref{tab:external-benchmarks-main}).

To test unseen real review workflows, we also run two public frontier LLM reviewing agents, Google Deep Review\footnote{\url{https://research.google/deepreview/}} and the Stanford Agentic Reviewer\footnote{\url{https://paperreview.ai}}, on PeerShift-ML test papers (Appendix~\ref{app:external-agents}). \peerreveal labels all 35 of their reviews as AI-generated, whereas every other detector misses most Deep Review reviews (Table~\ref{tab:external-agents}).

When one domain is held out from training and development, \peerreveal loses 6--9 points on that domain but still outperforms \peerrevealbase trained the same way and \semdetect trained on all three domains (Appendix~\ref{app:held-out-domains}).

\begin{table}[htbp]
\begin{minipage}[t]{0.52\linewidth}
\caption{Three-class Macro-F1 (\%) on external benchmarks. $^\dagger$Score-based detector with two cut points set on our calibration panel. Appendix~\ref{app:external-benchmarks} lists all detectors.}
\label{tab:external-benchmarks-main}
\centering\small
\setlength{\tabcolsep}{4pt}
\begin{tabular}{lrr}
\toprule
Detector & \citet{saha2026policies} & PeerPrism\\
\midrule
\roberta (reg.) & 66.59 & 72.88\\
\roberta (adv.) & 66.48 & 77.84\\
\semdetect (reg.) & 52.67 & 56.68\\
\semdetect (adv.) & 50.36 & 55.28\\
\editlens$^\dagger$ & 64.12 & 60.94\\
\tf$^\dagger$ & 55.48 & 47.37\\
\fastdetectgpt$^\dagger$ & 54.90 & 32.76\\
\anchor$^\dagger$ & 45.79 & 47.61\\
\midrule
\peerrevealbase & 62.87 & 75.91\\
\textbf{\peerreveal} & \textbf{67.79} & \textbf{78.16}\\
\bottomrule
\end{tabular}
\end{minipage}\hfill
\begin{minipage}[t]{0.45\linewidth}
\caption{Reviews labeled AI-generated, out of all reviews written by Google Deep Review and the Stanford Agentic Reviewer.}
\label{tab:external-agents}
\centering\small
\setlength{\tabcolsep}{3pt}
\begin{tabular}{lrr}
\toprule
Detector & Deep Review & Agentic\\
\midrule
\roberta (reg.) & 1/10 & 25/25\\
\roberta (adv.) & 3/10 & 25/25\\
\semdetect (reg.) & 0/10 & 14/25\\
\semdetect (adv.) & 2/10 & 20/25\\
\midrule
\peerrevealbase & 4/10 & 25/25\\
\textbf{\peerreveal} & \textbf{10/10} & \textbf{25/25}\\
\bottomrule
\end{tabular}
\end{minipage}
\end{table}
\end{revblock}

\section{Scope and Limitations}
\label{sec:analysis-limitations}

PeerShift expands model, workflow, and domain coverage, but cannot exhaust future review-production practices. \rev{Its AI-generated and AI-rewritten reviews come from our own pipelines, because reviews with known true authorship cannot be collected at scale. Section~\ref{sec:beyond-peershift} tests review production that we did not design, but its agent evaluation covers only 35 reviews. A larger collection of real AI-assisted reviews would give a stronger test.}

\begin{revblock}
Our three classes give one label per review, while a real review may mix human and AI-generated claims. On such reviews, \peerreveal labels more reviews as AI-generated as the AI-generated share grows (Appendix~\ref{app:hybrid}), but one label cannot show which claims came from the LLM. Because \peerreveal already represents each claim separately, attributing individual claims is a natural next step.

The adversarially trained generators still evade the final detector on about half of their reviews at 5\% human FPR (Table~\ref{tab:generator-evasion}). A fixed detector can always be targeted by an adversary trained against it, so robustness in detection is an eternal goal. Our framework can follow this goal by alternating the two sides: generators are retrained against the updated detector, and their reviews train the next detector. We ran one such round, with 9B generators (Qwen3.5, and GLM-4 on ML; Appendix~\ref{app:second-generator}); more rounds and larger generators remain to be studied.
\end{revblock}

Performance depends on the quality of claim extraction and manuscript-conditioned references. In particular, distinguishing faithful rewriting from human-authored expression remains harder than recognizing fully AI-generated judgement. Detection should therefore support review-process analysis rather than serve as sole evidence for individual misconduct.

\section{Conclusion}
\label{sec:conclusion}

\peerreveal combines learned claim interactions with adversarial supervision from a local Qwen3.5-9B generator. Quality-constrained mechanism search produces PeerShift, covering diverse review-production processes across three domains. Under matched corpora, \peerreveal outperforms \semdetect and \roberta in three-class detection under both training regimes, reaching 82.40\% pooled Macro-F1 on 5,430 reviews. These results show the value of learning from individual claims and diverse training reviews.

\subsection*{AI use statement}
Generative AI tools were used to assist with experiment orchestration, software development, data processing, and drafting and editing parts of the manuscript. The authors reviewed all AI-assisted work and take responsibility for the final scientific content, claims, analyses, and artifacts.

\subsection*{Ethics statement}
\peerreveal is intended to help editors and program chairs understand review provenance, not to adjudicate individual cases. Its outputs are probabilistic, its errors concentrate on the boundary between human-written and AI-rewritten reviews, and false accusations carry real costs for reviewers; detector predictions should therefore serve as one signal among several rather than as grounds for action on their own. All human reviews come from venues with public review records, reviewer identities are not used, and generated and rewritten reviews are used only for training and evaluation. Adversarial generators pose dual-use risks; here they serve only to improve detection.

\subsection*{Reproducibility statement}
Appendix~\ref{app:mechanism-details} gives the mechanism-search controller, the operation families, the numerical acceptance constraints, the frozen mechanisms selected for each domain, and their exact prompt texts; Appendix~\ref{app:generator-training-details} gives the detector-score calibration, the quality judge with its rubric and prompt, GRPO settings, and the generator evaluation protocol; Appendix~\ref{app:attention-ablation} gives all 15 attention structures with their results; Appendix~\ref{app:data-protocols} gives corpus composition, paper-level splits, training and model-selection settings, baseline implementations, the fixed decision rule for binary metrics, per-class results, human-versus-AI binary results, and generation and detection cost; Appendix~\ref{app:detector-implementation} gives the frozen claim-extraction, reference-selection, and encoding pipeline together with every detector hyperparameter; and Appendix~\ref{app:examples} gives one paper's three reviews in full with every detector's output on them. Final training configurations and execution commands accompany the code.

\defcitealias{kimiteam2026kimik25visualagentic}{Kimi Team et al., 2026}
\bibliography{references,references_additions}

\begin{thebibliography}{40}
\providecommand{\natexlab}[1]{#1}
\providecommand{\url}[1]{\texttt{#1}}
\expandafter\ifx\csname urlstyle\endcsname\relax
  \providecommand{\doi}[1]{doi: #1}\else
  \providecommand{\doi}{doi: \begingroup \urlstyle{rm}\Url}\fi

\bibitem[Bao et~al.(2024)Bao, Zhao, Teng, Yang, and Zhang]{ICLR2024_6b8c6f84}
Guangsheng Bao, Yanbin Zhao, Zhiyang Teng, Linyi Yang, and Yue Zhang.
\newblock Fast-DetectGPT: Efficient Zero-Shot Detection of Machine-Generated
  Text via Conditional Probability Curvature.
\newblock In B.~Kim, Y.~Yue, S.~Chaudhuri, K.~Fragkiadaki, M.~Khan, and Y.~Sun
  (eds.), \emph{International Conference on Learning Representations}, volume
  2024, pp.\  24814--24836, 2024.
\newblock URL
  \url{https://proceedings.iclr.cc/paper_files/paper/2024/file/6b8c6f846c3575e1d1ad496abea28826-Paper-Conference.pdf}.

\bibitem[Barde et~al.(2022)Barde, Karch, Nowrouzezahrai, Moulin{-}Frier, Pal,
  and Oudeyer]{DBLP:conf/iclr/BardeKNMPO22}
Paul Barde, Tristan Karch, Derek Nowrouzezahrai, Cl{\'{e}}ment Moulin{-}Frier,
  Christopher Pal, and Pierre{-}Yves Oudeyer.
\newblock Learning to Guide and to be Guided in the Architect-Builder Problem.
\newblock In \emph{The Tenth International Conference on Learning
  Representations, {ICLR} 2022, Virtual Event, April 25-29, 2022}.
  OpenReview.net, 2022.
\newblock URL \url{https://openreview.net/forum?id=swiyAeGzFhQ}.

\bibitem[Beltagy et~al.(2020)Beltagy, Peters, and
  Cohan]{beltagy2020longformerlongdocumenttransformer}
Iz~Beltagy, Matthew~E. Peters, and Arman Cohan.
\newblock Longformer: The Long-Document Transformer, 2020.
\newblock URL \url{https://arxiv.org/abs/2004.05150}.

\bibitem[Chen et~al.(2024)Chen, Wang, Chen, Yu, Ma, Zhao, Zhang, and
  Yu]{chen-etal-2024-dense}
Tong Chen, Hongwei Wang, Sihao Chen, Wenhao Yu, Kaixin Ma, Xinran Zhao,
  Hongming Zhang, and Dong Yu.
\newblock Dense {X} Retrieval: What Retrieval Granularity Should We Use?
\newblock In Yaser Al-Onaizan, Mohit Bansal, and Yun-Nung Chen (eds.),
  \emph{Proceedings of the 2024 Conference on Empirical Methods in Natural
  Language Processing}, pp.\  15159--15177, Miami, Florida, USA, November 2024.
  Association for Computational Linguistics.
\newblock \doi{10.18653/v1/2024.emnlp-main.845}.
\newblock URL \url{https://aclanthology.org/2024.emnlp-main.845/}.

\bibitem[Comanici et~al.(2025)Comanici, Bieber, Schaekermann, Pasupat,
  Sachdeva, Dhillon, Blistein, Ram, Zhang, Rosen, Marris, Petulla, Gaffney,
  Aharoni, Lintz, Pais, Jacobsson, Szpektor, Jiang, Haridasan, Omran, Saunshi,
  Bahri, Mishra, Chu, Boyd, Hekman, Parisi, Zhang, Kawintiranon, Bedrax-Weiss,
  Wang, Xu, Purkiss, Mendlovic, Deutel, Nguyen, Langley, Korn, Rossazza, Ramé,
  Waghmare, Miller, Byrd, Sheshan, Hadsell, Bhardwaj, Janus, Rissa, Horgan,
  Abdagic, Belenki, Allingham, Singh, Guidroz, Srinivasan, Schmit, Chiafullo,
  Elisseeff, Jha, Kolhar, Berrada, Ding, Si, Mallick, Och, Erell, Ni, Latkar,
  Yang, Sirkovic, Feng, Leland, Hornung, Wu, Blundell, Alvari, Huang, Yip,
  Deur, Liu, Surita, Duque, Damen, Jia, Guez, Mircea, Sinha, Magni, Stradomski,
  Marian, Galić, Chen, Husain, Singhal, Grewe, Aubet, Song, Blanco, Rechis,
  Ho, Munoz, Zheng, Hamrick, Mather, Taitelbaum, Rutherford, Lei, Chen, Shukla,
  Moreira, Doi, Isik, Shabat, Rogozińska, Kolipaka, Chang, Vušak,
  Venkatachary, Noghabi, Bharti, Jun, Zaks, Green, Challagundla, Wong,
  Mohammad, Hirsch, Cheng, Naim, Proleev, Vincent, Singh, Krikun, Krishnan,
  Ghahramani, Atias, Aggarwal, Kirov, Vytiniotis, Koh, Chronopoulou, Dogra,
  Ion, Tyen, Lee, Weissenberger, Strohman, Balakrishna, Rae, Velic,
  de~Liedekerke, Elyada, Yuan, Liu, Shani, Kishchenko, Alessio, Li, Song, Kwei,
  Jankowski, Pappu, Namiki, Ma, Tripuraneni, Cherry, Ikonomidis, Ling, Ji,
  Westberg, Wright, Yu, Parkinson, Ramaswamy, Connor, Yeganeh, Grover,
  Kenwright, Litchev, Apps, Tomala, Halim, Castro-Ros, Li, Boral, Sho, Yarom,
  Malmi, Klinghoffer, Lin, Ansell, S, Zhao, Zuo, Santoro, Cheng, Demmessie,
  Liu, Brichtova, Culp, Braun, Graur, Ng, Mehta, Phillips, Sundberg, Godbole,
  Liu, Katariya, Rim, Seyedhosseini, Ammirati, Valfridsson, Malihi, Knight,
  Toor, Lampe, Ittycheriah, Chiang, Yeung, Fréchette, Rao, Wang, Srivastava,
  Zhang, Rhodes, Brand, Weesner, Figotin, Gimeno, Fellinger, Marcenac, Leal,
  Marcus, Cotruta, Cabrera, Luo, Garrette, Axelrod, Baltateanu, Barker, Chen,
  Toma, Ingram, Riesa, Kulkarni, Zhang, Liu, Wang, Polacek, Wu, Hui, Reyes, Su,
  Barnes, Malhi, Siddiqui, Feng, Damaschin, Pighin, Steiner, Yang, Boppana,
  Ivanov, Kandoor, Shah, Mujika, Huang, Choquette-Choo, Patel, Yu, Creswell,
  Jerry, Liu, Barros, Razeghi, Roy, Culliton, Xiong, Pan, Strohmann, Powell,
  Seal, DeCarlo, Shyam, Katircioglu, Wang, Hardin, Odisho, Broder, Chang, Nair,
  Shtefan, O'Brien, Agarwal, Potluri, Goyal, Jhindal, Thakur, Stuken, Lyon,
  Toutanova, Feng, Wu, Horn, Wang, Cullum, Taubman, Shrivastava, Shi,
  Tomlinson, Patel, Tu, Oflazer, Pongetti, Yang, Taïga, Perot, Pierse, Han,
  Drori, Iturrate, Chakrabarti, Yeung, Dopson, ting Chen, Kulshreshtha, Guo,
  Pham, Schuster, Chen, Polozov, Xing, Zhou, Kacham, Kukliansky, Miech,
  Yaroshenko, Chi, Douglas, Fei, Blondel, Myla, Madmoni, Wu, Keysers, Kjems,
  Albuquerque, Yu, D'sa, Plantan, Ionescu, Elias, Gupta, Vuyyuru, Alcober,
  Zhou, Ji, Hartmann, Puttagunta, Song, Amid, Stefanoiu, Lee, Pucciarelli,
  Wang, Raul, Petrov, Tian, Anklin, Nti, Gomes, Schumacher, Vesom,
  Panagopoulos, Bousmalis, Andor, Jacob, Zhang, Rosgen, Kecman, Tung, Belias,
  Goodman, Covington, Wieder, Saxena, Davoodi, Huang, Maddineni, Roulet,
  Campbell-Ajala, Sessa, Xintian, Wu, Lai, Collins, Haig, Sakenas, Xu,
  Giustina, Shafey, Charoenpanit, Garg, Ainslie, Severson, Arenas, Pathak,
  Rajayogam, Feng, Bakker, Li, Wichers, Rogers, Geng, Li, Jagerman, Jia,
  Olmert, Sharon, Mauger, Mariserla, Ma, Mohabey, Kim, Andreev, Pollom, Love,
  Jain, Agrawal, Schroecker, Fortin, Warmuth, Liu, Leach, Blok, Girirajan,
  Aharoni, Uria, Sozanschi, Goldberg, Ionita, Ribeiro, Zlocha, Birodkar,
  Lachgar, Yuan, Choudhury, Ginsberg, Zheng, Dibb, Graves, Lokhande, Rasskin,
  Muraru, Quick, Tata, Sermanet, Chawla, Karo, Wang, Zhang, Keller, Dragan, Su,
  Chou, Liu, Tao, Prabhakara, Wilson, Liu, Wang, Evans, Du, Castaño, Prasad,
  Mahdy, Gerlach, Reid, Kahn, Zait, Pillai, Ulrich, Wang, Wassenberg, Farkash,
  Yalasangi, Wang, Bauza, Bucher, Liu, Yan, Leung, Sindhwani, Barnes, Singh,
  Jurin, Chang, Bhumihar, Eiger, Citovsky, Withbroe, Li, Xue, Santo, Stoyanov,
  Raimond, Zheng, Gao, Listík, Kwasiborski, Saputro, Ozturel, Mallya,
  Majmundar, West, Caron, Wei, Castrejon, Vikram, Ramachandran, Dhawan, Park,
  Smoot, van~den Driessche, Blau, Malik, Liang, Hirsch, dos Santos, Weinstein,
  van~den Oord, Lall, FitzGerald, Jiang, Yang, Webster, Elqursh, Pope, Rotival,
  Raposo, Zhu, Dean, Alabed, Tran, Gupta, Gleicher, Austin, Rosseel, Umekar,
  Das, Sun, Chen, Misiunas, Zhou, Di, Loo, Newlan, Li, Ramasesh, Xu, Chen,
  Gandhe, Soricut, Gupta, Hu, El-Sayed, Garcia, Brusilovsky, Chen, Bolt, Huang,
  Gurney, Zhang, Pritzel, Wilkiewicz, Seybold, Shamanna, Fischer, Dean, Gill,
  Mcilroy, Bhowmick, Selier, Yang, Cheng, Magay, Tan, Varma, Walder, Kocisky,
  Nakashima, Natsev, Kwong, Gog, Zhang, Dieleman, Jimma, Ryabtsev, Brahma,
  Steiner, Du, Žužul, Žanić, Raghavachari, Gierke, Zheng, Petrova, Dauphin,
  Liu, Kessler, Hand, Duvarney, Kim, Lee, Hussenot, Hui, Smith, Jain, Xia,
  Tomar, Amiri, Phan, Fuchs, Weyand, Tomasev, Cordell, Liu, Mallinson, Joshi,
  Crawford, Suggala, Chien, Fernando, Sanchez-Vargas, Williams, Crone, Luo,
  Karpov, Shan, Thurk, Strudel, Voigtlaender, Patil, Dozat, Khodaei, Singla,
  Ambroszczyk, Wu, Chang, Roark, Hegde, Ding, Filos, Wu, Pinto, Liu, Khanna,
  Pandey, Mcloughlin, Li, Haves, Zhou, Buchatskaya, Leal, de~Boursac, Akazawa,
  Anderson, Chen, Somandepalli, Liang, Goenka, Winkler, Grushetsky, Ding,
  Smith, Ye, Pont-Tuset, Li, Li, Golany, Wegner, Jiang, Barak, Shangguan,
  Vértes, Wong, Bornschein, Tudor, Bevilacqua, Schaul, Rawat, Zhao, Axiotis,
  Meng, McLean, Lai, Beattie, Kushman, Liu, Kutzman, Lang, Ye, Netrapalli,
  Mishra, Khan, Goel, Willoughby, Tian, Zhuang, Chen, Tsai, Kementsietsidis,
  Khare, Keeling, Xu, Waters, Altché, Popat, Mittal, Saxton, Badawy, Mathieu,
  Zheng, Zhou, Ranka, Shin, Duan, Salimans, Mihailescu, Shaham, Chang, Assael,
  Dikkala, Izzard, Cohen-Addad, Graves, Feinberg, Chung, Strouse, Karmon,
  Sharifzadeh, Ashwood, Pham, Blanton, Vasiloff, Barber, Geller, Zhou, Zubach,
  Huang, Zhang, Gupta, Young, Proskurnia, Votel, Gabeur, Barcik, Tripathi, Yu,
  Yan, Changpinyo, Pavetić, Coyle, Fujii, Mendez, Zhou, Rajamani, Hechtman,
  Cao, Juan, Tan, Dalibard, Du, Clay, Yao, Jia, Vijaykumar, Zhou, Bai, Hung,
  Pecht, Todorov, Khadke, Gupta, Lahoti, Autef, Duddu, Lee-Thorp, Bykovsky,
  Misiunas, Flennerhag, Thangaraj, McGiffin, Nado, Kunesch, Noever, Hertz,
  Liang, Stone, Palmer, Daruki, Pramanik, Põder, Kyker, Khan, Sluzhaev,
  Ritter, Ruderman, Zhou, Nagpal, Vodrahalli, Necula, Barham, Pavlick,
  Hartford, Shafran, Zhao, Mikuła, Eccles, Shimokawa, Garg, Vilnis, Chen,
  Shumailov, Lee, Abdelhamed, Xie, Cohen, Hlavnova, Malkin, Sitawarin, Lottes,
  Coquinot, Yu, Kumar, Zhang, Mahendru, Ahmed, Martens, Chen, Boag, Peng,
  Devin, Klimovskiy, Phuong, Vainstein, Xie, Ramabhadran, Howard, Yu, Goswami,
  Cui, Shleifer, Pinto, Yeh, Yang, Javanmardi, Ethier, Lee, Orbay, Kotecha,
  Bromberg, Shaw, Thornton, Rosenthal, Gu, Thomas, Gemp, Ayyar, Ushio, Selvan,
  Wee, Liu, Majzoubi, Yu, Abernethy, Liechty, Pan, Nguyen, Qiong, Hu, Perrin,
  Arora, Pitler, Wang, Shivakumar, Prost, Limonchik, Wang, Gao, Cour, Buch,
  Gui, Ivanova, Neubeck, Chan, Kim, Chen, Goyal, Chung, Liu, Su, Petrushkina,
  Shen, Joulin, Xu, Lin, Kulizhskaya, Chelba, Vasudevan, Collins, Bashlovkina,
  Lu, Fritz, Park, Zhou, Su, Tanburn, Sushkov, Rasquinha, Li, Prendki, Li, LV,
  Sharma, Fitoussi, Huang, Dai, Dao, Burrows, Prior, Qin, Pundak, Sjoesund,
  Khurshudov, Zhu, Webson, Kemp, Tan, Agrawal, Sargsyan, Cheng, Stephan,
  Kwiatkowski, Reid, Byravan, Michaely, Heess, Zhou, Goenka, Carpenter,
  Levskaya, Wang, Roberts, Leblond, Chikkerur, Ginzburg, Chang, Riachi,
  Chuqiao, Xu, Borsos, Pliskin, Pawar, Lustman, Kirkwood, Anand, Chaudhary,
  Kalb, Milan, Augenstein, Goldie, Prince, Raman, Sun, Xia, Cohen, Huo, Camp,
  Ellis, Zilka, Torres, Patel, Arora, Chan, Adler, Ayoub, Liang, Jamil, Jiang,
  Baumgartner, Sun, Karov, Akulov, Zheng, Cai, Fantacci, Rubin, Acha, Wang,
  D'Souza, Sathyanarayana, Dai, Rowe, Simanovsky, Goldman, Kuang, Pan,
  Rosenberg, Rojas-Esponda, Dutta, Zeng, Jurenka, Farquhar, Bansal, Iqbal,
  Roelofs, Joung, Beak, Ryu, Poplin, Wu, Alayrac, Buthpitiya, Ronneberger,
  Habtegebriel, Li, Cavallaro, Wei, Bensky, Denk, Ganapathy, Stanway, Joshi,
  Bertolini, Lo, Ma, Charles, Sampemane, Sahni, Chen, Askham, Gaddy, Young,
  Tan, Eyal, Bražinskas, Zhong, Wu, Epstein, Bailey, Hard, Lee, Goldshtein,
  Ruiz, Badawi, Lochbrunner, Kearns, Brown, Pardo, Weber, Yang, Jiang, Akin,
  Fu, Wainwright, Zou, Gaba, Manzagol, Kan, Song, Zainullina, Lin, Ko,
  Deshmukh, Jindal, Svensson, Tyam, Zhao, Kaeser-Chen, Baird, Moradi, Hall,
  Guo, Tsang, Liang, Pereira, Ganesh, Korotkov, Adamek, Thiagarajan, Tran,
  Chen, Tar, Jain, Dasgupta, Bilal, Reitter, Zhao, Vezzani, Gehman, Mehta,
  Beltrone, Dotiwalla, Guadarrama, Abbas, Karp, Georgiev, Ferng, Brockschmidt,
  Peng, Hirnschall, Verma, Bi, Xiao, Dabush, Xu, Wallis, Parker, Wang, Xu,
  Safarli, Tewari, Zhang, Kim, Gesmundo, Thomas, Levi, Chowdhury, Rao, Garst,
  Conway-Rahman, Ran, McKinney, Xiao, Yu, Agrawal, Stjerngren, Ionescu, Chen,
  Sharma, Chiu, Liu, Franko, Sanford, Cai, Michel, Ganapathy, Labanowski,
  Garrett, Vargas, Sun, Gale, Buschmann, Desjardins, Ghelani, Jain, Verma,
  Asawaroengchai, Eisenschlos, Harlalka, Kazawa, Metzler, Howland, Jian, Ades,
  Shah, Gangwani, Lee, Ring, Hernandez, Reich, Sinha, Sathe, Kovac, Gill,
  Kannan, D'olimpio, Sevenich, Whang, Kim, Sim, Chen, Zhang, Lall, Matias, Jia,
  Friesen, Nasso, Thapliyal, Perozzi, Yu, Shekhawat, Huda, Grabowski, Wang,
  Sreevatsa, Dib, Hassen, Schuh, Milutinovic, Welty, Quinn, Shah, Wang,
  Barth-Maron, Frye, Axelsson, Zhu, Ma, Giannoumis, Sedghi, Ye, Luan, Aydin,
  Chandra, Sampathkumar, Huang, Lavrenko, Eleryan, Hong, Hansen, Carthy,
  Samanta, Ćevid, Wang, Li, Voznesensky, Hoffman, Terzis, Sehwag, Fidel, He,
  Cai, He, Feng, Nikoltchev, Phatale, Chase, Lawton, Zhang, Ouyang, Tragut,
  Manshadi, Narayanan, Shen, Gao, Bolukbasi, Roy, Li, Golovin, Panait, Qin,
  Han, Anthony, Kudugunta, Patraucean, Ray, Chen, Yang, Bhatia, Talluri,
  Morris, Ražnatović, Brownfield, An, Peng, Kane, Zheng, Duduta, Kessinger,
  Noraky, Liu, Rong, Veličković, Rush, Goldin, Wei, Garlapati, Pantofaru,
  Kwon, Ni, Noland, Trapani, Beaufays, Roy, Chow, Turker, Cideron, Mei, Clark,
  Dou, Bošnjak, Leith, Du, Yazdanbakhsh, Nasr, Kwak, Sheth, Kaskasoli, Anand,
  Lakshminarayanan, Jerome, Bieber, Chu, Senges, Shen, Sridhar, Ndebele,
  Beyret, Mohamed, Chen, Freitag, Guo, Liu, Roit, Chen, Yan, Stone, Co-Reyes,
  Cole, Scellato, Azizi, Hashemi, Jin, Iyer, Valentine, György, Ahuja, Diaz,
  Lee, Clement, Kong, Garmon, Watts, Bhatia, Gupta, Miecnikowski, Vallet, Taly,
  Loper, Joshi, Atwood, Chick, Collier, Iliopoulos, Trostle, Gunel,
  Leal-Cavazos, Hrafnkelsson, Guzman, Ju, Forbes, Emond, Chauhan, Caine, Xiao,
  Zeng, Moufarek, Murphy, Meng, Gupta, Riedel, Das, Lawal, Narayan, Sosea,
  Swirhun, Friso, Neyshabur, Lu, Girgin, Wunder, Yvinec, Pyne, Carbune,
  Rijhwani, Guo, Doshi, Briukhov, Bain, Hitron, Wang, Gupta, Chen, Du, Zhang,
  Shah, Akula, Dylla, Kachra, Kuo, Zou, Wang, Xu, Zhu, Snyder, Menon, Firat,
  Mordatch, Yuan, Ponomareva, Blevins, Moore, Wang, Chen, Scholz, Dwornik, Lin,
  Li, Antognini, I, Song, Miller, Kalra, Raveret, Akerlund, Wu, Nystrom,
  Godbole, Liu, DeBalsi, Zhao, Liu, Caciularu, Lax, Khandelwal, Langston,
  Bailey, Lattanzi, Wang, Kovelamudi, Mondal, Guruganesh, Hua, Roval,
  Wesołowski, Ingale, Halcrow, Sohn, Angermueller, Raad, Stickgold, Lu, Kosik,
  Xie, Lillicrap, Huang, Zhang, Paulus, Farabet, Wertheim, Wang, Joshi, ling
  Ko, Wu, Agrawal, Lin, Sheng, Sung, Breland-King, Butterfield, Gawde, Singh,
  Zhang, Apte, Shetty, Hutter, Li, Salesky, Lebron, Kanerva, Paganini, Nguyen,
  Vallu, Peter, Velury, Kao, Hoover, Bortsova, Bishop, Jakobovits, Agostini,
  Agarwal, Liu, Kwong, Tavakkol, Bica, Greve, GP, Marcus, Hou, Duerig,
  Moroshko, Lacey, Davis, Amelot, Wang, Kim, Strinopoulos, Wan, Lan, Krishnan,
  Tang, Humphreys, Bai, Shtacher, Machado, Pang, Burke, Liu, Aravamudhan, Song,
  Hirst, Singh, Jou, Bai, Piccinno, Fu, Alazard, Meiri, Winter, Chen, Zhang,
  Heitkaemper, Lambert, Lee, Frömmgen, Rogulenko, Nair, Niemczyk, Bulyenov,
  Xu, Shemtov, Zadimoghaddam, Toropov, Wirth, Dai, Gollapudi, Zheng, Kurakin,
  Lee, Bullard, Serrano, Balazevic, Li, Schalkwyk, Murphy, Zhang, Sequeira,
  Datta, Agrawal, Sutton, Attaluri, Chiang, Farhan, Thornton, Lin, Choma,
  Nguyen, Dasgupta, Robinson, Comşa, Riley, Pillai, Mustafa, Golan, Zandieh,
  Lespiau, Porter, Ross, Rajayogam, Agarwal, Venugopalan, Shahriari, Yan, Xu,
  Tobin, Dubov, Shi, Recasens, Kovsharov, Borgeaud, Dery, Vasanth, Gribovskaya,
  Qiu, Mahdieh, Skut, Nielsen, Zheng, Yu, Bostock, Gupta, Archer, Rawles,
  Davies, Svyatkovskiy, Tsai, Halpern, Reisswig, Wydrowski, Chang, Puigcerver,
  Taege, Li, Schnider, Li, Dena, Xu, Telang, Shi, Zen, Kastner, Ko,
  Subramaniam, Kumar, Blois, Dai, Wieting, Lu, Zeldes, Xie, Hauth, Ţifrea, Li,
  El-Husseini, Abolafia, Zhou, Ding, Ghalebikesabi, Guía, Maksai, Ágoston
  Weisz, Arik, Sukhanov, Świetlik, Jia, Yu, Wang, Brand, Bloxwich, Kirmani,
  Chen, Go, Sprechmann, Kannen, Carin, Sandhu, Edkins, Nooteboom, Gupta,
  Maggiore, Azizi, Pritch, Yin, Gupta, Tarlow, Smith, Ivanov, Babaeizadeh,
  Goel, Kambala, Chu, Kastelic, Liu, Soltau, Stone, Agrawal, Kim, Soparkar,
  Tadepalli, Bunyan, Soh, Kannan, Kim, Chen, Halumi, Roy, Wang, Sercinoglu,
  Gibson, Bhatnagar, Sano, von Dincklage, Ren, Mitrevski, Olšák, She,
  Doersch, Jilei, Wang, Liu, Tan, Yakar, Warkentin, Ramirez, Lebsack, Dillon,
  Mathews, Cobley, Wu, Chen, Simon, Nath, Sainath, Bendebury, Julian,
  Mankalale, Ćurko, Zacchello, Brown, Sodhia, Howard, Caelles, Gupta, Evans,
  Bulanova, Katzen, Goldenberg, Tsitsulin, Stanton, Schillings, Kovalev, Fry,
  Shah, Lin, Upadhyay, Li, Radpour, Maggioni, Xiong, Haas, Brennan, Kamath,
  Savinov, Nagrani, Yacovone, Kappedal, Andriopoulos, Lao, Li, Rozhdestvenskiy,
  Hashimoto, Audibert, Austin, Rodriguez, Ruoss, Honke, Karkhanis, Xiong, Wei,
  Huang, Leng, Premachandran, Bileschi, Evangelopoulos, Mensink, Pavagadhi,
  Teplyashin, Chang, Xue, Tanzer, Goldman, Patel, Li, Wiesner, Zheng,
  Stewart-Binks, Han, Li, Luo, Lenc, Lučić, Xue, Mullins, Guseynov, Chang,
  Galatzer-Levy, Zhang, Bingham, Hu, Hartman, Ma, Griffith, Irpan, Radebaugh,
  Yue, Fan, Ungureanu, Sorokin, Teufel, Li, Anil, Paparas, Wang, Lin, Peng,
  Shum, Petrovic, Brady, Nguyen, Macherey, Li, Singh, Yenugula, Iinuma, Chen,
  Kopparapu, Stern, Dave, Thekkath, Perot, Kumar, Li, Xiao, Bilotti, Bateni,
  Noble, Lee, Vázquez-Reina, Salazar, Yang, Wang, Gruzewska, Rao, Raghuram,
  Xu, Ben-David, Mei, Dalmia, Zhang, Liu, Bansal, Pankov, Schwarcz, Burns,
  Chan, Sanghai, Liang, Liang, He, Stuart, Narayanan, Zhu, Frank, Fatemi,
  Sabne, Lang, Bhattacharya, Settle, Wang, McMahan, Tacchetti, Soares, Hadian,
  Cabi, Chung, Putikhin, Li, Chen, Tarango, Michalewski, Kazemi, Masoom,
  Sheftel, Shivanna, Vadali, Comanescu, Reid, Moore, Neelakantan, Sander,
  Herzig, Rosenberg, Dehghani, Choi, Fink, Hayes, Ge, Weng, Ho, Karro, Krishna,
  Thiet, Skerry-Ryan, Eppens, Andreetto, Sarma, Bonacina, Ayan, Nawhal, Shan,
  Dusenberry, Thakoor, Gubbi, Nguyen, Tsarfaty, Albanie, Mitrović, Gandhi,
  Chen, Epasto, Stephanov, Jin, Gehman, Amini, Weber, Behbahani, Xu, Allamanis,
  Chen, Ott, Sha, Jastrzebski, Qi, Greene, Wu, Toki, Vlasic, Shapiro,
  Kotikalapudi, Shen, Saeki, Xie, Cassirer, Bharadwaj, Kiyono, Bhojanapalli,
  Rosenfeld, Ritter, Mao, Oliveira, Egyed, Bandemer, Parisotto, Kinoshita,
  Pluto, Maniatis, Li, Guo, Ghiasi, Tarbouriech, Chatterjee, Jin, Katrina, Xu,
  Palomaki, Arnold, Sewak, Piccinini, Sharma, Albrecht, Purser-haskell,
  Vaswani, Chen, Wisniewski, Cao, Aslanides, Phu, Sieb, Agubuzu, Zheng, Sohn,
  Selvi, Andreassen, Subudhi, Eruvbetine, Woodman, Mery, Krause, Ren, Ma, Luo,
  Chen, Fan, Griffiths, Schuler, Li, Zhang, Sarr, Luo, Patana, Watson,
  Naboulsi, Collins, Sidhwani, Hoogeboom, Silver, Caveness, Zhao, Rodriguez,
  Deines, Bai, Griffin, Tagliasacchi, Xue, Babbula, Pang, Ding, Shen, Peake,
  Crocker, Raghvendra, Swisher, Han, Singh, Wu, Pchelin, Munkhdalai, Alon,
  Bacon, Robles, Bulian, Johnson, Powell, Ferreira, Li, Benzing, Velimirović,
  Soyer, Kong, Tony, Nguyên, Yang, Liu, van Amersfoort, Gillick, Sun,
  Rauschmayr, Zhang, Zhan, Zhou, Frolov, Yang, Vnukov, Rouillard, Li, Mandhane,
  Fallen, Venkataraman, Hu, Brennan, Lee, Chang, Sundermeyer, Pan, Ke, Tong,
  Fabrikant, Bono, Gu, Foley, Mao, Delakis, Bhaswar, Frostig, Li, Zipori, Hope,
  Kozlova, Mishra, Djolonga, Schiff, Merey, Briakou, Morgan, Wan, Hassidim,
  Skerry-Ryan, Sengupta, Jasarevic, Kallakuri, Kunkle, Brennan, Lieber,
  Mansoor, Walker, Zhang, Xie, Žužić, Chukwuka, Druinsky, Cho, Yao, Naeem,
  Butt, Kim, Jia, Jordan, Lelkes, Kurzeja, Wang, Zhao, Over, Chakladar,
  Prasetya, Jha, Ganapathy, Cong, Shroff, Saroufim, Miryoosefi, Hammad, Nasir,
  Xi, Gao, Maeng, Hora, Cheng, Haghani, Lewenberg, Lu, Matysiak, Raisinghani,
  Wang, Baugher, Sukthankar, Giang, Schultz, Fiedel, Chen, Lee, Dey, Zheng,
  Paul, Smith, Ly, Wang, Bansal, Perz, Ricco, Blank, Keshava, Sharma, Chow,
  Lad, Jalan, Osindero, Swanson, Scott, Ilić, Li, Jonnalagadda, Soudagar,
  Xiong, Batsaikhan, Jarrett, Kumar, Shah, Lawlor, Waters, Graham, May, Ramos,
  Lefdal, Cankara, Cano, O'Donoghue, Borovik, Liu, Grimstad, Alnahlawi,
  Tsihlas, Hudson, Grigorev, Jia, Huang, Igwe, Lebedev, Tang, Krivokon, Garcia,
  Tan, Jia, Stys, Vashishth, Liang, Venkatraman, Gu, Kementsietsidis, Zhu,
  Jung, Bai, Hosseini, Ahmed, Gupta, Yuan, Ashraf, Nigam, Vasudevan, Awasthi,
  Gilady, Mariet, Eskander, Li, Hu, Garrido, Schlattner, Zhang, Saxena, Dević,
  Muralidharan, Murthy, Zhou, Choi, Wongpanich, Wang, Shah, Xu, Huang, Spencer,
  Chen, Cohan, Wang, Tompson, Wu, Haroun, Li, Huergo, Yang, Yin, Wendt,
  Bendersky, Chaabouni, Snaider, Ferret, Jindal, Thompson, Xue, Bishop, Phal,
  Sharma, Sung, Radhakrishnan, Shomrat, Ingle, Vij, Gilmer, Istin, Sobell, Lu,
  Nottage, Sadigh, Willcock, Zhang, Xu, Brown, Lee, Wang, Zhu, Tay, Kim,
  Gutierrez, Sharma, Xian, Seo, Cui, Pochernina, Baetu, Jastrzębski, Ly,
  Elhawaty, Suh, Sezener, Wang, Yuen, Tucker, Cai, Yang, Wang, Muzio, Qian,
  Yoo, Lockhart, McKee, Guo, Mehrotra, Mendonça, Mehta, Ben, Tekur, Mu, Zhu,
  Krakovna, Lee, Maschinot, Cevey, Choe, Bai, Srinivasan, Gasaway, Young,
  Siegler, Holtmann-Rice, Piratla, Baumli, Yogev, Hofer, van Hasselt, Grant,
  Chervonyi, Silver, Hogue, Agarwal, Wang, Singh, Flynn, Lipschultz, David,
  Bellot, Yang, Le, Graziano, Olszewska, Hui, Maurya, Parotsidis, Chen,
  Oguntebi, Kelley, Baddepudi, Mauerer, Shaw, Siegman, Yang, Shetty, Roy, Song,
  Stokowiec, Burnell, Savant, Busa-Fekete, Miao, Ghosh, MacDermed, Lippe,
  Dektiarev, Behrman, Mentzer, Nguyen, Wei, Verma, Knutsen, Dasari, Yan,
  Mitrichev, Wang, Shejwalkar, Austin, Sunkara, Potti, Virin, Wright, Liu,
  Riva, Pot, Kochanski, Le, Balasubramaniam, Dhar, Liao, Bloniarz, Shukla,
  Cole, Lee, Zhang, Kafle, Vashishtha, Mahmoudieh, Chen, Hoffmann, Srinivasan,
  Lago, Shalom, Wang, Elabd, Sharma, Oh, Kothawade, Le, Monteiro, Yang,
  Alarakyia, Geirhos, Mincu, Garnes, Kobayashi, Mariooryad, Krasowiak, Zhixin,
  Lai, Mourad, Wang, Bu, Aharoni, Chen, Goyal, Zubov, Bapna, Dabir, Kothari,
  Lamerigts, Cao, Shar, Yew, Kulkarni, Mahaarachchi, Joshi, Zhu, Lichtarge,
  Zhou, Muckenhirn, Selo, Vinyals, Chen, Brohan, Mehta, Cogan, Wang, Geri, Ko,
  Chen, Viola, Shivam, Wang, Elish, Popa, Pereira, Liu, Koster, Kim, Zhang,
  Ebrahimi, Talukdar, Zheng, Poklukar, Mikhalap, Johnson, Vijayakumar,
  Omernick, Dibb, Dubey, Hu, Suman, Aggarwal, Kornakov, Xia, Lowe, Kolganov,
  Xiao, Nikolaev, Hemingray, Li, Iljazi, Rybiński, Sandhu, Lu, Luong,
  Jenatton, Govindaraj, Hui, Li, Dulac-Arnold, Park, Wang, Modi, Pouget-Abadie,
  Greller, Gupta, Berry, Ramachandran, Xie, McCafferty, Wang, Gupta, Lim,
  Bratanič, Brock, Akolzin, Sproch, Karliner, Kim, Goedeckemeyer, Shazeer,
  Schmid, Calandriello, Bhatia, Choromanski, Montgomery, Dua, Ramalho, King,
  Gao, Nguyen, Lindner, Pitta, Johnson, Salama, Ardila, Han, Farnese, Odoom,
  Wang, Ding, Rink, Smith, Lehri, Cohen, Vats, He, Gopavarapu, Paszke, Patel,
  Gansbeke, Loher, Castro, Voitovich, von Glehn, George, Niklaus, Eaton-Rosen,
  Rakićević, Jue, Perel, Zhang, Bahat, Pouget, Xing, Huot, Shenoy, Bos,
  Coriou, Richter, Noy, Wang, Ontanon, Qin, Makarchuk, Hassabis, Li, Sharma,
  Venkatesan, Kemaev, Daniel, Huang, Shah, Ponce, Warren, Chen, Faruqui, Wu,
  Andačić, Payrits, McDuff, Hume, Cao, Tessler, Wang, Wang, Rendulic,
  Agustsson, Johnson, Lando, Howard, Padmanabhan, Daswani, Banino, Kilgore,
  Heek, Ji, Caceres, Li, Kassner, Vlaskin, Liu, Grills, Hou, Sukkerd, Cheon,
  Shetty, Markeeva, Stanczyk, Iyer, Gong, Gao, Gopalakrishnan, Blyth, Reynolds,
  Bhoopchand, Bilenko, Gharibian, Zayats, Faust, Singh, Ma, Jiao,
  Vijayanarasimhan, Aroyo, Yadav, Chakera, Kakarla, Meshram, Gregor, Botea,
  Senter, Jia, Kovacs, Sharma, Baur, Kang, He, Zhuo, Kostelac, Laish, Peng,
  O'Bryan, Kasenberg, Rao, Leurent, Zhang, Stevens, Salazar, Zhang, Lobov,
  Walker, Porter, Redshaw, Ke, Rao, Lee, Lam, Moffitt, Kim, Qiao, Koo, Dadashi,
  Song, Sundararajan, Xu, Kawamoto, Zhong, Barbu, Reddy, Verzetti, Li,
  Papamakarios, Klimczak-Plucińska, Cassin, Kavukcuoglu, Swavely, Vaucher,
  Zhao, Hemsley, Tschannen, Ge, Menghani, Yu, Ha, He, Wu, Song, Sterneck,
  Zinke, Calian, Marsden, Ruiz, Hessel, Gueta, Lee, Farris, Gupta, Li, Saleh,
  Misra, Xiao, Mendolicchio, Buttimore, Krayvanova, Nayakanti, Wiethoff, Pande,
  Mirhoseini, Lao, Liu, Hua, Chen, Malkov, Kalashnikov, Gupta, Audhkhasi, Zhai,
  Kopalle, Jain, Ofek, Meyer, Baatarsukh, Strejček, Qian, Freedman, Figueira,
  Sokolik, Bachem, Lin, Kharrat, Hidey, Xu, Duan, Li, Ersoy, Everett, Cen,
  Santamaria-Fernandez, Taubenfeld, Mackinnon, Deng, Zablotskaia, Viswanadha,
  Goel, Yates, Deng, Choy, Chen, Sinha, Mossin, Wang, Szlam, Hao, Rubenstein,
  Toksoz-Exley, Aperghis, Zhong, Ahn, Isard, Lacombe, Luisier, Anastasiou,
  Kalley, Prabhu, Dunleavy, Bijwadia, Mao-Jones, Chen, Pasumarthi, Wood,
  Dostmohamed, Hurley, Simsa, Parrish, Pajarskas, Harvey, Skopek, Kochinski,
  Rey, Rieser, Zhou, Lee, Acharya, Li, Jiang, Zhang, Gipson, Mahintorabi,
  Gelmi, Khajehnouri, Yeh, Lee, Matthey, Baker, Pham, Fu, Pak, Gupta,
  Vasconcelos, Sadovsky, Walker, Hsiao, Zochbauer, Marzoca, Velan, Zeng,
  Baechler, Driess, Jain, Huang, Tao, Maggs, Levine, Schneider, Gemzer, Petit,
  Han, Fisher, Zelle, Biles, Ie, Fadeeva, Liu, Franco, Collister, Zhang, Wang,
  Zhao, Kieliger, Shuster, Zhu, Gong, Chan, Sun, Basu, Zimmermann, Hayes,
  Bapna, Snoek, Yang, Datta, Abdallah, Kilgour, Li, Mah, Jun, Rivière,
  Karmarkar, Spalink, Huang, Gonzalez, Tran, Nowak, Palowitch, Chadwick,
  Talius, Mehta, Sellam, Fränken, Nicosia, He, Kini, Amos, Basu, Jobe, Shaw,
  Xu, Evans, Ikeda, Yan, Jin, Wang, Yadav, Labzovsky, Sampath, Ma, Schumann,
  Siddhant, Shah, Youssef, Agarwal, Dabney, Tonioni, Ambar, Li, Guyon, Li,
  Soergel, Fang, Karadzhov, Udrescu, Trinh, Raunak, Noury, Guo, Gupta,
  Finkelstein, Petek, Liang, Billock, Sun, Wood, Song, Yu, Matejovicova, Cohen,
  Andra, D'Ambrosio, Deng, Nallatamby, Songhori, Dangovski, Lampinen, Botadra,
  Hillier, Cao, Baddi, Kuncoro, Yoshino, Bhagatwala, Ranzato, Schaeffer, Liu,
  Ye, Sarvana, Nham, Kuang, Gao, Baek, Mittal, Wahid, Gergely, Ni, Feldman,
  Muir, Lamblin, Macherey, Dyer, Kilpatrick, Campos, Bhutani, Fort, Ahmad,
  Severyn, Chatziprimou, Ferludin, Dimarco, Kusupati, Heyward, Bahir, Villela,
  Millican, Marcus, Bahargam, Unlu, Roth, Wei, Gopal, Ghoshal, Lee, Lin, Lees,
  Lee, Hosseini, Fan, Neel, Wu, Altun, Cai, Piqueras, Woodward, Bissacco,
  Haykal, Bordbar, Sundaram, Hodkinson, Toyama, Polovets, Myers, Sinha,
  Levinboim, Krishnakumar, Chhaparia, Sholokhova, Gundavarapu, Jawahar,
  Qureshi, Hu, Momchev, Rahtz, Wu, S, Dhamdhere, Guo, Gupta, Eslami, Schain,
  Blokzijl, Welling, Orr, Bolelli, Perez-Nieves, Sirotenko, Prasad, Kar, Pigem,
  Terzi, Weisz, Ghosh, Mavalankar, Madeka, Daugaard, Adam, Shah, Berman, Tran,
  Baker, Andrejczuk, Chole, Raboshchuk, Mirzazadeh, Kagohara, Wu, Schallhart,
  Orlando, Wang, Rrustemi, Xiong, Liu, Vezer, Ramsden, yiin Chang, Mudgal, Li,
  Vieillard, Hoshen, Ahmad, Slone, Hua, Potikha, Rossini, Stritar, Prakash,
  Wang, Dong, Nazari, Nehoran, Tekelioglu, Li, Badola, Funkhouser, Li, Yerram,
  Ganeshan, Formoso, Langner, Shi, Li, Yamamori, Panda, Saade, Scarpati,
  Breaux, Carey, Zhou, Hsieh, Bridgers, Butryna, Gupta, Tulsyan, Woo, Eltyshev,
  Grathwohl, Parks, Benjamin, Panigrahy, Dodhia, Freitas, Sauer, Song, Alet,
  Tolins, Paduraru, Zhou, Albert, Zhang, Shu, Bansal, Nguyen, Globerson, Xiao,
  Manyika, Hennigan, Rong, Matak, Bakalov, Sharma, Sinopalnikov, Pierson,
  Roller, Brown, Gao, Fukuzawa, Ghafouri, Vassigh, Barr, Wang, Korsun, Jayaram,
  Ren, Zaman, Khan, Lunts, Deutsch, Uthus, Katz, Samsikova, Khalifa, Sethi,
  Sun, Tang, Alon, Luo, Yu, Nayyar, Petrini, Truong, Hellendoorn, Chinaev,
  Alberti, Wang, Hu, Mirrokni, Balashankar, Aharon, Mehta, Iscen, Kready,
  Manning, Mohananey, Chen, Tripathi, Wu, Petrovski, Hwang, Baeuml,
  Chandrakaladharan, Liu, Coaguila, Chen, Ma, Tafti, Tatineni, Spitz, Ye,
  Vicol, Rosca, Puigdomènech, Yahav, Ghemawat, Lin, Kirk, Nabulsi, Brin,
  Bohnet, Caluwaerts, Veerubhotla, Zheng, Dai, Petrov, Xu, Mehran, Xu,
  Zintgraf, Choi, Hombaiah, Thoppilan, Reddi, Lew, Li, Webster, Sawhney,
  Lamprou, Shakeri, Lunayach, Chen, Bagri, Salcianu, Chen, Donchev, Magister,
  Nørly, Rodrigues, Izo, Noga, Zou, Köppe, Zhou, Lee, Long, Eisenbud, Chen,
  Schenck, To, Zhong, Taropa, Truong, Levy, Martins, Zhang, Semturs, Zhang,
  Yakubovich, Moreno, McConnaughey, Lu, Redmond, Weerts, Bitton, Refice,
  Lacasse, Conmy, Tallec, Odell, Forbes-Pollard, Socala, Hoech, Kohli, Walton,
  Wang, Sazanovich, Zhu, Kapishnikov, Galt, Denton, Murdoch, Sikora, Mohamed,
  Wei, First, McConnell, Cobo, Qin, Avrahami, Balle, Watanabe, Louis, Kraft,
  Ariafar, Gu, Rives, Yoon, Rusu, Cobon-Kerr, Hahn, Luo, Yuvein, Zhu, Ahuja,
  Benenson, Kaufman, Yu, Hightower, Zhang, Ni, Hendricks, Wang, Yona, Jain,
  Barrio, Bhupatiraju, Velusamy, Dafoe, Riedel, Thomas, Yuan, Bellaiche,
  Panthaplackel, Kloboves, Jauhari, Akbulut, Davchev, Gladchenko, Madras,
  Chuklin, Hill, Yuan, Madhavan, Leonhard, Scandinaro, Chen, Niu, Douillard,
  Damoc, Onoe, Pedregosa, Bertsch, Leichner, Pagadora, Malmaud, Ponda, Twigg,
  Duzhyi, Shen, Wang, Garg, Chen, Evci, Lee, Liu, Kojima, Yamaguchi, Rajendran,
  Piergiovanni, Rajendran, Fornoni, Ibagon, Ragan, Khan, Blitzer, Bunner, Sun,
  Kosakai, Lundberg, Elue, Guu, Park, Park, Narayanaswamy, Wu, Mudigonda, Cohn,
  Mu, Kumar, Graesser, Zhang, Killam, Zhuang, Giménez, Jishi, Ley-Wild, Zhai,
  Osawa, Cedillo, Liu, Upadhyay, Sieniek, Sharma, Paine, Angelova, Addepalli,
  Parada, Majumder, Lamp, Kumar, Deng, Myaskovsky, Sabolić, Dudek, York,
  de~Chaumont~Quitry, Nie, Cattle, Gunjan, Piot, Khawaja, Bang, Wang,
  Khodadadeh, R, Rawlani, Powell, Lee, Griesser, Oh, Magalhaes, Li, Tokumine,
  Vogel, Hsu, BC, Jindal, Cohen, Yang, Yuan, de~Cesare, Bruguier, Xu, Roy,
  Jacovi, Belov, Arya, Meadowlark, Cohen-Ganor, Ye, Morris-Suzuki, Banzal,
  Song, Ponnuramu, Zhang, Scrivener, Zaiem, Rochman, Han, Ghazi, Lee, Drath,
  Suo, Girgis, Shenoy, Nguyen, Eck, Gupta, Yan, Carreira, Gulati, Sang,
  Mirylenka, Cooney, Chou, Ling, Fan, Coleman, Tubone, Kumar, Baldridge,
  Hernandez-Campos, Lazaridou, Besley, Yona, Bulut, Wellens, Pierigiovanni,
  George, Green, Han, Tao, Clark, You, Abdolmaleki, Fu, Chen, Chaugule,
  Chandorkar, Rahman, Thompson, Koanantakool, Bernico, Ren, Vlasov,
  Vassilvitskii, Kula, Liang, Kim, Huang, Ye, Lepikhin, and
  Helmholz]{comanici2025gemini25pushingfrontier}
Gheorghe Comanici, Eric Bieber, Mike Schaekermann, Ice Pasupat, Noveen
  Sachdeva, Inderjit Dhillon, Marcel Blistein, Ori Ram, Dan Zhang, Evan Rosen
  et~al.
\newblock Gemini 2.5: Pushing the Frontier with Advanced Reasoning,
  Multimodality, Long Context, and Next Generation Agentic Capabilities, 2025.
\newblock URL \url{https://arxiv.org/abs/2507.06261}.

\bibitem[DeepSeek-AI(2024)]{deepseekai2024deepseekv3technicalreport}
DeepSeek-AI.
\newblock DeepSeek-V3 Technical Report, 2024.
\newblock URL \url{https://arxiv.org/abs/2412.19437}.

\bibitem[DeepSeek-AI et~al.(2025)DeepSeek-AI, Liu, Mei, Lin, Xue, Wang, Xu, Wu,
  Zhang, Lin, Dong, Lu, Zhao, Deng, Xu, Ruan, Dai, Guo, Yang, Chen, Li, Zhou,
  Lin, Dai, Hao, Chen, Li, Zhang, Xu, Li, Liang, Wei, Zhang, Luo, Ji, Ding,
  Tang, Cao, Gao, Qu, Zeng, Huang, Li, Xu, Hu, Chen, Xiang, Yuan, Cheng, Zhu,
  Ran, Jiang, Qiu, Li, Song, Dong, Gao, Guan, Huang, Zhou, Huang, Yu, Wang,
  Zhang, Wang, Zhao, Yin, Guo, Luo, Ma, Wang, Zhang, Di, Xu, Zhang, Zhang,
  Tang, Zhou, Huang, Cong, Wang, Wang, Zhu, Li, Chen, Du, Xu, Ge, Zhang, Pan,
  Wang, Yin, Xu, Shen, Zhang, Liu, Lu, Zhou, Chen, Cai, Chen, Hu, Liu, Hu, Ma,
  Wang, Yu, Zhou, Pan, Zhou, Ni, Yun, Pei, Ye, Yue, Zeng, Liu, Liang, Pang,
  Luo, Gao, Zhang, Gao, Wang, Bi, Liu, Wang, Chen, Zhang, Nie, Cheng, Liu, Xie,
  Liu, Yu, Li, Yang, Li, Chen, Su, Pan, Lin, Fu, Wang, Zhang, Xu, Ma, Li, Li,
  Zhao, Sun, Wang, Qian, Yu, Zhang, Ding, Shi, Xiong, He, Zhou, Zhong, Piao,
  Wang, Chen, Tan, Wei, Ma, Liu, Yang, Guo, Wu, Wu, Cheng, Ou, Xu, Wang, Gong,
  Wu, Zou, Li, Xiong, Luo, You, Liu, Zhou, Wu, Ren, Zhao, Ren, Sha, Fu, Xu,
  Xie, Zhang, Hao, Gou, Ma, Yan, Shao, Huang, Wu, Li, Zhang, Xu, Wang, Gu, Zhu,
  Li, Zhang, Xie, Gao, Pan, Yao, Feng, Li, Cai, Ni, Xu, Li, Tian, Chen, Jin,
  Li, Zhou, Sun, Li, Jin, Shen, Chen, Song, Zhou, Zhu, Huang, Li, Zheng, Zhu,
  Ma, Huang, Xu, Zhang, Ji, Liang, Guo, Chen, Xia, Wang, Li, Zhang, Chen, Sun,
  Wu, Ye, Wang, Xiao, An, Wang, Sun, Wang, Tang, Zha, Zhang, Ju, Zhang, and
  Qu]{deepseekai2025deepseekv32pushingfrontieropen}
DeepSeek-AI, Aixin Liu, Aoxue Mei, Bangcai Lin, Bing Xue, Bingxuan Wang,
  Bingzheng Xu, Bochao Wu, Bowei Zhang, Chaofan Lin et~al.
\newblock DeepSeek-V3.2: Pushing the Frontier of Open Large Language Models,
  2025.
\newblock URL \url{https://arxiv.org/abs/2512.02556}.

\bibitem[DeepSeek-AI et~al.(2026)DeepSeek-AI, Xu, Lin, Xue, Wang, Xu, Wu,
  Zhang, Lin, Dong, Ling, Lu, Zhao, Deng, Hou, Xu, Shao, Ruan, Sun, Dai, Guo,
  Yang, Chen, Li, Ji, Li, Wei, Lin, Yuan, Xia, Dai, Hao, Chen, Cao, Meng, Li,
  Yu, Zhang, Xu, Li, Liang, Zhang, Luo, Wei, Yuan, Zhang, Luo, Chen, Ji, Zhang,
  Ding, Tang, Cao, Gao, Qu, Zeng, Yang, Zhu, Luo, Song, Yu, Huang, Cai, Liang,
  Zhou, Ye, Li, Xu, Hu, Yang, Chen, Yan, Chen, Zhou, Xiang, Yuan, Cheng, Zhou,
  Zhu, Yu, Sun, Ran, Jiang, Qiu, Li, Zheng, Song, Dong, Gao, Guan, Zhou, Huang,
  Yu, Wang, Zhang, Wang, Xia, Zhang, Zhao, Guo, Luo, Ma, Zhu, Wang, Cai, Zhang,
  Chen, Di, Xu, Mei, Wang, Zhang, Zhang, Tang, Li, Zhou, Han, Wang, Huang,
  Wang, Cong, Wang, Zhang, Wang, Zhu, Li, Chen, Du, Jiang, Tian, Xu, Lu, Xu,
  Ge, Zhang, Pan, Wang, Chen, Yin, Xu, Shen, Zhang, Chen, Liu, Lu, Sun, Zhou,
  Chen, Cai, Nie, Wu, Chen, Hu, Liu, Hu, Ma, Wang, Yu, Zhou, Pan, Yu, Zhou, Ni,
  Yun, Jin, Pei, Ye, Lin, Ji, Cui, Yue, Yu, Wang, Zhang, Xiao, Zeng, An, Zhao,
  Liu, Liang, Pang, Luo, Yao, Gao, Yang, Huang, Hou, Zhang, Ma, Gao, He, Wang,
  Wang, Bi, Liu, Wang, Chen, Zhang, Nie, Sun, Wang, Cheng, Liu, Xie, Liu, Liu,
  Yu, Li, Yang, Zhang, Chen, Wang, Su, Chen, Lin, Fu, Yan, Wang, Ma, Luo,
  Zhang, Xu, Ma, Huang, Li, Li, Xu, Zhao, Sun, Wang, Qian, Shao, Yu, Zhang,
  Ding, Shi, Wu, Xiong, Ma, He, Tang, Zhou, Luo, Zhong, Piao, Wang, Zhang,
  Chen, Tan, Wei, Ma, Liu, Yang, Guo, Wu, Wu, Li, Cheng, Ou, Xu, Li, Wang,
  Yang, Xu, Wu, Meng, Zou, Zha, Xiong, Chen, Lin, Cao, Wang, Zhang, Yan, Lin,
  Gu, Luo, You, Liu, Zhou, Zhou, Huang, Wu, Wang, Zhao, Ren, Zhang, Sha, Fu,
  Ju, Xu, Xie, Zhang, Gao, Hao, Gou, Ma, Yan, Shao, Huang, Chen, Wu, Ren, Wu,
  Li, Zhang, Xu, Wang, Qu, Gu, Zhu, Li, Zhang, Xie, Gao, Wan, Pan, and
  Yao]{deepseekai2026deepseekv4highlyefficientmilliontoken}
DeepSeek-AI, Anyi Xu, Bangcai Lin, Bing Xue, Bingxuan Wang, Bingzheng Xu,
  Bochao Wu, Bowei Zhang, Chaofan Lin, Chen Dong et~al.
\newblock DeepSeek-V4: Towards Highly Efficient Million-Token Context
  Intelligence, 2026.
\newblock URL \url{https://arxiv.org/abs/2606.19348}.

\bibitem[Duarte et~al.(2024)Duarte, Marques, Gra{\c{c}}a, Freire, Li, and
  Oliveira]{duarte-etal-2024-lumberchunker}
Andr{\'e}~V. Duarte, Jo{\~a}o~DS Marques, Miguel Gra{\c{c}}a, Miguel Freire,
  Lei Li, and Arlindo~L. Oliveira.
\newblock {L}umber{C}hunker: Long-Form Narrative Document Segmentation.
\newblock In Yaser Al-Onaizan, Mohit Bansal, and Yun-Nung Chen (eds.),
  \emph{Findings of the Association for Computational Linguistics: EMNLP 2024},
  pp.\  6473--6486, Miami, Florida, USA, November 2024. Association for
  Computational Linguistics.
\newblock \doi{10.18653/v1/2024.findings-emnlp.377}.
\newblock URL \url{https://aclanthology.org/2024.findings-emnlp.377/}.

\bibitem[Duarte et~al.(2026)Duarte, Tufts, Oke, Fang, Oliveira, and
  Li]{duarte2026semdetect}
Andre~Vicente Duarte, Brian Tufts, Aditya Oke, Fei Fang, Arlindo~L. Oliveira,
  and Lei Li.
\newblock Sem-Detect: Semantic Level Detection of {AI} Generated Peer-Reviews.
\newblock In \emph{Forty-third International Conference on Machine Learning},
  2026.
\newblock URL \url{https://openreview.net/forum?id=3S9eVockjL}.

\bibitem[Dugan et~al.(2024)Dugan, Hwang, Trhl{\'i}k, Zhu, Ludan, Xu, Ippolito,
  and Callison-Burch]{dugan-etal-2024-raid}
Liam Dugan, Alyssa Hwang, Filip Trhl{\'i}k, Andrew Zhu, Josh~Magnus Ludan,
  Hainiu Xu, Daphne Ippolito, and Chris Callison-Burch.
\newblock {RAID}: A Shared Benchmark for Robust Evaluation of Machine-Generated
  Text Detectors.
\newblock In Lun-Wei Ku, Andre Martins, and Vivek Srikumar (eds.),
  \emph{Proceedings of the 62nd Annual Meeting of the Association for
  Computational Linguistics (Volume 1: Long Papers)}, pp.\  12463--12492,
  Bangkok, Thailand, August 2024. Association for Computational Linguistics.
\newblock \doi{10.18653/v1/2024.acl-long.674}.
\newblock URL \url{https://aclanthology.org/2024.acl-long.674/}.

\bibitem[GLM-5-Team et~al.(2026)GLM-5-Team, Zeng, Lv, Hou, Du, Zheng, Chen,
  Yin, Ge, Huang, Xie, Zhu, Yin, Wang, Pan, Zeng, Zhang, Wang, Chen, Zhang,
  Jiao, Guo, Wang, Du, Wu, Wang, Li, Fan, Zhong, Liu, Zhao, Du, Dong, Lu,
  Shuang-Li, Cao, Liu, Jiang, Chen, Zhang, Huang, Dong, Xu, Wei, An, Niu, Zhu,
  Wen, Cen, Bai, Qiao, Wang, Wang, Zhu, Liu, Li, Wang, Wen, Huang, Cai, Yu, Li,
  Hu, Zhang, Zhang, Lin, Yang, Wang, Ai, Zhu, Yi, Chen, Wen, Sun, Zhao, Hu,
  Zhang, Liu, Zhang, Peng, Tai, Zhang, Liu, Wang, Yan, Ge, Liu, Chu, Zhao,
  Wang, Zhao, Ren, Wang, Zhang, Gui, Zhao, Li, An, Li, Yuan, Du, Liu, Zhi,
  Duan, Zhou, Wei, Wang, Luo, Zhang, Sha, Xu, Wu, Ding, Chen, Li, Lin, Ta, Zou,
  Song, Yang, Tu, Yang, Wu, Zhang, Li, Li, Fan, Qin, Tian, Zhang, Yu, Liang,
  Kuang, Cheng, Li, Yan, Hu, Ling, Fan, Xia, Zhang, Zhang, Pan, Zou, Zhang,
  Liu, Wu, Li, Wang, Zhu, Tan, Zhou, Pan, Zhang, Su, Geng, Yan, Tan, Bi, Shen,
  Yang, Li, Liu, Wang, Li, Wu, Zhang, Duan, Zhang, Liu, Jiang, Yan, Zhang, Wei,
  Chen, Feng, Yao, Chai, Wang, Zhang, Xu, Huang, Wang, Li, Dong, and
  Tang]{glm5team2026glm5vibecodingagentic}
GLM-5-Team, Aohan Zeng, Xin Lv, Zhenyu Hou, Zhengxiao Du, Qinkai Zheng, Bin
  Chen, Da~Yin, Chendi Ge, Chenghua Huang et~al.
\newblock GLM-5: from Vibe Coding to Agentic Engineering, 2026.
\newblock URL \url{https://arxiv.org/abs/2602.15763}.

\bibitem[Hans et~al.(2024)Hans, Schwarzschild, Cherepanova, Kazemi, Saha,
  Goldblum, Geiping, and Goldstein]{pmlr-v235-hans24a}
Abhimanyu Hans, Avi Schwarzschild, Valeriia Cherepanova, Hamid Kazemi,
  Aniruddha Saha, Micah Goldblum, Jonas Geiping, and Tom Goldstein.
\newblock Spotting {LLM}s With Binoculars: Zero-Shot Detection of
  Machine-Generated Text.
\newblock In Ruslan Salakhutdinov, Zico Kolter, Katherine Heller, Adrian
  Weller, Nuria Oliver, Jonathan Scarlett, and Felix Berkenkamp (eds.),
  \emph{Proceedings of the 41st International Conference on Machine Learning},
  volume 235 of \emph{Proceedings of Machine Learning Research}, pp.\
  17519--17537. PMLR, 21--27 Jul 2024.
\newblock URL \url{https://proceedings.mlr.press/v235/hans24a.html}.

\bibitem[Hu et~al.(2023)Hu, Chen, and Ho]{NEURIPS2023_30e15e59}
Xiaomeng Hu, Pin-Yu Chen, and Tsung-Yi Ho.
\newblock RADAR: Robust AI-Text Detection via Adversarial Learning.
\newblock In A.~Oh, T.~Naumann, A.~Globerson, K.~Saenko, M.~Hardt, and
  S.~Levine (eds.), \emph{Advances in Neural Information Processing Systems},
  volume~36, pp.\  15077--15095. Curran Associates, Inc., 2023.
\newblock \doi{10.52202/075280-0662}.
\newblock URL
  \url{https://proceedings.neurips.cc/paper_files/paper/2023/file/30e15e5941ae0cdab7ef58cc8d59a4ca-Paper-Conference.pdf}.

\bibitem[Ippolito et~al.(2020)Ippolito, Duckworth, Callison-Burch, and
  Eck]{ippolito-etal-2020-automatic}
Daphne Ippolito, Daniel Duckworth, Chris Callison-Burch, and Douglas Eck.
\newblock Automatic Detection of Generated Text is Easiest when Humans are
  Fooled.
\newblock In Dan Jurafsky, Joyce Chai, Natalie Schluter, and Joel Tetreault
  (eds.), \emph{Proceedings of the 58th Annual Meeting of the Association for
  Computational Linguistics}, pp.\  1808--1822, Online, July 2020. Association
  for Computational Linguistics.
\newblock \doi{10.18653/v1/2020.acl-main.164}.
\newblock URL \url{https://aclanthology.org/2020.acl-main.164/}.

\bibitem[Jiang et~al.(2023)Jiang, Sablayrolles, Mensch, Bamford, Chaplot,
  de~las Casas, Bressand, Lengyel, Lample, Saulnier, Lavaud, Lachaux, Stock,
  Scao, Lavril, Wang, Lacroix, and Sayed]{jiang2023mistral7b}
Albert~Q. Jiang, Alexandre Sablayrolles, Arthur Mensch, Chris Bamford,
  Devendra~Singh Chaplot, Diego de~las Casas, Florian Bressand, Gianna Lengyel,
  Guillaume Lample, Lucile Saulnier et~al.
\newblock Mistral 7B, 2023.
\newblock URL \url{https://arxiv.org/abs/2310.06825}.

\bibitem[Ke et~al.(2017)Ke, Meng, Finley, Wang, Chen, Ma, Ye, and
  Liu]{NIPS2017_6449f44a}
Guolin Ke, Qi~Meng, Thomas Finley, Taifeng Wang, Wei Chen, Weidong Ma, Qiwei
  Ye, and Tie-Yan Liu.
\newblock LightGBM: A Highly Efficient Gradient Boosting Decision Tree.
\newblock In I.~Guyon, U.~Von Luxburg, S.~Bengio, H.~Wallach, R.~Fergus,
  S.~Vishwanathan, and R.~Garnett (eds.), \emph{Advances in Neural Information
  Processing Systems}, volume~30. Curran Associates, Inc., 2017.
\newblock URL
  \url{https://proceedings.neurips.cc/paper_files/paper/2017/file/6449f44a102fde848669bdd9eb6b76fa-Paper.pdf}.

\bibitem[Kirchenbauer et~al.(2023)Kirchenbauer, Geiping, Wen, Katz, Miers, and
  Goldstein]{pmlr-v202-kirchenbauer23a}
John Kirchenbauer, Jonas Geiping, Yuxin Wen, Jonathan Katz, Ian Miers, and Tom
  Goldstein.
\newblock A Watermark for Large Language Models.
\newblock In Andreas Krause, Emma Brunskill, Kyunghyun Cho, Barbara Engelhardt,
  Sivan Sabato, and Jonathan Scarlett (eds.), \emph{Proceedings of the 40th
  International Conference on Machine Learning}, volume 202 of
  \emph{Proceedings of Machine Learning Research}, pp.\  17061--17084. PMLR,
  23--29 Jul 2023.
\newblock URL \url{https://proceedings.mlr.press/v202/kirchenbauer23a.html}.

\bibitem[Krishna et~al.(2023)Krishna, Song, Karpinska, Wieting, and
  Iyyer]{NEURIPS2023_575c4500}
Kalpesh Krishna, Yixiao Song, Marzena Karpinska, John Wieting, and Mohit Iyyer.
\newblock Paraphrasing evades detectors of AI-generated text, but retrieval is
  an effective defense.
\newblock In A.~Oh, T.~Naumann, A.~Globerson, K.~Saenko, M.~Hardt, and
  S.~Levine (eds.), \emph{Advances in Neural Information Processing Systems},
  volume~36, pp.\  27469--27500. Curran Associates, Inc., 2023.
\newblock \doi{10.52202/075280-1195}.
\newblock URL
  \url{https://proceedings.neurips.cc/paper_files/paper/2023/file/575c450013d0e99e4b0ecf82bd1afaa4-Paper-Conference.pdf}.

\bibitem[Kumar et~al.(2024)Kumar, Sahu, Gacche, Ghosal, and
  Ekbal]{kumar-etal-2024-quis}
Sandeep Kumar, Mohit Sahu, Vardhan Gacche, Tirthankar Ghosal, and Asif Ekbal.
\newblock `Quis custodiet ipsos custodes?' Who will watch the watchmen? On
  Detecting {AI}-generated peer-reviews.
\newblock In Yaser Al-Onaizan, Mohit Bansal, and Yun-Nung Chen (eds.),
  \emph{Proceedings of the 2024 Conference on Empirical Methods in Natural
  Language Processing}, pp.\  22663--22679, Miami, Florida, USA, November 2024.
  Association for Computational Linguistics.
\newblock \doi{10.18653/v1/2024.emnlp-main.1262}.
\newblock URL \url{https://aclanthology.org/2024.emnlp-main.1262/}.

\bibitem[Li et~al.(2024)Li, Li, Cui, Bi, Wang, Wang, Yang, Shi, and
  Zhang]{li-etal-2024-mage}
Yafu Li, Qintong Li, Leyang Cui, Wei Bi, Zhilin Wang, Longyue Wang, Linyi Yang,
  Shuming Shi, and Yue Zhang.
\newblock {MAGE}: Machine-generated Text Detection in the Wild.
\newblock In Lun-Wei Ku, Andre Martins, and Vivek Srikumar (eds.),
  \emph{Proceedings of the 62nd Annual Meeting of the Association for
  Computational Linguistics (Volume 1: Long Papers)}, pp.\  36--53, Bangkok,
  Thailand, August 2024. Association for Computational Linguistics.
\newblock \doi{10.18653/v1/2024.acl-long.3}.
\newblock URL \url{https://aclanthology.org/2024.acl-long.3/}.

\bibitem[Liang et~al.(2024)Liang, Izzo, Zhang, Lepp, Cao, Zhao, Chen, Ye, Liu,
  Huang, Mcfarland, and Zou]{pmlr-v235-liang24b}
Weixin Liang, Zachary Izzo, Yaohui Zhang, Haley Lepp, Hancheng Cao, Xuandong
  Zhao, Lingjiao Chen, Haotian Ye, Sheng Liu, Zhi Huang et~al.
\newblock Monitoring {AI}-Modified Content at Scale: A Case Study on the Impact
  of {C}hat{GPT} on {AI} Conference Peer Reviews.
\newblock In Ruslan Salakhutdinov, Zico Kolter, Katherine Heller, Adrian
  Weller, Nuria Oliver, Jonathan Scarlett, and Felix Berkenkamp (eds.),
  \emph{Proceedings of the 41st International Conference on Machine Learning},
  volume 235 of \emph{Proceedings of Machine Learning Research}, pp.\
  29575--29620. PMLR, 21--27 Jul 2024.
\newblock URL \url{https://proceedings.mlr.press/v235/liang24b.html}.

\bibitem[Liu et~al.(2019)Liu, Ott, Goyal, Du, Joshi, Chen, Levy, Lewis,
  Zettlemoyer, and Stoyanov]{liu2019robertarobustlyoptimizedbert}
Yinhan Liu, Myle Ott, Naman Goyal, Jingfei Du, Mandar Joshi, Danqi Chen, Omer
  Levy, Mike Lewis, Luke Zettlemoyer, and Veselin Stoyanov.
\newblock RoBERTa: A Robustly Optimized BERT Pretraining Approach, 2019.
\newblock URL \url{https://arxiv.org/abs/1907.11692}.

\bibitem[Loshchilov \& Hutter(2019)Loshchilov and
  Hutter]{DBLP:conf/iclr/LoshchilovH19}
Ilya Loshchilov and Frank Hutter.
\newblock Decoupled Weight Decay Regularization.
\newblock In \emph{7th International Conference on Learning Representations,
  {ICLR} 2019, New Orleans, LA, USA, May 6-9, 2019}. OpenReview.net, 2019.
\newblock URL \url{https://openreview.net/forum?id=Bkg6RiCqY7}.

\bibitem[Mitchell et~al.(2023)Mitchell, Lee, Khazatsky, Manning, and
  Finn]{pmlr-v202-mitchell23a}
Eric Mitchell, Yoonho Lee, Alexander Khazatsky, Christopher~D Manning, and
  Chelsea Finn.
\newblock {D}etect{GPT}: Zero-Shot Machine-Generated Text Detection using
  Probability Curvature.
\newblock In Andreas Krause, Emma Brunskill, Kyunghyun Cho, Barbara Engelhardt,
  Sivan Sabato, and Jonathan Scarlett (eds.), \emph{Proceedings of the 40th
  International Conference on Machine Learning}, volume 202 of
  \emph{Proceedings of Machine Learning Research}, pp.\  24950--24962. PMLR,
  23--29 Jul 2023.
\newblock URL \url{https://proceedings.mlr.press/v202/mitchell23a.html}.

\bibitem[Pham et~al.(2022)Pham, Liu, and Hoi]{DBLP:conf/iclr/PhamLH22}
Quang Pham, Chenghao Liu, and Steven C.~H. Hoi.
\newblock Continual Normalization: Rethinking Batch Normalization for Online
  Continual Learning.
\newblock In \emph{The Tenth International Conference on Learning
  Representations, {ICLR} 2022, Virtual Event, April 25-29, 2022}.
  OpenReview.net, 2022.
\newblock URL \url{https://openreview.net/forum?id=vwLLQ-HwqhZ}.

\bibitem[{Qwen Team}(2026)]{qwen3.5}
{Qwen Team}.
\newblock {Qwen3.5}: Towards Native Multimodal Agents, February 2026.
\newblock URL \url{https://qwen.ai/blog?id=qwen3.5}.

\bibitem[Rao et~al.(2025)Rao, Kumar, Lakkaraju, and
  Shah]{10.1371/journal.pone.0331871}
Vishisht~Srihari Rao, Aounon Kumar, Himabindu Lakkaraju, and Nihar~B. Shah.
\newblock Detecting LLM-generated peer reviews.
\newblock \emph{PLOS ONE}, 20\penalty0 (9):\penalty0 1--22, 09 2025.
\newblock \doi{10.1371/journal.pone.0331871}.
\newblock URL \url{https://doi.org/10.1371/journal.pone.0331871}.

\bibitem[Sadallah et~al.(2025)Sadallah, Baumg{\"a}rtner, Gurevych, and
  Briscoe]{sadallah-etal-2025-good}
Abdelrahman Sadallah, Tim Baumg{\"a}rtner, Iryna Gurevych, and Ted Briscoe.
\newblock The Good, the Bad and the Constructive: Automatically Measuring Peer
  Review{'}s Utility for Authors.
\newblock In Christos Christodoulopoulos, Tanmoy Chakraborty, Carolyn Rose, and
  Violet Peng (eds.), \emph{Proceedings of the 2025 Conference on Empirical
  Methods in Natural Language Processing}, pp.\  28991--29021, Suzhou, China,
  November 2025. Association for Computational Linguistics.
\newblock ISBN 979-8-89176-332-6.
\newblock \doi{10.18653/v1/2025.emnlp-main.1476}.
\newblock URL \url{https://aclanthology.org/2025.emnlp-main.1476/}.

\bibitem[Sadeghian et~al.(2026)Sadeghian, Daqiq, Cheraghi, Ebrahimi, Arabzadeh,
  and Bagheri]{sadeghian2026peerprism}
Soroush Sadeghian, Alireza Daqiq, Radin Cheraghi, Sajad Ebrahimi, Negar
  Arabzadeh, and Ebrahim Bagheri.
\newblock {PeerPrism}: Peer Evaluation Expertise vs Review-writing {AI}.
\newblock In \emph{Proceedings of the 49th International ACM SIGIR Conference
  on Research and Development in Information Retrieval}, 2026.
\newblock \doi{10.1145/3805712.3808602}.
\newblock URL \url{https://arxiv.org/abs/2604.14513}.

\bibitem[Saha et~al.(2026)Saha, Juneja, Chaudhuri, Sajeevan, Shah, and
  Pruthi]{saha2026policies}
Rounak Saha, Gurusha Juneja, Dayita Chaudhuri, Naveeja Sajeevan, Nihar~B. Shah,
  and Danish Pruthi.
\newblock Policies Permitting {LLM} Use for Polishing Peer Reviews Are
  Currently Not Enforceable.
\newblock In \emph{Proceedings of the 43rd International Conference on Machine
  Learning}, 2026.
\newblock URL \url{https://arxiv.org/abs/2603.20450}.

\bibitem[Shao et~al.(2024)Shao, Wang, Zhu, Xu, Song, Bi, Zhang, Zhang, Li, Wu,
  and Guo]{shao2024deepseekmathpushinglimitsmathematical}
Zhihong Shao, Peiyi Wang, Qihao Zhu, Runxin Xu, Junxiao Song, Xiao Bi, Haowei
  Zhang, Mingchuan Zhang, Y.~K. Li, Y.~Wu et~al.
\newblock DeepSeekMath: Pushing the Limits of Mathematical Reasoning in Open
  Language Models, 2024.
\newblock URL \url{https://arxiv.org/abs/2402.03300}.

\bibitem[Team et~al.(2026)Team, Bai, Bai, Bao, Cai, Cao, Charles, Che, Chen,
  Chen, Chen, Chen, Chen, Chen, Chen, Chen, Chen, Chen, Chen, Chen, Chen, Chen,
  Chen, Chen, Chen, Chen, Chen, Chen, Chen, Cheng, Chu, Cui, Deng, Diao, Ding,
  Dong, Dong, Dong, Dong, Du, Du, Du, Du, Du, Fan, Fang, Feng, Feng, Fu, Fu,
  Gao, Gao, Ge, Geng, Gong, Gong, Gongque, Gu, Gu, Gu, Guan, Guo, Hao, He, He,
  He, Hong, Hu, Hu, Hu, Hu, Huang, Huang, Huang, Huang, Jiang, Jiang, Jin,
  Jing, Lai, Li, Li, Li, Li, Li, Li, Li, Li, Li, Li, Li, Li, Li, Li, Li, Li,
  Li, Li, Li, Li, Li, Li, Li, Liao, Lin, Lin, Lin, Lin, Liu, Liu, Liu, Liu,
  Liu, Liu, Liu, Liu, Liu, Liu, Liu, Liu, Liu, Liu, Liu, Liu, Liu, Liu, Lu, Lu,
  Lu, Luo, Luo, Luo, Ma, Ma, Mao, Mei, Men, Meng, Meng, Miao, Ni, Ouyang, Pan,
  Pang, Qian, Qin, Qin, Qiu, Qu, Shang, Shao, Shen, Shen, Shi, Shi, Shi, Song,
  Song, Song, Song, Su, Su, Su, Sui, Sun, Sun, Sun, Sung, Tai, Tang, Tang,
  Tang, Tang, Tao, Teng, Tian, Tian, Wang, Wang, Wang, Wang, Wang, Wang, Wang,
  Wang, Wang, Wang, Wang, Wang, Wang, Wang, Wang, Wang, Wang, Wang, Wang, Wang,
  Wang, Wang, Wang, Wang, Wang, Wang, Wang, Wang, Wang, Wang, Wang, Wang, Wang,
  Wang, Wang, Wang, Wang, Wang, Wei, Wei, Wen, Wen, Wu, Wu, Wu, Wu, Wu, Wu, Wu,
  Wu, Wu, Xiao, Xie, Xie, Xie, Xin, Xing, Xu, Xu, Xu, Xu, Xu, Xu, Xu, Xu, Xu,
  Xu, Xu, Xu, Xu, Xu, Xu, Yan, Yan, Yang, Yang, Yang, Yang, Yang, Yang, Yang,
  Yang, Yang, Yang, Yang, Yang, Yang, Yang, Yao, Ye, Ye, Ye, Yin, Yu, Yu, Yu,
  Yu, Yuan, Yuan, Yuan, Yue, Zeng, Zha, Zhan, Zhang, Zhang, Zhang, Zhang,
  Zhang, Zhang, Zhang, Zhang, Zhang, Zhang, Zhang, Zhang, Zhang, Zhang, Zhang,
  Zhang, Zhang, Zhang, Zhao, Zhao, Zhao, Zhao, Zhao, Zhao, Zhao, Zheng, Zheng,
  Zheng, Zheng, Zhong, Zhong, Zhong, Zhou, Zhou, Zhou, Zhou, Zhu, Zhu, Zhu,
  Zhu, Zhu, Zhuang, Zhuang, Zou, and Zu]{kimiteam2026kimik25visualagentic}
Kimi Team, Tongtong Bai, Yifan Bai, Yiping Bao, S.~H. Cai, Yuan Cao,
  Y.~Charles, H.~S. Che, Cheng Chen, Guanduo Chen et~al.
\newblock Kimi K2.5: Visual Agentic Intelligence, 2026.
\newblock URL \url{https://arxiv.org/abs/2602.02276}.

\bibitem[Thai et~al.(2026)Thai, Emi, Masrour, and Iyyer]{ICLR2026_c7d936e6}
Katherine Thai, Bradley Emi, Elyas Masrour, and Mohit Iyyer.
\newblock EditLens: Quantifying the Extent of AI Editing in Text.
\newblock In C.~Vondrick, B.~Hariharan, C.~Raffel, L.~Pinto, D.~Yang, and
  A.~Faust (eds.), \emph{International Conference on Learning Representations},
  volume 2026, pp.\  122997--123024, 2026.
\newblock URL
  \url{https://proceedings.iclr.cc/paper_files/paper/2026/file/c7d936e696f57468b6b80a7b1463178c-Paper-Conference.pdf}.

\bibitem[Vaswani et~al.(2017)Vaswani, Shazeer, Parmar, Uszkoreit, Jones, Gomez,
  Kaiser, and Polosukhin]{DBLP:conf/nips/VaswaniSPUJGKP17}
Ashish Vaswani, Noam Shazeer, Niki Parmar, Jakob Uszkoreit, Llion Jones,
  Aidan~N. Gomez, Lukasz Kaiser, and Illia Polosukhin.
\newblock Attention is All you Need.
\newblock In Isabelle Guyon, Ulrike von Luxburg, Samy Bengio, Hanna~M. Wallach,
  Rob Fergus, S.~V.~N. Vishwanathan, and Roman Garnett (eds.), \emph{Advances
  in Neural Information Processing Systems 30: Annual Conference on Neural
  Information Processing Systems 2017, December 4-9, 2017, Long Beach, CA,
  {USA}}, pp.\  5998--6008, 2017.
\newblock URL
  \url{https://proceedings.neurips.cc/paper/2017/hash/3f5ee243547dee91fbd053c1c4a845aa-Abstract.html}.

\bibitem[Wu et~al.(2025)Wu, Yang, Zhan, Yuan, Chao, and
  Wong]{wu-etal-2025-survey}
Junchao Wu, Shu Yang, Runzhe Zhan, Yulin Yuan, Lidia~Sam Chao, and Derek~Fai
  Wong.
\newblock A Survey on {LLM}-Generated Text Detection: Necessity, Methods, and
  Future Directions.
\newblock \emph{Computational Linguistics}, 51\penalty0 (1):\penalty0 275--338,
  March 2025.
\newblock \doi{10.1162/coli_a_00549}.
\newblock URL \url{https://aclanthology.org/2025.cl-1.8/}.

\bibitem[Yang et~al.(2025)Yang, Li, Yang, Zhang, Hui, Zheng, Yu, Gao, Huang,
  Lv, Zheng, Liu, Zhou, Huang, Hu, Ge, Wei, Lin, Tang, Yang, Tu, Zhang, Yang,
  Yang, Zhou, Zhou, Lin, Dang, Bao, Yang, Yu, Deng, Li, Xue, Li, Zhang, Wang,
  Zhu, Men, Gao, Liu, Luo, Li, Tang, Yin, Ren, Wang, Zhang, Ren, Fan, Su,
  Zhang, Zhang, Wan, Liu, Wang, Cui, Zhang, Zhou, and
  Qiu]{yang2025qwen3technicalreport}
An~Yang, Anfeng Li, Baosong Yang, Beichen Zhang, Binyuan Hui, Bo~Zheng, Bowen
  Yu, Chang Gao, Chengen Huang, Chenxu Lv et~al.
\newblock Qwen3 Technical Report, 2025.
\newblock URL \url{https://arxiv.org/abs/2505.09388}.

\bibitem[Yu et~al.(2026)Yu, Luo, Madasu, Lal, and Howard]{ICLR2026_64592277}
Sungduk Yu, Man Luo, Avinash Madasu, Vasudev Lal, and Phillip Howard.
\newblock Is Your Paper Being Reviewed by an LLM? Benchmarking AI Text
  Detection in Peer Review.
\newblock In C.~Vondrick, B.~Hariharan, C.~Raffel, L.~Pinto, D.~Yang, and
  A.~Faust (eds.), \emph{International Conference on Learning Representations},
  volume 2026, pp.\  61517--61553, 2026.
\newblock URL
  \url{https://proceedings.iclr.cc/paper_files/paper/2026/file/64592277a1cbb02ed6e26b6e9e5f458c-Paper-Conference.pdf}.

\bibitem[Zhang et~al.(2025)Zhang, Li, Long, Zhang, Lin, Yang, Xie, Yang, Liu,
  Lin, Huang, and Zhou]{qwen3embedding}
Yanzhao Zhang, Mingxin Li, Dingkun Long, Xin Zhang, Huan Lin, Baosong Yang,
  Pengjun Xie, An~Yang, Dayiheng Liu, Junyang Lin et~al.
\newblock Qwen3 Embedding: Advancing Text Embedding and Reranking Through
  Foundation Models.
\newblock \emph{arXiv preprint arXiv:2506.05176}, 2025.

\bibitem[Zhou et~al.(2026)Zhou, Zhang, Dai, Hershcovich, and Li]{Zhou_2026}
Li~Zhou, Ruijie Zhang, Xunlian Dai, Daniel Hershcovich, and Haizhou Li.
\newblock ScholarLens: Tracking the growing penetration of Large Language
  Models in scholarly writing and peer review.
\newblock \emph{Engineering Applications of Artificial Intelligence},
  177:\penalty0 114988, August 2026.
\newblock ISSN 0952-1976.
\newblock \doi{10.1016/j.engappai.2026.114988}.
\newblock URL \url{http://dx.doi.org/10.1016/j.engappai.2026.114988}.

\end{thebibliography}
\bibliographystyle{peerreveal_references}

\appendix
\raggedbottom
\section{Review-Mechanism Search Details}
\label{app:mechanism-details}

\subsection{Full Search Controller}
\label{app:search-controller}

Algorithm~\ref{alg:review-mechanism-search-full} expands the compact controller in Algorithm~\ref{alg:review-mechanism-search}. It records every complete-program evaluation and uses the trace when proposing replacements, combinations, and failure-driven repairs.

\begin{algorithm}[H]
\caption{Full automatic quality-constrained mechanism-search controller}
\label{alg:review-mechanism-search-full}
\small
\begin{algorithmic}[1]
\Require base prompt $p_0$; class-specific operation registry $\mathcal{O}$; balanced development schedule $\mathcal{D}$; detector panel $\mathcal{E}$; search budget $B$; output size $K$
\State $\mathcal{F}\gets\{\varnothing\}$; $\mathcal{A}\gets\varnothing$; $\mathcal{T}\gets\varnothing$
\For{$b=1,\ldots,B$}
    \State $\mathcal{P}\gets\Call{Propose}{\mathcal{F},\mathcal{A},\mathcal{T},\mathcal{O};\ \textsc{Add},\textsc{Replace},\textsc{Combine},\textsc{Repair}}$
    \ForAll{complete programs $z\in\mathcal{P}$}
        \State $p_z\gets\Call{Compile}{p_0,z}$; $R_z\gets\Call{BalancedGenerate}{p_z,\mathcal{D}}$
        \State $(D_z,Q_z,V_z)\gets\Call{Evaluate}{R_z,\mathcal{E}}$
        \State $U_z\gets\Call{MarginalDiversity}{z,R_z,\mathcal{A}}$
        \State append $(z,D_z,U_z,Q_z,V_z)$ to trace $\mathcal{T}$
        \If{$Q_z\geq\tau_Q$ and $V_z\geq\tau_V$}
            \State $\mathcal{A}\gets\mathcal{A}\cup\{z\}$
        \EndIf
    \EndFor
    \State $\mathcal{F}\gets\Call{SelectFrontier}{\mathcal{P},\mathcal{A};\{D_z,U_z\}_{z\in\mathcal{P}}}$
\EndFor
\State $\mathcal{A}\gets\Call{Confirm}{\mathcal{A},\mathcal{D}}$ \Comment{larger balanced pilots}
\State $\mathcal{S}^{*}\gets\Call{SelectComplementary}{\mathcal{A},K;\{D_z,U_z\}_{z\in\mathcal{A}}}$
\State \Return $\Call{Freeze}{\mathcal{S}^{*}}$ and trace $\mathcal{T}$
\end{algorithmic}
\end{algorithm}

$\mathcal{F}$ is the search frontier, $\mathcal{A}$ contains quality- and validity-qualified programs, and $\mathcal{T}$ stores their complete evaluations. \textsc{Add} introduces an operation, \textsc{Replace} changes one operation, \textsc{Combine} merges compatible programs, and \textsc{Repair} corrects diagnosed validity failures. Difficulty and marginal diversity determine frontier promotion and final complementary selection.

\paragraph{Development detectors.}
The ML search uses four frozen three-class detectors: a fine-tuned RoBERTa-large~\citep{liu2019robertarobustlyoptimizedbert}, \semdetect~\citep{duarte2026semdetect}, and two auxiliary claim-based models, Target-only and Pairwise. Nature Communications and PeerJ use the latter three. This panel covers whole-review classification, predefined semantic and textual features, and learned claim representations. RoBERTa averages chunk representations before a three-class linear head; \semdetect uses its nine-feature LightGBM classifier. The auxiliary models operate on frozen 1,024-dimensional claim embeddings and are described below. Both auxiliary models are trained on the ML regular corpus using an 85/15 training--validation split and remain frozen during search. \rev{They are earlier versions of our own claim-based architecture, so the search also targets the method family of \peerreveal, although \peerreveal itself is not in the panel.}

\paragraph{Target-only.}
Each target claim is projected to 64 dimensions with layer normalization and GELU, then encoded by a 128-unit feed-forward layer with ReLU and dropout. A learned query scores the claim vectors; softmax-normalized scores weight their sum into a review representation. An MLP with a 128-unit hidden layer outputs three-class logits. This model uses neither reference claims nor textual features.

\paragraph{Pairwise.}
Separate projections map target and reference claims to 64 dimensions. A pair-scoring MLP with 64 hidden units takes the two vectors, their elementwise absolute difference, and their elementwise product. For each target claim, softmax over these scores forms a weighted reference vector. The target vector, weighted reference vector, their absolute difference and product, and the maximum pair score are encoded into a 128-dimensional claim representation. Learned-query attention pooling then summarizes the target claims. A separate 32-unit feed-forward layer encodes the four textual features; its output is concatenated with the pooled claim representation and passed to a 128-unit MLP with a three-class output.

\subsection{Acceptance Constraints}
\label{app:acceptance-constraints}

The thresholds $\tau_Q$ and $\tau_V$ of Section~\ref{sec:evaluation-construction} are fixed before the search. A candidate program joins $\mathcal{A}$ only if all of the following hold over its pilot reviews: at least $85\%$ pass the quality judge of Appendix~\ref{app:quality-judge}; at least $95\%$ of the pilot reviews are distinct; and the program reduces target-class recall relative to the matched control for at least two development detectors. For ML the gate additionally requires that the median review length and the median retained-claim count each reach $60\%$ of the matched human median. For Nature Communications and PeerJ it instead requires that at least $85\%$ of the pilot reviews pass the class-specific validity check.

Validity is checked per class. A fully AI-generated review passes when it is paper-grounded, contains a judgement and a supporting reason, is consistent with the requested rating, and is on topic. A rewrite passes when it preserves the source review's central judgement, main reasons, and stance, contains a judgement and a supporting reason, is an independent reconstruction, adds no substantive claim, and is on topic. Empty outputs and outputs carrying prompt residue are rejected before either check.

In the ML search, promotion from pilot to confirmation additionally requires, on a larger balanced pilot, that at least three of the four development detectors lose recall, that no detector gain more than $.05$ recall, and that the median recall reduction reach $.10$.

Among qualified programs, difficulty is ranked lexicographically by the minimum per-detector recall reduction relative to the matched control, then the median reduction, then the negated worst-case absolute recall; among equally difficult programs, the one with less operation overlap with those already selected is preferred. Per-detector recalls, mean AI probabilities, and error directions are recorded for every evaluation.

\subsection{Class-Specific Operation Spaces}

AI operations span four families: \textbf{evidence selection} over assumptions, baselines, results, conditions, scope, and limitations; \textbf{critique formation} via dependency analysis, baseline--question alignment, alternative interpretation, consequence reasoning, and counterfactuals; \textbf{judgement formation} linking evidence to confidence, interpretation, or recommendation; and \textbf{grounding repair} preserving metric direction, experimental conditions, evidence scope, and strength--boundary consistency.

The rewrite registry begins from source-order bullet notes extracted from the human review (Appendix~\ref{app:frozen-prompts}), which record its central judgement, reasons, and stance. Operations vary organization, emphasis, judgement placement, prose and section structure, argumentative expansion, and connections among the original reasons. Failure-driven repair removes direct paraphrasing, incomplete reasoning, unsupported additions, and loss of source stance.

\subsection{Selected Mechanisms}
\label{app:selected-mechanisms}

The search runs separately for each domain over the shared operation library, so the selected mechanisms differ across domains. Table~\ref{tab:challenge-ai-mechanisms} lists the selected AI-generation mechanisms and Table~\ref{tab:challenge-rewrite-mechanisms} the selected rewrite mechanisms, together with the number of PeerShift reviews each mechanism produced. Appendix~\ref{app:frozen-prompts} gives the exact prompt text of every mechanism.

\begin{table}[htbp]
\caption{Selected AI-generation mechanisms by domain.}
\label{tab:challenge-ai-mechanisms}
\centering\small
\setlength{\tabcolsep}{4pt}
\begin{tabular}{ll>{\raggedright\arraybackslash}p{8.2cm}r}
\toprule
Domain & ID & Reasoning-operation composition & Reviews \\
\midrule
ML & ML-A1 & Assumption dependency + alternative result interpretations & 152 \\
 & ML-A2 & Baseline--question fit + alternative result interpretations + observation--consequence reasoning & 152 \\
 & ML-A3 & Baseline--question fit + decision counterfactual & 153 \\
 & ML-A4 & Verified same-evidence strength/boundary judgement & 153 \\
\midrule
Nature & NC-A1 & Methodological benefit--cost tradeoff & 150 \\
 & NC-A2 & Constructive resolution effect + methodological tradeoff & 150 \\
 & NC-A3 & Claim--evidence clarification questions + alternative result interpretations, with grounding repair & 150 \\
 & NC-A4 & Cross-section claim--evidence consistency + methodological tradeoff & 150 \\
\midrule
PeerJ & PJ-A1 & Verified strength--limitation chain, with grounding repair & 200 \\
 & PJ-A2 & Cross-section claim--evidence consistency & 200 \\
 & PJ-A3 & Verified decision counterfactual, with grounding repair & 200 \\
\bottomrule
\end{tabular}
\end{table}

\begin{table}[!tb]
\caption{Selected rewrite mechanisms by domain, with the reconstruction instruction appended to the shared rewrite prompt (Appendix~\ref{app:frozen-prompts}).}
\label{tab:challenge-rewrite-mechanisms}
\centering\small
\setlength{\tabcolsep}{4pt}
\begin{tabular}{ll>{\raggedright\arraybackslash}p{9.4cm}r}
\toprule
Domain & ID & Appended reconstruction instruction & Reviews \\
\midrule
ML & ML-R1 & Make the final judgement feel calibrated to the evidence, but keep the wording less polished and recognizably reviewer-like. & 102 \\
 & ML-R2 & Embed the recommendation inside the discussion instead of isolating it. Keep the prose slightly rough but coherent. & 102 \\
 & ML-R3 & Write in less polished but still professional reviewer prose. Keep the judgement and reasons explicit, and avoid generic assistant-like smoothing. & 102 \\
 & ML-R4 & Embed the recommendation in the discussion rather than isolating it. Keep the wording conservative and slightly less polished, while preserving every main reason from the notes. & 102 \\
 & ML-R5 & Reconstruct the review in plain, slightly less polished reviewer prose. Include a clear judgement and reason, but do not add new claims or uplift the recommendation beyond the source notes. & 101 \\
 & ML-R6 & Reconstruct a complete reviewer note from the bullet notes. Use natural but not overly polished prose, preserve the stance and all main reasons, and include one explicit judgement with its reason. & 101 \\
\midrule
Nature & NC-R1 & Use less-polished but professional reviewer wording. Use a natural reviewer narrative rather than rubric prose. Use plain or straightforward reviewer prose. Avoid template-like transitions and phrasing. & 100 \\
 & NC-R2 & Use less-polished but professional reviewer wording. Avoid generic assistant-like smoothing. & 100 \\
 & NC-R3 & Use a natural reviewer narrative rather than rubric prose. & 100 \\
 & NC-R4 & Use plain or straightforward reviewer prose. Use less-polished but professional reviewer wording. Avoid generic assistant-like smoothing. & 100 \\
 & NC-R5 & Embed the recommendation inside the discussion. & 100 \\
 & NC-R6 & Embed the recommendation inside the discussion. Use slightly rough but coherent prose. & 100 \\
\midrule
PeerJ & PJ-R1 & Place the main judgement before its supporting reasons. & 120 \\
 & PJ-R2 & Use a natural or plain reviewer voice. Use light reviewer hedging. Permit uneven transitions found in reviewer prose. State an explicit judgement together with its reason. & 120 \\
 & PJ-R3 & Preserve mixed confidence and both promising and concerning aspects. & 120 \\
 & PJ-R4 & Preserve mixed confidence and both promising and concerning aspects. Use plain or straightforward reviewer prose. & 120 \\
 & PJ-R5 & Produce a shorter reconstruction while retaining central content. Be concise while retaining substantive judgement and reasoning. & 120 \\
\bottomrule
\end{tabular}
\end{table}

\subsection{Frozen Prompt Texts}
\label{app:frozen-prompts}

\paragraph{AI generation.} Every AI-generated review uses one base prompt, which fixes the venue and the requested rating; the requested rating is taken from the human reviews of the same paper, and the paper text is supplied with the prompt:
\begin{quote}\small
Review the given paper for \textit{[a top AI conference $\mid$ a Nature Communications journal submission $\mid$ a PeerJ Psychology / Social Sciences journal submission]}. Be critical, focused, constructive, and paper-grounded, and make the final recommendation exactly ``\textit{[rating]}.'' Give a complete assessment of the contribution, supported strengths, substantive weaknesses or limitations, clarification questions where they matter, and the reasoning behind the judgement. Organize the review as natural reviewer prose without a prescribed five-part heading scaffold.
\end{quote}
A mechanism appends a single block whose operation sentences are the only part that varies:
\begin{quote}\small
Evaluation-reasoning guidance:\\
Shape the assessment through these paper-grounded reasoning relations when applicable: \textit{[operation sentences]}. Tie each resulting judgement to evidence that is verifiably present, and treat uncertainty as a scoped question rather than an asserted deficiency. Keep the assessment complete and preserve the exact requested recommendation.
\end{quote}
The operation sentences of the selected mechanisms are:
\begin{description}\small
\item[ML-A1] Identify assumptions on which an important conclusion depends; Explain what changes if an assumption is weakened, violated, or left unverified; Reflect that dependency in the assessment; When a reported result admits more than one explanation, distinguish the plausible interpretations; Identify what evidence favors one interpretation over the others; Calibrate the assessment to the remaining ambiguity.
\item[ML-A2] Evaluate whether each important comparison actually tests the question the paper asks; Explain what the baseline can establish and what it cannot; Reflect any mismatch in the interpretation of the reported advantage; When a reported result admits more than one explanation, distinguish the plausible interpretations; Identify what evidence favors one interpretation over the others; Calibrate the assessment to the remaining ambiguity; Begin evaluative points from a concrete observation in the paper; Explain the consequence of that observation for validity, usefulness, or interpretation; Make the consequence proportional to what the observation actually supports.
\item[ML-A3] Evaluate whether each important comparison actually tests the question the paper asks; Explain what the baseline can establish and what it cannot; Reflect any mismatch in the interpretation of the reported advantage; State which verified change in evidence or interpretation would alter the recommendation; Explain why the current record falls on one side of that decision boundary; Keep the counterfactual plausible and specific to the paper.
\item[ML-A4] Before writing the final review, silently verify every material statement against the provided paper; For each decisive comparison, theoretical statement, or experimental interpretation, name its Section, Table, Figure, Equation, or Theorem when one is available; preserve the metric meaning, direction, values, assumptions, and scope exactly; Do not use external facts or assert statistical significance, computational cost, missing baselines, missing experiments, or missing ablations unless the paper explicitly establishes them; If an absence or interpretation cannot be verified, retain the issue as a clarification question rather than a factual assertion; Use the same explicitly reported observation to explain both a genuine strength and the boundary of what that strength supports, and carry that evidence-weighted distinction into the requested recommendation while still providing the complete review requested by the base prompt.
\item[NC-A1] Select a methodological choice that creates a benefit and a cost; Explain whether the benefit-cost tradeoff is appropriate for the paper's stated objective; Tie the methodological tradeoff to the overall assessment.
\item[NC-A2] Select an identified issue for which a concrete author response would matter; Select a methodological choice that creates a benefit and a cost; Explain what a concrete requested change would resolve and why that resolution matters; Explain whether the benefit-cost tradeoff is appropriate for the paper's stated objective; Connect the effect of the requested resolution to the overall assessment; Tie the methodological tradeoff to the overall assessment.
\item[NC-A3] Identify a central position stated by the paper before evaluating it; Select a reported result whose interpretation affects the assessment; Select a specific unresolved interpretation that warrants clarification; Distinguish plausible interpretations of the result and identify what evidence favors one over another; Explain how different answers to the unresolved interpretation would change confidence or judgement; Separate what the evidence resolves from an unresolved issue that should remain a question; Calibrate confidence and recommendation to the ambiguity that remains; Connect the decision-relevant clarification to the requested recommendation; Express an unverifiable absence or interpretation as a clarification question rather than a factual assertion; Do not generalize beyond the evidence or invent an unreported factual, experimental, theoretical, or evaluative claim.
\item[NC-A4] Select a methodological choice that creates a benefit and a cost; Select related claims or evidence stated in different parts of the paper; Explain whether the benefit-cost tradeoff is appropriate for the paper's stated objective; Determine whether the selected statements reinforce one another or create a substantive inconsistency; Tie the methodological tradeoff to the overall assessment; Explain how the consistency or conflict affects confidence and recommendation.
\item[PJ-A1] Select a genuine paper-supported strength whose value depends on a condition or limitation; Explain whether the limitation merely qualifies the strength or substantially undercuts it; Carry the strength-limitation dependency into the judgement; Preserve metric meaning, direction, values, assumptions, and scope exactly, including reported conditions; Name the relevant Section, Table, Figure, Equation, or Theorem when one is available; Express an unverifiable absence or interpretation as a clarification question rather than a factual assertion; Do not generalize beyond the evidence or invent an unreported factual, experimental, theoretical, or evaluative claim; Use only facts and evidence contained in the provided paper.
\item[PJ-A2] Select related claims or evidence stated in different parts of the paper; Determine whether the selected statements reinforce one another or create a substantive inconsistency; Explain how the consistency or conflict affects confidence and recommendation.
\item[PJ-A3] Identify the paper-supported evidence or interpretation that is decision-critical; State which verified clarification or change in evidence would alter the recommendation; Explain why the current evidence falls on the stated side of the decision boundary; Preserve metric meaning, direction, values, assumptions, and scope exactly, including reported conditions; Name the relevant Section, Table, Figure, Equation, or Theorem when one is available; Express an unverifiable absence or interpretation as a clarification question rather than a factual assertion; Do not generalize beyond the evidence or invent an unreported factual, experimental, theoretical, or evaluative claim; Use only facts and evidence contained in the provided paper.
\end{description}

\paragraph{Rewriting.} Rewrite writers do not see the human review. A fixed extraction step (Mistral Large 3) first converts each source review into source-order bullet notes. For Nature Communications and PeerJ the extraction instruction is ``Extract source-order notes from one academic peer review. Return only bullet notes. Preserve the central judgement, every main supporting reason, strengths, weaknesses, technical concerns, author questions, stance, and any recommendation. Do not add content.'' For ML it is ``Convert the provided peer review into concise source-order bullet notes. Preserve the reviewer's stance, main points, questions, and judgement if they appear in the source. Do not add new content, do not improve the review, and do not impose section headings. Return only bullet notes.'' The writer then receives the system prompt ``Return only the rewritten peer review text. Do not include reasoning, JSON, markdown fences, instruction summaries, bullet notes, or comments about being an AI model.'' and the reconstruction prompt:
\begin{quote}\small
The following text is source-order bullet notes extracted from one human peer review. Reconstruct a meaningful rewritten peer review from those notes.\\
Hard constraints:\\
-- Preserve the original reviewer's main judgement, reasons, and stance.\\
-- Do not add substantive new claims beyond the bullet notes.\\
-- Do not change the final judgement.\\
-- Include at least one clear judgement about the paper and the reason for that judgement.\\
-- Do not use donor human-style profile statistics; do not mention this instruction.
\end{quote}
For ML, the mechanism follows as ``Style constraint: \textit{[instruction]}'' and the bullet notes close the prompt. For Nature Communications and PeerJ, the prompt additionally states ``Independently reconstruct the complete review from the notes. Direct paraphrasing or imitation of sentences from the source review is prohibited.'', supplies the paper abstract as context, and lists the mechanism's operations as numbered lines under ``Reconstruction mechanism:'' before the notes. Table~\ref{tab:challenge-rewrite-mechanisms} gives the appended text of every mechanism.

\subsection{PeerShift Composition}

\begin{table}[H]
\caption{Composition of PeerShift by domain. Parentheses give the number of writer models and mechanisms behind each generated class.}
\label{tab:challenge-composition}
\centering
\small
\begin{tabular}{lrll}
\toprule
Domain & Human & Fully AI-generated & AI-rewritten \\
\midrule
ML & 610 & 610 (10 writers, 4 mechanisms) & 610 (5 writers, 6 mechanisms) \\
Nature Communications & 600 & 600 (10 writers, 4 mechanisms) & 600 (5 writers, 6 mechanisms) \\
PeerJ & 600 & 600 (10 writers, 3 mechanisms) & 600 (5 writers, 5 mechanisms) \\
\bottomrule
\end{tabular}
\end{table}

The generated classes span distinct writer-model pools, generation procedures, and review styles. Every generated review must contain a reasoned judgement; AI-generated reviews must also match the requested rating and pass the paper-grounding check, and rewrites must keep the source judgement and pass the source-fidelity check. Mechanisms are selected and frozen at the program level before balanced full-set generation.

The PeerJ source corpus covers 460 Psychology and 140 Social Sciences papers; its initial manuscripts are paired with first-round human reviews. Across the three subsets, the final set contains 1,810 reviews per class and 5,430 in total.

Table~\ref{tab:writer-versions} lists generation models and review counts. Writers instantiate the selected reasoning or reconstruction programs; test reviews are separate from those generated locally for detector training.

\begin{table}[!tb]
\caption{PeerShift generation models and per-domain counts. $\dag$ links to official documentation.}
\label{tab:writer-versions}
\centering\small
\begin{tabular}{llrrr}
\toprule
Class & Writer model & ML & Nature & PeerJ\\
\midrule
AI & Gemini 3.1 Pro Preview~\href{https://ai.google.dev/gemini-api/docs/models/gemini-3.1-pro-preview}{\textsuperscript{\dag}} & 61 & 60 & 60\\
 & DeepSeek V4 Pro~\citep{deepseekai2026deepseekv4highlyefficientmilliontoken} & 61 & 60 & 60\\
 & MiniMax M2.5~\href{https://huggingface.co/MiniMaxAI/MiniMax-M2.5}{\textsuperscript{\dag}} & 61 & 60 & 60\\
 & Mistral Large 3 (675B)~\href{https://huggingface.co/mistralai/Mistral-Large-3-675B-Instruct-2512}{\textsuperscript{\dag}} & 61 & 60 & 60\\
 & Qwen3-Next-80B-A3B~\href{https://huggingface.co/Qwen/Qwen3-Next-80B-A3B-Instruct}{\textsuperscript{\dag}} & 61 & 60 & 60\\
 & GLM-5~\citep{glm5team2026glm5vibecodingagentic} & 61 & 60 & 60\\
 & Kimi K2.5~\citepalias{kimiteam2026kimik25visualagentic} & 61 & 60 & 60\\
 & Grok 4.3~\href{https://docs.x.ai/developers/models/grok-4.3}{\textsuperscript{\dag}} & 61 & 60 & 60\\
 & Claude Sonnet 4.6~\href{https://www.anthropic.com/news/claude-sonnet-4-6}{\textsuperscript{\dag}} & 61 & 60 & 60\\
 & GPT-5.5~\href{https://developers.openai.com/api/docs/models/gpt-5.5}{\textsuperscript{\dag}} & 61 & 60 & 60\\
\midrule
Rewrite & Gemini 3.5 Flash~\href{https://ai.google.dev/gemini-api/docs/whats-new-gemini-3.5}{\textsuperscript{\dag}} & 122 & 120 & 120\\
 & DeepSeek V4 Pro & 122 & 120 & 120\\
 & Kimi K2.5 & 122 & 120 & 120\\
 & Grok 4.3 & 122 & 120 & 120\\
 & GPT-5.5 & 122 & 120 & 120\\
\bottomrule
\end{tabular}
\end{table}

\subsection{\texorpdfstring{\rev{Faithfulness of AI-Rewritten Reviews}}{Faithfulness of AI-Rewritten Reviews}}
\label{app:rewrite-faithfulness}
\begin{revblock}
Every AI-rewritten review is reconstructed from a real human review of the same paper, and that source review is also in the test set, so each pair can be checked directly. An LLM judge (Claude Sonnet 4.6) compared all 1,810 pairs. It finds 91.2\% of rewrites faithful (95\% CI 89.8--92.4; ML 97.9, Nature Communications 86.7, PeerJ 88.8). The overall judgement is kept in 99.5\% of pairs and the stance in 99.4\%, and 95.4\% of the main reasons are kept on average. Unfaithful rewrites mostly drop a detail, such as a citation or a line-level fix, without changing the reviewer's opinion.

Like \semdetect's rewrites, each PeerShift rewrite starts from one human review. We compare their claims with the 1,024-dimensional claim embeddings of \peerreveal: for each claim of the source review, we take its highest cosine similarity to any claim of the rewrite and average over claims ($\times 100$). PeerShift-ML rewrites reach 78.4, close to the 81.2 of \semdetect's rewrites on its Original test set, and well above the 62.9 of the AI-generated review of the same paper.
\end{revblock}

\section{AI Generator Training Details}
\label{app:generator-training-details}

\subsection{Detector Score Mappings}
\label{app:detector-score-mappings}
The weights in Equation~\ref{eq:combined-evasion} are $.30$, $.40$, $.15$, and $.15$ for \semdetect, \peerreveal, \radar~\citep{NEURIPS2023_30e15e59}, and \anchor~\citep{ICLR2026_64592277}, respectively. \rev{They are fixed before generator training. \peerreveal receives the largest weight because the generated data is meant to improve it, and \semdetect is the strongest prior three-class detector; \radar and \anchor add two further detector families so that the generator does not overfit to one detector family.} \semdetect's evasion score is $P(\mathrm{Human})+.3P(\mathrm{Rewrite})$. For \peerreveal, we use the human-class logit margin
\begin{equation}
m_H=\ell_H-\log\!\left(\exp(\ell_{\mathrm{AI}})+\exp(\ell_{\mathrm{Rewrite}})\right).
\end{equation}
Three increasing calibration anchors map to $0$, $.5$, and $1$; linear interpolation between anchors and clipping outside them produce the evasion score. \radar's human-probability signal is mapped through a fixed empirical cumulative distribution. \anchor provides a binary evasion score at its threshold calibrated to a $.01$ false-positive rate. The mappings stay fixed throughout optimization. \rev{The \peerreveal checkpoint that supplies the reward is an earlier version of our detector, trained only on the ML regular corpus before any three-domain, adversarial, or PeerShift data existed. It reaches 96.8, 94.7, and 89.2\% three-class Macro-F1 on \semdetect's Original, OOD-M, and OOD-M+P test sets. It stays frozen and is shared by the generators of all three domains. The released \peerreveal is trained afterwards and never supplies a reward. The development and test sets use the same writer models on different papers; Appendix~\ref{app:selection-exposure} shows that this overlap does not affect checkpoint selection.}

\subsection{Quality Judge}
\label{app:quality-judge}

The same frozen judge scores review quality for the mechanism-search acceptance constraints (Appendix~\ref{app:acceptance-constraints}), for generator credit assignment, and for generator evaluation. It is Mistral Large 3 (675B)\footnote{\url{https://huggingface.co/mistralai/Mistral-Large-3-675B-Instruct-2512}} served through Amazon Bedrock at temperature $0$ with a budget of $512$ output tokens; a JSON schema is supplied in the prompt text and decoding is unconstrained. The judge receives the paper title, the manuscript text truncated to $12{,}000$ characters, and the review truncated to $20{,}000$ characters. It does not see the writer model, the mechanism, the detector scores, or whether the review is a training sample. Its system prompt is:
\begin{quote}\small
You are a strict peer-review quality judge. Your task is to evaluate whether a submitted review is a coherent, useful, paper-specific peer review. You must judge the review as a review, not as generic academic-sounding text. Penalize text that is repetitive, globally incoherent, semantically thin, corrupted, full of invented or irrelevant terms, or not grounded in the paper. Do not reward length by itself. A concise coherent review is better than a long noisy one. Return only valid JSON.
\end{quote}
The judge rates five dimensions from $1$ to $5$ (Table~\ref{tab:quality-rubric}) and returns a validity flag with a one-sentence rationale. Its score averages the first four dimensions; format compliance is recorded but excluded, treating free-form human and structured AI-generated reviews equally.

During generator training the judge runs in parallel with RevUtil~\citep{sadallah-etal-2025-good}\footnote{\url{https://huggingface.co/boda/RevUtil_Llama-3.1-8B-Instruct_score_only}}, a Llama-3.1-8B-Instruct adapter that scores the whole review. A review passes only when both averages reach $3$, and the quality score is the smaller of the two; when RevUtil returns no parsable score, the judge's average is used in its place.

\begin{table}[h]
\caption{Quality-judge rubric. The quality score averages the first four dimensions; format compliance is reported separately and does not enter the score or the eligibility rule.}
\label{tab:quality-rubric}
\centering\small
\begin{tabular}{lp{8.8cm}}
\toprule
Dimension & Question put to the judge\\
\midrule
Coherence & Does the review make sense as a whole, with clear logical flow and no word salad?\\
Paper specificity & Is the review grounded in concrete details from this paper rather than generic academic criticism?\\
Usefulness & Would the authors receive useful guidance from this review?\\
Non-repetition & Does the review avoid repetitive phrasing, repeated claims, and filler?\\
\midrule
Format compliance & Does the review have a recognizable peer-review structure?\\
\bottomrule
\end{tabular}
\end{table}

Eligibility is decided from the numbers: a review is accepted whenever its mean score is at least $3$ and no individual dimension falls below $3$, and rejected whenever its mean score is below $3$, overriding the returned flag in both directions. This constraint rejects generic or incoherent evasive reviews during generator training and mechanism search.

\subsection{Quality-Constrained Group Credit}
\label{app:group-credit}
Let $V$ contain the eligible reviews in a four-review group. Eligibility requires valid output formatting, a quality score of at least $3$ on a five-point scale, an accepted length of $180$--$840$ words, and no input-artifact exploitation. When $|V|\geq2$ and the evasion-score range within $V$ is at least $.01$, the basic advantage for $i\in V$ is
\begin{equation}
\widetilde A_i=E(r_i)-\frac{1}{|V|}\sum_{j\in V}E(r_j).
\end{equation}
The basic advantage is zero otherwise, and is not divided by a group standard deviation. Truncation, length, and input-artifact failures receive negative credit relative to eligible peers, followed by recentering over the affected samples and eligible peers. Formatting failures are handled by a separate reward term. Other quality failures do not enter the evasion baseline. Groups without an eligible review use only nonpositive length-recovery credit where applicable.

\subsection{Policy Optimization}
\label{app:policy-optimization}
Each update samples $12$ paper--rating--strategy conditions with four independent reviews per condition, yielding $48$ reviews. We optimize all generator parameters with the clipped GRPO~\citep{shao2024deepseekmathpushinglimitsmathematical} objective, using a clipping parameter of $.2$, one policy epoch per rollout batch, and no KL penalty. AdamW~\citep{DBLP:conf/iclr/LoshchilovH19} uses a learning rate of $3\times10^{-6}$, weight decay of $.01$, and gradient clipping at $2$. Detector and quality scorers remain frozen throughout the update.

\subsection{Generator Evaluation}
\label{app:generator-evaluation}
Domain-specific generators are independently trained from the same Qwen3.5-9B~\citep{qwen3.5} base model. Before and after training, we use the same 17 held-out papers per domain, ratings, prompts, and sampling settings, generating four reviews per condition. ML uses all three strategy conditions; Nature Communications and PeerJ use paper-only generation.

A review is eligible if its output format is valid, the quality judge accepts it as a valid review, and its mean quality-dimension score is at least 3 out of 5. Eligibility sets are fixed before detector evaluation and shared by all thirteen detectors. Before and after training retain their respective eligible sets: 198 and 203 ML reviews, 68 and 50 Nature Communications reviews, and 67 and 50 PeerJ reviews. Evasion is the number of eligible reviews classified as non-AI divided by the number of eligible reviews. Pooling across domains gives denominators of 333 before and 303 after training.

Each detector is calibrated on the same 1,711 human-written PeerShift reviews (545 ML, 583 Nature Communications, and 583 PeerJ); the human reviews of the 51 generator-evaluation papers are excluded, so that no paper contributes both calibration reviews and evaluated reviews. Thresholds are determined from human calibration scores, and tied scores are not split. Tables~\ref{tab:generator-evasion-domain-1} and~\ref{tab:generator-evasion-domain-5} report domain-wise evasion at 1\% and 5\% human FPR, respectively. At 5\% FPR, empirical false-positive rates range from 4.44\% to 4.97\% on this calibration set.

At 1\% FPR, evasion against \peerrevealbase increases by 48.99, 63.65, and 6.96 percentage points on ML, Nature Communications, and PeerJ, respectively; paired-paper bootstrap 95\% confidence intervals are [40.20, 56.78], [47.39, 79.41], and [$-0.70$, 14.25], using 10,000 resamples. At 5\% FPR, the corresponding increases are 74.27, 85.06, and 76.57 points, with confidence intervals [68.88, 79.90], [74.14, 94.55], and [63.03, 89.15]. Evasion against \editlens~\citep{ICLR2026_c7d936e6} also increases in every domain at 1\% FPR, by 34.26, 16.18, and 30.84 points with intervals [24.64, 44.25], [4.41, 30.88], and [17.65, 44.40]; none of these intervals crosses zero. After generator training, adversarially trained \roberta has the lowest pooled evasion at 1\% human FPR (42.90\%), while regular \roberta has the lowest at 5\% (30.03\%). Before training, \editlens is almost never evaded at 5\% human FPR: only 1 of the 333 eligible reviews is classified as non-AI, so editing extent alone separates untrained generator output from human writing almost perfectly.

\begin{table}[H]
\caption{Domain-wise evasion rates (\%) at 1\% human FPR. Each cell shows before $\rightarrow$ after training. $\dagger$ marks detector families used for generator-training feedback.}
\label{tab:generator-evasion-domain-1}
\centering\small
\setlength{\tabcolsep}{5pt}
\begin{tabular}{lrrr}
\toprule
Detector & ML & Nature & PeerJ\\
\midrule
\logrank & 100.00 $\rightarrow$ 100.00 & 98.53 $\rightarrow$ 100.00 & 100.00 $\rightarrow$ 100.00\\
\fastdetectgpt & 100.00 $\rightarrow$ 100.00 & 69.12 $\rightarrow$ 100.00 & 89.55 $\rightarrow$ 100.00\\
\binoculars & 100.00 $\rightarrow$ 100.00 & 100.00 $\rightarrow$ 100.00 & 100.00 $\rightarrow$ 100.00\\
\radar{}$^\dagger$ & 97.98 $\rightarrow$ 100.00 & 100.00 $\rightarrow$ 100.00 & 100.00 $\rightarrow$ 100.00\\
\anchor{}$^\dagger$ & 73.74 $\rightarrow$ 99.51 & 75.00 $\rightarrow$ 100.00 & 85.07 $\rightarrow$ 100.00\\
\tf & 83.84 $\rightarrow$ 91.13 & 92.65 $\rightarrow$ 100.00 & 97.01 $\rightarrow$ 100.00\\
\mage & 98.99 $\rightarrow$ 100.00 & 100.00 $\rightarrow$ 100.00 & 100.00 $\rightarrow$ 100.00\\
\editlens & 48.99 $\rightarrow$ 83.25 & 83.82 $\rightarrow$ 100.00 & 67.16 $\rightarrow$ 98.00\\
\roberta (regular) & 3.54 $\rightarrow$ 54.19 & 4.41 $\rightarrow$ 16.00 & 49.25 $\rightarrow$ 30.00\\
\roberta (adversarial) & 0.51 $\rightarrow$ 63.05 & 0.00 $\rightarrow$ 4.00 & 2.99 $\rightarrow$ 0.00\\
\midrule
\semdetect (regular)$^\dagger$ &95.45 $\rightarrow$ 100.00 & 54.41 $\rightarrow$ 100.00 & 59.70 $\rightarrow$ 100.00\\
\peerrevealbase{}$^\dagger$ & 51.01 $\rightarrow$ 100.00 & 32.35 $\rightarrow$ 96.00 & 91.04 $\rightarrow$ 98.00\\
\peerreveal{}$^\dagger$ & 2.53 $\rightarrow$ 89.16 & 2.94 $\rightarrow$ 68.00 & 20.90 $\rightarrow$ 8.00\\
\midrule
Eligible reviews & $198\rightarrow203$ & $68\rightarrow50$ & $67\rightarrow50$\\
\bottomrule
\end{tabular}
\end{table}

\begin{table}[H]
\caption{Domain-wise evasion rates (\%) at 5\% human FPR. Each cell shows before $\rightarrow$ after training. $\dagger$ marks detector families used for generator-training feedback.}
\label{tab:generator-evasion-domain-5}
\centering\small
\setlength{\tabcolsep}{5pt}
\begin{tabular}{lrrr}
\toprule
Detector & ML & Nature & PeerJ\\
\midrule
\logrank & 100.00 $\rightarrow$ 100.00 & 98.53 $\rightarrow$ 100.00 & 100.00 $\rightarrow$ 100.00\\
\fastdetectgpt & 100.00 $\rightarrow$ 100.00 & 42.65 $\rightarrow$ 100.00 & 55.22 $\rightarrow$ 96.00\\
\binoculars & 100.00 $\rightarrow$ 100.00 & 100.00 $\rightarrow$ 100.00 & 100.00 $\rightarrow$ 100.00\\
\radar{}$^\dagger$ & 84.85 $\rightarrow$ 73.40 & 91.18 $\rightarrow$ 96.00 & 80.60 $\rightarrow$ 82.00\\
\anchor{}$^\dagger$ & 41.41 $\rightarrow$ 92.61 & 42.65 $\rightarrow$ 100.00 & 41.79 $\rightarrow$ 100.00\\
\tf & 44.95 $\rightarrow$ 67.00 & 75.00 $\rightarrow$ 100.00 & 80.60 $\rightarrow$ 94.00\\
\mage & 91.41 $\rightarrow$ 100.00 & 98.53 $\rightarrow$ 100.00 & 92.54 $\rightarrow$ 100.00\\
\editlens & 0.51 $\rightarrow$ 20.20 & 0.00 $\rightarrow$ 94.00 & 0.00 $\rightarrow$ 70.00\\
\roberta (regular) & 3.54 $\rightarrow$ 37.44 & 2.94 $\rightarrow$ 10.00 & 37.31 $\rightarrow$ 20.00\\
\roberta (adversarial) & 0.51 $\rightarrow$ 62.56 & 0.00 $\rightarrow$ 4.00 & 2.99 $\rightarrow$ 0.00\\
\midrule
\semdetect (regular)$^\dagger$ &69.19 $\rightarrow$ 100.00 & 50.00 $\rightarrow$ 100.00 & 50.75 $\rightarrow$ 100.00\\
\peerrevealbase{}$^\dagger$ & 24.75 $\rightarrow$ 99.01 & 2.94 $\rightarrow$ 88.00 & 13.43 $\rightarrow$ 90.00\\
\peerreveal{}$^\dagger$ & 0.00 $\rightarrow$ 63.05 & 0.00 $\rightarrow$ 40.00 & 0.00 $\rightarrow$ 0.00\\
\midrule
Eligible reviews & $198\rightarrow203$ & $68\rightarrow50$ & $67\rightarrow50$\\
\bottomrule
\end{tabular}
\end{table}

\subsection{\texorpdfstring{\rev{A Second Generator Family}}{A Second Generator Family}}
\label{app:second-generator}
\begin{revblock}
To test whether the framework transfers to another model family, we train GLM-4-9B-0414 on the ML domain with the same framework, reward detectors, and number of steps (67) as Qwen3.5-9B. Training is stable, and 95\% of its reviews pass the quality filter. We then repeat Stage B from the same \peerrevealbase checkpoint, replacing only the ML AI-generated training reviews with the same number of GLM reviews and keeping all other data (3 seeds).

\begin{table}[h]
\caption{Stage B with adversarial ML data from GLM-4-9B or Qwen3.5-9B (three-class Macro-F1, \%, 3 seeds).}
\label{tab:second-generator}
\centering\small
\begin{tabular}{lrrrr}
\toprule
Generator & ML & Nature & PeerJ & Pooled\\
\midrule
GLM-4-9B & 75.53 & 80.62 & 83.16 & 79.86 $\pm$ 1.36\\
Qwen3.5-9B & \textbf{77.66} & \textbf{81.03} & \textbf{84.58} & \textbf{81.17 $\pm$ 1.39}\\
\bottomrule
\end{tabular}
\end{table}

GLM recovers most of the gain over \peerrevealbase (71.03): +8.8 against +10.1 points. The gap comes mostly from ML AI-generated recall (81.0 against 89.1\%); GLM reviews are also shorter (372 against 532 words on average).
\end{revblock}

\section{Attention-Mask Ablation}
\label{app:attention-ablation}

\begin{figure}[H]
\centering
\includegraphics[width=\linewidth]{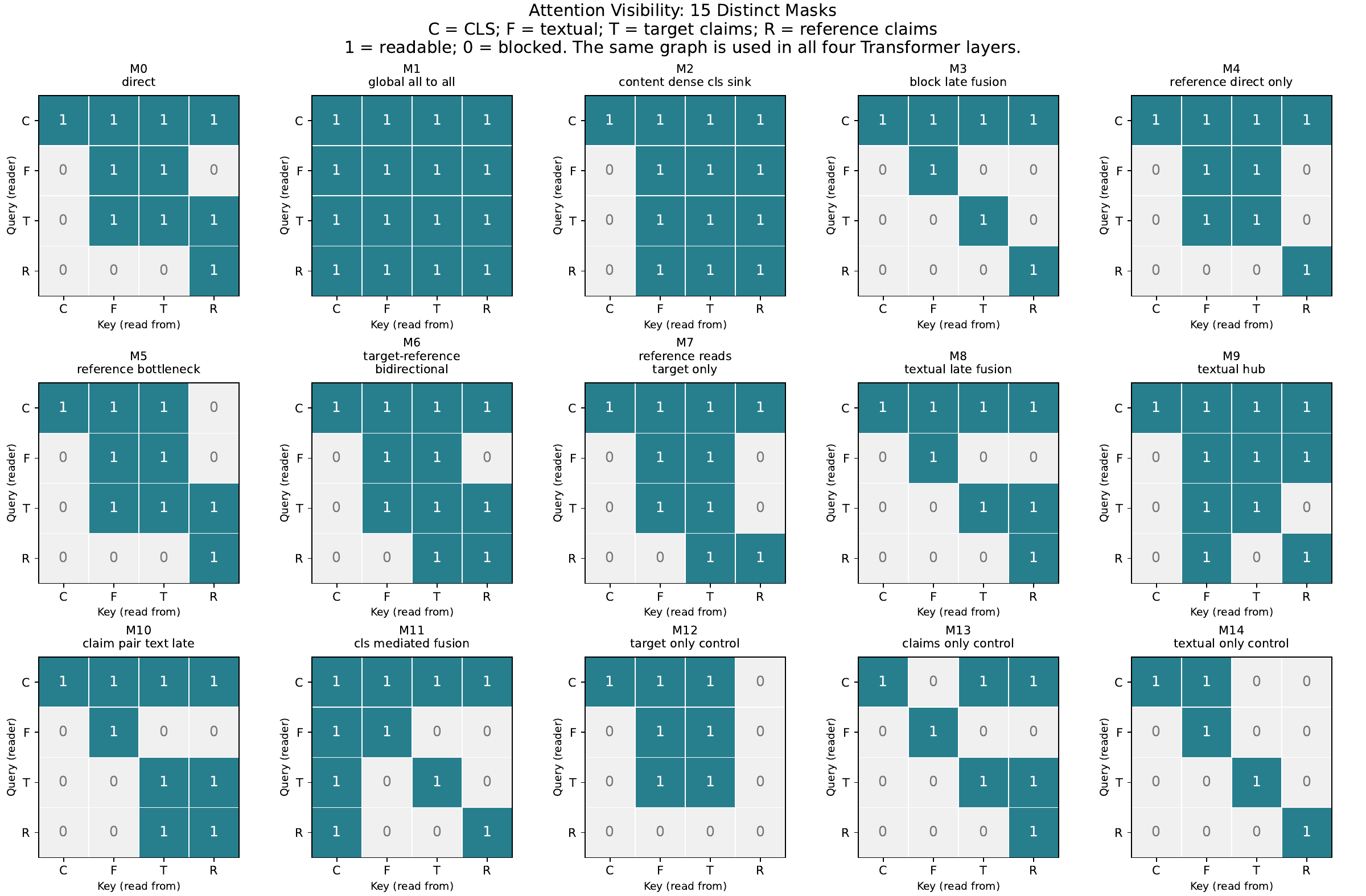}
\caption{The 15 attention masks. C denotes CLS, F textual features, T target claims, and R reference claims. Our mask appears at the upper left.}
\label{fig:attention-ablation-masks}
\end{figure}

\paragraph{Compared structures.}
Figure~\ref{fig:attention-ablation-masks} shows the 15 visibility patterns. Rows denote query blocks and columns denote key/value blocks. 1 permits attention; 0 blocks it. All four Transformer layers use the same mask. Our mask lets target claims read reference claims and exchange information with textual features, while reference claims attend within their own block. CLS reads every block. Alternatives change these connections or remove an input source while preserving model capacity.

\paragraph{Training and selection.}
Structures use matched initialization, data, sampling, losses, and training budgets. Each undergoes regular and adversarial training, with final model selection on development data. Table~\ref{tab:attention-followup} gives per-domain and pooled Macro-F1 on PeerShift.

\begin{table}[H]
\caption{Attention-mask results on PeerShift after development-set selection: Macro-F1 (\%).}
\label{tab:attention-followup}
\centering\small
\setlength{\tabcolsep}{4pt}
\begin{tabular}{llrrrr}
\toprule
ID & Attention structure & ML & Nature & PeerJ & Pooled\\
\midrule
M0 & \textbf{Our directed attention mask} & \textbf{79.95} & \textbf{81.98} & 85.18 & \textbf{82.40}\\
M1 & Fully connected attention & 78.22 & 80.00 & 82.72 & 80.39\\
M2 & Dense content attention with CLS readout & 77.82 & 78.56 & 82.24 & 79.71\\
M3 & Blockwise late fusion & 78.25 & 80.59 & 84.49 & 81.17\\
M4 & Reference access through CLS only & 79.78 & 80.18 & 82.23 & 80.78\\
M5 & Reference bottleneck & 78.30 & 79.79 & 84.01 & 80.80\\
M6 & Bidirectional target--reference attention & 78.55 & 76.62 & 81.26 & 78.85\\
M7 & Reference reads target only & 79.37 & 77.79 & 81.03 & 79.46\\
M8 & Textual late fusion & 77.06 & 80.51 & 84.50 & 80.77\\
M9 & Textual hub & 78.76 & 78.99 & 84.57 & 80.79\\
M10 & Claim-pair attention with textual late fusion & 76.79 & 78.75 & 83.82 & 79.95\\
M11 & CLS-mediated fusion & 78.25 & 81.37 & 83.21 & 81.08\\
M12 & Without reference information & 77.74 & 75.79 & 82.27 & 78.66\\
M13 & Without textual information & 77.02 & 80.57 & \textbf{85.62} & 81.12\\
M14 & Textual information only & 46.24 & 45.61 & 51.64 & 48.58\\
\bottomrule
\end{tabular}
\end{table}

We compare the directed blockwise mask with 14 alternatives under matched capacity and training; ours is best. The two evidence sources are complementary: Macro-F1 drops 3.74 points without reference claims and 1.28 without textual features, which alone reach 48.58\%.

\section{Training Data, Protocols, and Additional Results}
\label{app:data-protocols}

\subsection{Regular and Adversarial Training Corpora}
\label{app:training-corpora}

The ML regular corpus follows \semdetect. For Nature Communications and PeerJ, we collect manuscript--human-review pairs from public review records and generate corresponding regular AI reviews and faithful rewrites using conventional prompts. The PeerJ corpus is restricted to Psychology or Social Sciences. \rev{Some of these human reviews postdate the release of ChatGPT; Appendix~\ref{app:post-chatgpt} shows that they do not affect the results.} Paper-level partitions separate training and development from the final evaluation. Table~\ref{tab:training-corpora} reports the training corpora.

\begin{table}[h]
\caption{Three-domain training corpora. Both stages use regular-corpus human reviews.}
\label{tab:training-corpora}
\centering\small
\begin{tabular}{lrrrrrr}
\toprule
 & \multicolumn{4}{c}{Regular} & \multicolumn{2}{c}{Adversarial}\\
Domain & Human & AI & Rewrite & Total & AI & Rewrite\\
\midrule
ML & 2,088 & 4,580 & 8,392 & 15,060 & 5,053 & 1,259\\
Nature & 926 & 1,600 & 3,704 & 6,230 & 300 & 413\\
PeerJ & 2,075 & 4,488 & 8,299 & 14,862 & 960 & 479\\
\midrule
Total & 5,089 & 10,668 & 20,395 & 36,152 & 6,313 & 2,151\\
\bottomrule
\end{tabular}
\end{table}

The regular pools contain 541 ML, 400 Nature Communications, and 1,122 PeerJ papers. The ML regular writer pool includes DeepSeek V3.1~\citep{deepseekai2024deepseekv3technicalreport}\footnote{\url{https://huggingface.co/deepseek-ai/DeepSeek-V3.1}}, Gemini 2.5 Flash~\citep{comanici2025gemini25pushingfrontier}\footnote{\url{https://ai.google.dev/gemini-api/docs/models/gemini-2.5-flash}}, Gemini 2.5 Pro\footnote{\url{https://ai.google.dev/gemini-api/docs/models/gemini-2.5-pro}}, and Qwen3-235B~\citep{yang2025qwen3technicalreport}. The new-domain regular pool uses DeepSeek V3.2~\citep{deepseekai2025deepseekv32pushingfrontieropen}, GPT-oss-120B\footnote{\url{https://huggingface.co/openai/gpt-oss-120b}}, Mistral Large 3\footnote{\url{https://huggingface.co/mistralai/Mistral-Large-3-675B-Instruct-2512}}, and Qwen3-235B. PeerShift instead applies the frozen reasoning and reconstruction mechanisms in Appendices~\ref{app:selected-mechanisms}--\ref{app:frozen-prompts}, rather than regular training prompts on new papers.

\subsection{ML-Only Comparison}
\label{app:ml-only}
Both detectors use the original ML regular training corpus, without new-domain or adversarial data. \semdetect uses its original nine-feature LightGBM~\citep{NIPS2017_6449f44a} classifier. \peerrevealbase uses the factorized classification head, trained on the 85\% training partition and selected on the remaining 15\% development partition. The same selected checkpoint is evaluated on Original, OOD-M, OOD-M+P, and PeerShift-ML. Table~\ref{tab:ml-only-recall} gives the class recalls. Reproduction configurations accompany the code.

\begin{table}[htbp]
\caption{Class recall (\%) for the ML-only models in Table~\ref{tab:ml-only}.}
\label{tab:ml-only-recall}
\centering\small
\begin{tabular}{llrrr}
\toprule
Detector & Test set & AI & Rewrite & Human\\
\midrule
\semdetect & Original & 91.55 & 92.31 & 63.93\\
 & OOD-M & 68.86 & 82.31 & 63.69\\
 & OOD-M+P & 68.06 & 76.65 & 63.69\\
 & PeerShift-ML & 15.08 & 31.97 & 63.93\\
\midrule
\peerrevealbase & Original & 98.76 & 96.48 & 92.13\\
 & OOD-M & 94.19 & 94.52 & 93.11\\
 & OOD-M+P & 86.61 & 88.70 & 93.30\\
 & PeerShift-ML & 30.00 & 31.31 & 92.13\\
\bottomrule
\end{tabular}
\end{table}

\subsection{Three-Domain Training and Model Selection}
\label{app:model-selection}
\peerrevealbase is trained on the three-domain regular corpus; \peerreveal continues training on a mixture of regular and adversarial reviews. \semdetect is trained separately on the same corpora while retaining its nine feature definitions and reference reviews. Preprocessing is fitted on training data; model selection uses three-domain mean development Macro-F1.

Development sets use separate papers and the same generation mechanisms as PeerShift, containing 1,101 ML, 207 Nature Communications, and 255 PeerJ reviews. The mechanisms are shared between development and test; they are the frozen artifacts of the search in Appendices~\ref{app:selected-mechanisms}--\ref{app:frozen-prompts}. \rev{Appendix~\ref{app:selection-exposure} measures how much this sharing affects the results.} Final evaluation uses PeerShift, with overall metrics computed from pooled predictions. Final training configurations and execution commands accompany the code.

\subsection{Baseline Implementations}
\label{app:baseline-implementations}

Released detectors keep their published scoring models and decision procedures; only the calibration data and the evaluation set change. \logrank~\citep{ippolito-etal-2020-automatic} scores reviews with GPT-2 Medium\footnote{\url{https://huggingface.co/openai-community/gpt2-medium}}. \fastdetectgpt~\citep{ICLR2024_6b8c6f84} uses GPT-Neo-2.7B\footnote{\url{https://huggingface.co/EleutherAI/gpt-neo-2.7B}} with the analytic sampling-discrepancy estimator. \binoculars~\citep{pmlr-v235-hans24a} contrasts Falcon-7B as observer with Falcon-7B-Instruct as performer\footnote{\url{https://huggingface.co/tiiuae/falcon-7b}} over the first 512 tokens. \radar keeps its released RADAR-Vicuna-7B\footnote{\url{https://huggingface.co/TrustSafeAI/RADAR-Vicuna-7B}} weights without fine-tuning and averages scores over 512-token chunks covering the whole review. \mage~\citep{li-etal-2024-mage} uses its released detector on the whole review. \editlens keeps its released RoBERTa-large classifier over four AI-editing-extent buckets together with the authors' text cleaning; we read its AI score as the expected bucket index normalized to $[0,1]$ and average it over consecutive 510-token chunks so that the whole review is covered. \tf~\citep{kumar-etal-2024-quis} scores reviews from part-of-speech-tagged adjective, noun, and adverb frequency profiles estimated on AI and human reviews, and is the only baseline that uses no neural language model.

\anchor embeds the target review and one same-paper GPT-5.5 reference review with the same frozen Qwen3-Embedding-0.6B~\citep{qwen3embedding} encoder that \peerreveal uses, and scores their cosine similarity.

Our \roberta baseline uses RoBERTa-base with a three-class classification head on review text. Its regular and adversarial versions use the same respective training corpora as \peerrevealbase and \peerreveal.

\subsection{Per-Class Results}
\label{app:per-class-results}

\begin{table}[h]
\caption{PeerShift class recall (\%) for \peerrevealbase and \peerreveal (Table~\ref{tab:three-domain-main}).}
\label{tab:class-recall}
\centering\small
\begin{tabular}{llrrr}
\toprule
Training & Domain & AI & Rewrite & Human\\
\midrule
\textsc{Base} & ML & 56.89 & 30.16 & 93.77\\
 & Nature & 72.00 & 72.00 & 76.67\\
 & PeerJ & 69.33 & 87.50 & 74.67\\
\textsc{Full} & ML & 93.77 & 62.62 & 84.43\\
 & Nature & 91.67 & 75.50 & 78.83\\
 & PeerJ & 88.67 & 85.17 & 81.33\\
\bottomrule
\end{tabular}
\end{table}

Rows and columns of the pooled confusion matrices follow AI, rewrite, human:
\begin{equation}
C_{\textsc{Base}}=\begin{bmatrix}1195&539&76\\120&1141&549\\32&298&1480\end{bmatrix},\qquad
C_{\textsc{Full}}=\begin{bmatrix}1654&145&11\\152&1346&312\\31&303&1476\end{bmatrix}.
\end{equation}
Their Macro-F1 scores are 70.52\% and 82.40\%.

Tables~\ref{tab:recall-by-writer} and~\ref{tab:recall-by-mechanism} break the AI and rewrite recalls down by writer model and by mechanism (Appendix~\ref{app:selected-mechanisms}). \peerreveal matches or improves \peerrevealbase for most writer-model and mechanism groups, with the largest gains where \peerrevealbase was weakest: Grok 4.3 AI reviews rise from 1.64\%, 6.67\%, and 18.33\% to 86.89\%, 80.00\%, and 76.67\% on ML, Nature Communications, and PeerJ, and ML rewrites by Grok 4.3 and DeepSeek V4 Pro from 14.75\% and 14.75\% to 61.48\% and 65.57\%. For AI-generated reviews, recall varies more across writer models than across mechanisms within a domain; the hardest AI writers for \peerreveal are MiniMax M2.5 on ML (77.05\%), MiniMax M2.5 on Nature Communications (76.67\%), and Gemini 3.1 Pro Preview on PeerJ (70.00\%).

\begin{table}[htbp]
\caption{PeerShift recall (\%) of \peerrevealbase (\textsc{Base}) and \peerreveal (\textsc{Full}) by writer model. Each AI cell covers 61 (ML) or 60 (Nature, PeerJ) reviews and each rewrite cell 122 or 120.}
\label{tab:recall-by-writer}
\centering\small
\setlength{\tabcolsep}{4pt}
\begin{tabular}{llrrrrrr}
\toprule
 & & \multicolumn{2}{c}{ML} & \multicolumn{2}{c}{Nature} & \multicolumn{2}{c}{PeerJ}\\
Class & Writer model & \textsc{Base} & \textsc{Full} & \textsc{Base} & \textsc{Full} & \textsc{Base} & \textsc{Full}\\
\midrule
AI & Gemini 3.1 Pro Preview & 37.70 & 93.44 & 31.67 & 83.33 & 33.33 & 70.00\\
 & DeepSeek V4 Pro & 75.41 & 98.36 & 86.67 & 100.00 & 75.00 & 95.00\\
 & MiniMax M2.5 & 40.98 & 77.05 & 65.00 & 76.67 & 73.33 & 76.67\\
 & Mistral Large 3 (675B) & 96.72 & 100.00 & 91.67 & 83.33 & 88.33 & 93.33\\
 & Qwen3-Next-80B-A3B & 98.36 & 100.00 & 98.33 & 98.33 & 96.67 & 100.00\\
 & GLM-5 & 39.34 & 95.08 & 66.67 & 95.00 & 48.33 & 83.33\\
 & Kimi K2.5 & 70.49 & 100.00 & 85.00 & 100.00 & 75.00 & 96.67\\
 & Grok 4.3 & 1.64 & 86.89 & 6.67 & 80.00 & 18.33 & 76.67\\
 & Claude Sonnet 4.6 & 68.85 & 91.80 & 90.00 & 100.00 & 86.67 & 96.67\\
 & GPT-5.5 & 39.34 & 95.08 & 98.33 & 100.00 & 98.33 & 98.33\\
\midrule
Rewrite & Gemini 3.5 Flash & 59.84 & 73.77 & 80.00 & 84.17 & 86.67 & 84.17\\
 & DeepSeek V4 Pro & 14.75 & 65.57 & 73.33 & 74.17 & 85.83 & 85.00\\
 & Kimi K2.5 & 50.00 & 76.23 & 80.00 & 72.50 & 93.33 & 86.67\\
 & Grok 4.3 & 14.75 & 61.48 & 68.33 & 79.17 & 85.00 & 86.67\\
 & GPT-5.5 & 11.48 & 36.07 & 58.33 & 67.50 & 86.67 & 83.33\\
\bottomrule
\end{tabular}
\end{table}

\begin{table}[htbp]
\caption{PeerShift recall (\%) of \peerrevealbase (\textsc{Base}) and \peerreveal (\textsc{Full}) by mechanism; IDs follow Tables~\ref{tab:challenge-ai-mechanisms} and~\ref{tab:challenge-rewrite-mechanisms}.}
\label{tab:recall-by-mechanism}
\centering\small
\setlength{\tabcolsep}{4pt}
\begin{tabular}[t]{lrrr}
\toprule
AI mechanism & Reviews & \textsc{Base} & \textsc{Full}\\
\midrule
ML-A1 & 152 & 55.26 & 94.08\\
ML-A2 & 152 & 56.58 & 96.05\\
ML-A3 & 153 & 56.21 & 91.50\\
ML-A4 & 153 & 59.48 & 93.46\\
NC-A1 & 150 & 74.00 & 90.00\\
NC-A2 & 150 & 73.33 & 92.67\\
NC-A3 & 150 & 68.67 & 90.00\\
NC-A4 & 150 & 72.00 & 94.00\\
PJ-A1 & 200 & 71.50 & 93.00\\
PJ-A2 & 200 & 67.00 & 88.00\\
PJ-A3 & 200 & 69.50 & 85.00\\
\bottomrule
\end{tabular}
\hspace{1.5em}
\begin{tabular}[t]{lrrr}
\toprule
Rewrite mechanism & Reviews & \textsc{Base} & \textsc{Full}\\
\midrule
ML-R1 & 102 & 21.57 & 58.82\\
ML-R2 & 102 & 40.20 & 69.61\\
ML-R3 & 102 & 22.55 & 57.84\\
ML-R4 & 102 & 35.29 & 62.75\\
ML-R5 & 101 & 27.72 & 63.37\\
ML-R6 & 101 & 33.66 & 63.37\\
NC-R1 & 100 & 62.00 & 67.00\\
NC-R2 & 100 & 62.00 & 74.00\\
NC-R3 & 100 & 89.00 & 84.00\\
NC-R4 & 100 & 70.00 & 75.00\\
NC-R5 & 100 & 82.00 & 82.00\\
NC-R6 & 100 & 67.00 & 71.00\\
PJ-R1 & 120 & 93.33 & 85.83\\
PJ-R2 & 120 & 76.67 & 75.83\\
PJ-R3 & 120 & 88.33 & 88.33\\
PJ-R4 & 120 & 88.33 & 84.17\\
PJ-R5 & 120 & 90.83 & 91.67\\
\bottomrule
\end{tabular}
\end{table}

\subsection{Additional Binary Results}
\label{app:binary-results}
Three-class models contribute $P(\mathrm{AI})$, and ROC curves and AUROC are computed from continuous scores. Binary Macro-F1 uses discrete predictions. Three-class detectors predict AI when $P(\mathrm{AI})\geq.5$, without threshold tuning. A binary Macro-F1 near 40\% corresponds to predicting almost no review as AI-generated; regular \semdetect reaches this value on PeerShift (Table~\ref{tab:binary-main}) because its $P(\mathrm{AI})$ rarely exceeds .5 under this rule, despite an AUROC of 77.87\%. Every other method uses one global threshold that maximizes binary Macro-F1 on a calibration panel of 2,665 regular-corpus reviews from 145 papers (48 ML, 50 Nature Communications, and 47 PeerJ; 780 AI-generated, 1,508 AI-rewritten, and 377 human-written), none of which shares a paper with PeerShift; ties are resolved toward the higher threshold. Table~\ref{tab:macro-f1-thresholds} lists the resulting thresholds. ROC operating points use both human-written and AI-rewritten reviews as negatives. We report the highest empirical TPR with FPR at or below each stated limit, retaining complete score ties without interpolation. All 14 evaluated configurations cover the same 5,430 reviews (1,810 positives and 3,620 negatives).

\begin{table}[htbp]
\caption{Binary Macro-F1 decision thresholds, selected on the 2,665-review calibration panel. A review is predicted AI-generated when its score reaches the threshold.}
\label{tab:macro-f1-thresholds}
\centering\small
\begin{tabular}{lrr}
\toprule
Method & Threshold & Calibration Macro-F1 (\%)\\
\midrule
\logrank & $-1.1281$ & 41.55\\
\fastdetectgpt & $1.4018$ & 56.70\\
\binoculars & $-0.9472$ & 49.89\\
\radar & $0.6339$ & 57.41\\
\anchor & $0.8735$ & 86.61\\
\tf & $0.2915$ & 54.04\\
\mage & $0.9998$ & 41.43\\
\editlens & $0.4292$ & 56.63\\
\bottomrule
\end{tabular}
\end{table}

\begin{figure}[t]
\centering
\includegraphics[width=\linewidth]{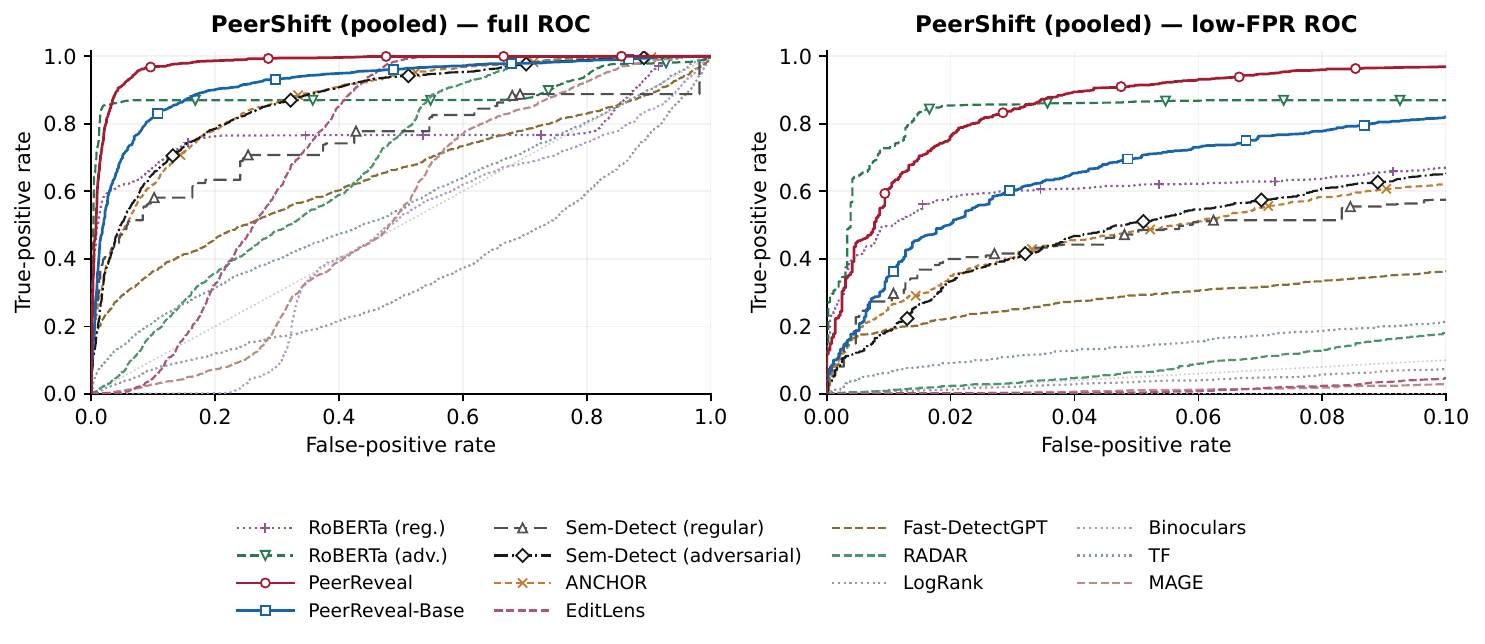}
\caption{Pooled binary ROC curves on PeerShift. Left: full ROC. Right: FPR from 0 to 0.1.}
\label{fig:binary-roc-pooled}
\end{figure}

\begin{figure}[p]
\centering
\includegraphics[width=\linewidth]{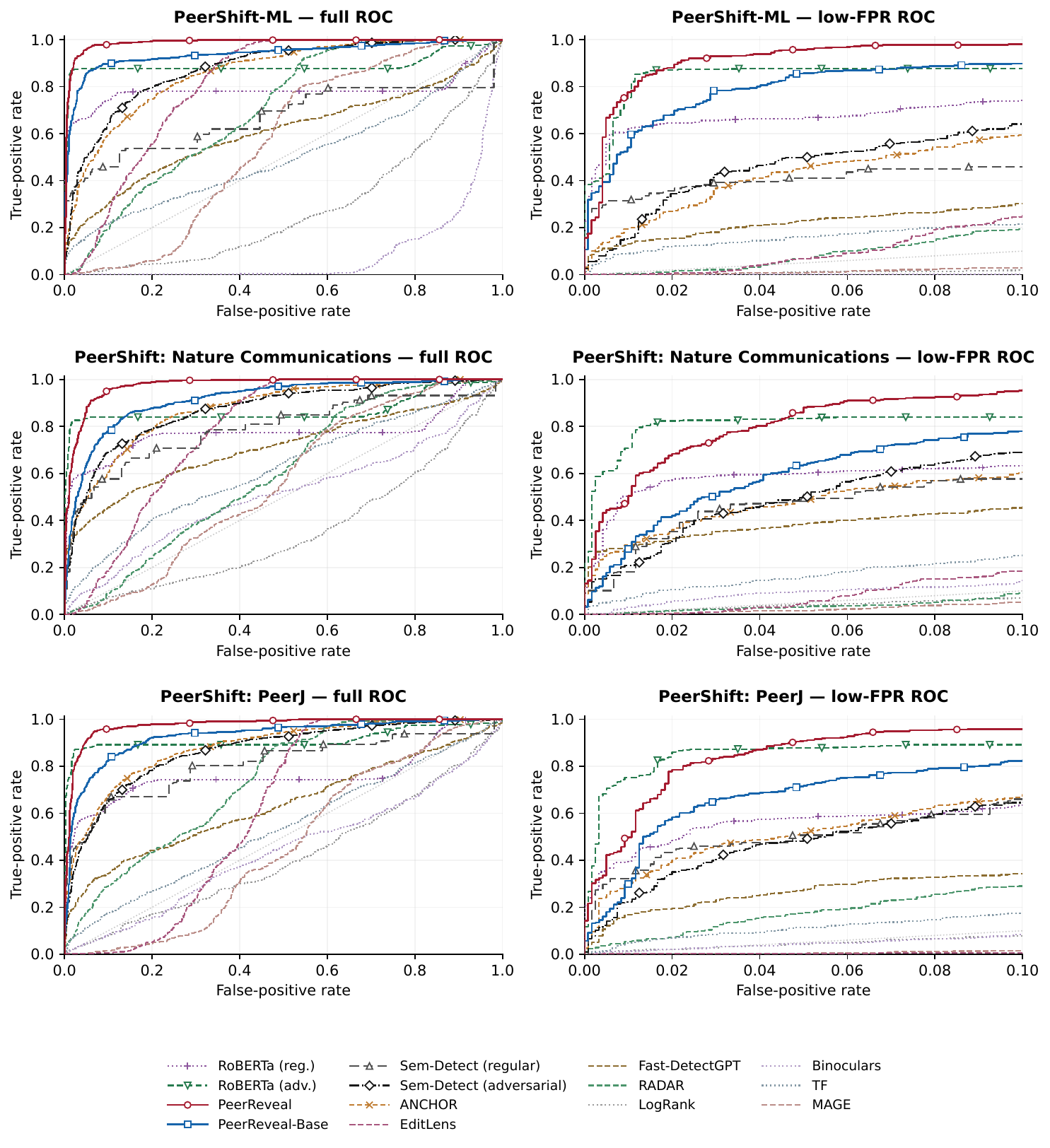}
\caption{Binary ROC curves on PeerShift for the same detectors evaluated on ML (top), Nature Communications (middle), and PeerJ (bottom). Left: full ROC. Right: FPR from 0 to 0.1.}
\label{fig:binary-roc-domains}
\end{figure}

Table~\ref{tab:binary-domain} gives the complete per-domain results. Figures~\ref{fig:binary-roc-pooled} and~\ref{fig:binary-roc-domains} show full and low-FPR ROC curves for PeerShift (pooled) and each domain.
\begin{table}[h]
\caption{Per-domain binary results on PeerShift (\%): AUROC / TPR@1\% / Macro-F1.}
\label{tab:binary-domain}
\centering\small
\begin{tabular}{p{3.4cm}ccc}
\toprule
Detector & ML & Nature & PeerJ\\
\midrule
\logrank & 27.88 / 0.00 / 39.98 & 35.11 / 0.83 / 39.96 & 41.39 / 1.50 / 40.16\\
\fastdetectgpt & 60.74 / 13.11 / 58.45 & 71.01 / 28.50 / 69.12 & 63.37 / 16.83 / 62.83\\
\binoculars & 8.97 / 0.00 / 25.61 & 50.99 / 3.00 / 54.81 & 44.07 / 1.33 / 49.38\\
\radar & 69.84 / 0.49 / 60.87 & 60.05 / 0.50 / 53.10 & 73.40 / 5.50 / 64.16\\
\anchor & 86.17 / 19.51 / 73.72 & 87.53 / 29.67 / 73.04 & 88.19 / 30.50 / 75.63\\
\tf & 50.45 / 9.02 / 51.07 & 61.77 / 7.00 / 52.05 & 53.76 / 4.67 / 49.81\\
\mage & 56.14 / 0.00 / 39.98 & 55.97 / 0.17 / 40.00 & 46.19 / 0.17 / 40.00\\
\editlens & 80.70 / 0.00 / 66.18 & 77.96 / 0.17 / 67.33 & 61.02 / 0.17 / 55.86\\
\roberta (reg.) & 78.68 / 61.97 / 71.14 & 77.70 / 50.50 / 66.61 & 76.48 / 39.67 / 64.98\\
\roberta (adv.) & 89.07 / 75.41 / 85.17 & 86.93 / 71.00 / 81.19 & 91.22 / 75.17 / 84.81\\
\semdetect (reg.) & 71.62 / 31.48 / 40.00 & 81.16 / 18.00 / 40.00 & 82.38 / 32.17 / 40.00\\
\semdetect (adv.) & 87.78 / 15.57 / 72.96 & 87.14 / 20.83 / 77.16 & 86.72 / 22.17 / 73.49\\
\peerrevealbase & 94.02 / 54.10 / 80.39 & 91.60 / 27.83 / 83.64 & 92.84 / 29.83 / 84.12\\
\peerreveal & 98.78 / 77.54 / \textbf{94.73} & 97.55 / 50.50 / \textbf{91.79} & 97.61 / 49.83 / \textbf{92.51}\\
\bottomrule
\end{tabular}
\end{table}

\FloatBarrier
\subsection{Human-versus-AI Binary Results}
\label{app:human-vs-ai}

Table~\ref{tab:human-vs-ai} repeats the binary evaluation with AI-rewritten reviews removed, so that fully AI-generated reviews are compared with human-written reviews only, as in the binary setting of \semdetect. AUROC and TPR at 1\% FPR use the same scores as Table~\ref{tab:binary-main}. For Macro-F1, three-class detectors keep the $P(\mathrm{AI})\geq.5$ rule, and every other method uses a threshold re-selected by the same rule on the 780 AI-generated and 377 human-written reviews of the calibration panel. \peerreveal has the highest AUROC (99.23\%) and TPR at 1\% FPR (87.90\%). \editlens achieves 98.04\% AUROC and the highest Macro-F1 (96.07\%), but falls to 73.11\% AUROC with rewrites included as negatives (Table~\ref{tab:binary-main}).

\begin{table}[H]
\caption{Pooled human-versus-AI binary detection on PeerShift (\%). Positive: AI-generated (1,810); negative: human-written (1,810). AI-rewritten reviews are excluded.}
\label{tab:human-vs-ai}
\centering\small
\begin{tabular}{lrrr}
\toprule
Detector & AUROC & TPR@1\% & Macro-F1\\
\midrule
\logrank & 32.90 & 0.66 & 34.45\\
\fastdetectgpt & 66.63 & 20.17 & 62.40\\
\binoculars & 47.11 & 0.00 & 50.34\\
\radar & 77.56 & 4.97 & 66.12\\
\anchor & 86.01 & 21.60 & 77.25\\
\tf & 65.62 & 9.12 & 60.43\\
\mage & 55.22 & 0.17 & 56.35\\
\editlens & 98.04 & 37.35 & \textbf{96.07}\\
\midrule
\roberta (regular) & 75.82 & 59.12 & 62.42\\
\roberta (adversarial) & 86.97 & 85.64 & 80.38\\
\semdetect (regular) & 79.28 & 36.80 & 33.33\\
\semdetect (adversarial) & 86.65 & 14.86 & 70.85\\
\peerrevealbase & 95.41 & 58.84 & 81.56\\
\peerreveal & \textbf{99.23} & \textbf{87.90} & 94.77\\
\bottomrule
\end{tabular}
\end{table}

\subsection{Cost and Compute}
\label{app:cost-compute}

Constructing PeerShift is dominated by review generation rather than by detector training. Generating its 3,620 AI-generated and AI-rewritten reviews through API providers cost about \$600, roughly \$0.17 per review. Detection is inexpensive. Each target review needs one claim-extraction call, which costs about \$0.0026 per review with Gemini 2.5 Flash (\$0.0013 with batch calls). The three references are generated once per manuscript and rating and shared by every review of that manuscript at that rating; at public Amazon Bedrock prices they cost about \$0.02--\$0.04 per manuscript and rating, plus three extraction calls. The claim encoder, the reference language model, and the detector run locally, and all features are computed once and reused by \peerrevealbase, \peerreveal, the \semdetect comparisons, and every training stage.

Generator training is the largest GPU cost. The final generators are reached after 67 (ML), 39 (Nature Communications), and 23 (PeerJ) GRPO updates, each sampling 48 complete reviews and scoring them with four frozen detectors and the frozen quality scorers, for $3{,}216$, $1{,}872$, and $1{,}104$ generated reviews respectively. \peerreveal training is negligible by comparison: with $300{,}295$ trainable parameters over precomputed features, Stage A runs 30 epochs of 406 steps and Stage B 20 epochs of 406 steps, both at batch size 64 on a single GPU.

\subsection{\texorpdfstring{\rev{External Benchmarks}}{External Benchmarks}}
\label{app:external-benchmarks}
\begin{revblock}
We evaluate all detectors, unchanged, on two recent benchmarks built by other groups. \citet{saha2026policies} polish and regenerate pre-ChatGPT PeerRead reviews (ACL 2017, CoNLL 2016, ICLR 2017, NeurIPS 2013--17) with five LLMs; we use its hard subset of 158 papers. PeerPrism~\citep{sadeghian2026peerprism} rewrites and generates reviews of ICLR and NeurIPS 2021--24 papers with six LLMs; we drop 7 papers that overlap with our data, leaving 153. None of these writer models is in our training data.

Human reviews map to human, polished or rewritten reviews to AI-rewritten, and fully generated reviews to AI-generated. Mixed categories (AI-HI in \citet{saha2026policies}; the three hybrid types in PeerPrism) are excluded. Macro-F1 is computed on 4,560 reviews for \citet{saha2026policies} (1,570 AI-generated, 2,487 AI-rewritten, 503 human) and on 2,618 for PeerPrism (986, 994, 638). References are generated as for Nature Communications and PeerJ (Appendix~\ref{app:reference-reviews}).

Score-based detectors output one AI-likeness score. We split it into three classes with two cut points that maximize three-class Macro-F1 on the calibration panel of Appendix~\ref{app:binary-results}; the external sets are never used to set them.

\begin{table}[h]
\caption{Three-class Macro-F1 (\%) on the external benchmarks. $^\dagger$Score-based detector with two cut points set on our calibration panel.}
\label{tab:external-benchmarks}
\centering\small
\begin{tabular}{lrr}
\toprule
Detector & \citet{saha2026policies} & PeerPrism\\
\midrule
\textbf{\peerreveal} & \textbf{67.79} & \textbf{78.16}\\
\peerrevealbase & 62.87 & 75.91\\
\roberta (reg.) & 66.59 & 72.88\\
\roberta (adv.) & 66.48 & 77.84\\
\semdetect (reg.) & 52.67 & 56.68\\
\semdetect (adv.) & 50.36 & 55.28\\
\midrule
\editlens$^\dagger$ & 64.12 & 60.94\\
\tf$^\dagger$ & 55.48 & 47.37\\
\fastdetectgpt$^\dagger$ & 54.90 & 32.76\\
\anchor$^\dagger$ & 45.79 & 47.61\\
\binoculars$^\dagger$ & 43.33 & 32.79\\
\logrank$^\dagger$ & 38.03 & 26.13\\
\mage$^\dagger$ & 30.44 & 30.26\\
\radar$^\dagger$ & 28.41 & 34.75\\
\bottomrule
\end{tabular}
\end{table}

\peerreveal is best on both benchmarks. Its recall of AI-generated, AI-rewritten, and human reviews is 48.0, 82.5, and 85.7\% on \citet{saha2026policies}, and 70.9, 82.4, and 79.3\% on PeerPrism. On \citet{saha2026policies}, half of the AI-generated reviews are labeled AI-rewritten, so most errors stay on the AI side.
\end{revblock}

\subsection{\texorpdfstring{\rev{Development-Set Exposure and External Reviewing Agents}}{Development-Set Exposure and External Reviewing Agents}}
\label{app:selection-exposure}
\begin{revblock}
\paragraph{Development-set exposure.} PeerShift's generation mechanisms never appear in training, but the development set used for checkpoint selection shares them (Appendix~\ref{app:model-selection}). Table~\ref{tab:selection-exposure} measures how much this matters. Without any development data, \peerreveal loses at most about 1.4 points. Removing a mechanism or a writer from the development set almost never changes the selected checkpoint.

\begin{table}[h]
\caption{Effect of the development set on checkpoint selection (pooled three-class Macro-F1, \%).}
\label{tab:selection-exposure}
\centering\small
\begin{tabular}{lr}
\toprule
Setting & Result\\
\midrule
Development-based selection (5 seeds) & 81.46 $\pm$ 1.10\\
No development data: last epoch in both stages (5 seeds) & 80.08 $\pm$ 0.83\\
Each ML mechanism removed from development, selection re-run & same checkpoint in 40/40\\
Each writer removed from development, selection re-run & same checkpoint in 57/60\\
\bottomrule
\end{tabular}
\end{table}

\paragraph{External reviewing agents.}
\label{app:external-agents}
To test unseen real review workflows, we run two public frontier LLM reviewing agents, Google Deep Review and the Stanford Agentic Reviewer (Section~\ref{sec:beyond-peershift}). Neither appears anywhere in our pipeline. They reviewed papers drawn from a random sample of PeerShift-ML test papers: 10 for Deep Review (Gemini 4 Argon, medium depth) and 25 for the Agentic Reviewer.

We keep the part of each output that corresponds to a peer review. A Deep Review report opens with a high-level summary (summary, strengths, and weaknesses) in the form of a conference review, followed by a section-by-section audit of the paper with page references. The audit makes up most of the report: for one paper, the full report has about 71k characters, against about 5k for the high-level summary. We therefore use the high-level summary. From the Agentic Reviewer, we drop the scoring rubric, which contains scores rather than review text. Claims and references are produced as for PeerShift.

\peerreveal labels all 35 reviews as AI-generated (Table~\ref{tab:external-agents}). The other detectors label most Deep Review reviews as AI-rewritten.
\end{revblock}

\subsection{\texorpdfstring{\rev{Held-Out Domains and Writer Families}}{Held-Out Domains and Writer Families}}
\label{app:held-out-domains}
\begin{revblock}
We retrain \peerreveal and \peerrevealbase with one domain removed from training and development (3 seeds) and test on that domain (Table~\ref{tab:lodo}). An unseen domain costs \peerreveal 6--9 points, but adversarial training still helps there (+2.5--8.7 over \peerrevealbase), and \peerreveal still beats \semdetect trained on all three domains, including the held-out one.

\begin{table}[h]
\caption{Three-class Macro-F1 (\%) on a domain held out from training and development (3 seeds). ``(all)'': trained on all three domains; \semdetect uses augmented training (Table~\ref{tab:three-domain-main}).}
\label{tab:lodo}
\centering\small
\begin{tabular}{lrrrr}
\toprule
Held-out domain & \peerreveal & \peerrevealbase & \peerreveal (all) & \semdetect (all)\\
\midrule
ML & \textbf{68.8} & 60.1 & 77.8 & 51.5\\
Nature & \textbf{74.9} & 72.4 & 80.9 & 55.6\\
PeerJ & \textbf{76.7} & 68.9 & 84.8 & 60.2\\
\bottomrule
\end{tabular}
\end{table}

We also retrain with one writer family (Gemini, DeepSeek, Mistral, GPT, or Qwen) removed from training and development. Of that family's PeerShift reviews, 98.7--100\% are still labeled AI-generated or AI-rewritten, compared with 99.1--100\% with full training. Pooled Macro-F1 stays at 79.4--81.5 (81.2 with full training).
\end{revblock}

\subsection{\texorpdfstring{\rev{Control with Ordinary Generator Data}}{Control with Ordinary Generator Data}}
\label{app:ordinary-control}
\begin{revblock}
To test whether the gain comes from adversarial training or simply from more data, the untrained Qwen3.5-9B generates the same amount of ordinary AI reviews and rewrites, with the conventional prompts of the regular corpora (Appendix~\ref{app:training-corpora}). These replace the adversarial data in Stage B, starting from the same \peerrevealbase checkpoint (3 seeds). Pooled Macro-F1 reaches 72.42 $\pm$ 0.65, barely above \peerrevealbase (71.03), against 81.17 $\pm$ 1.39 with the adversarial data (Table~\ref{tab:ordinary-control}).

\begin{table}[h]
\caption{Stage B with ordinary or adversarial data from the same Qwen3.5-9B model (three-class Macro-F1, \%, 3 seeds).}
\label{tab:ordinary-control}
\centering\small
\begin{tabular}{lrrrr}
\toprule
Stage B data & ML & Nature & PeerJ & Pooled\\
\midrule
Ordinary (untrained generator) & 64.24 & 74.24 & 77.99 & 72.42\\
Adversarial (trained generator) & \textbf{77.66} & \textbf{81.03} & \textbf{84.58} & \textbf{81.17}\\
\bottomrule
\end{tabular}
\end{table}
\end{revblock}

\subsection{\texorpdfstring{\rev{Input Ablation over Five Seeds}}{Input Ablation over Five Seeds}}
\label{app:input-ablation}
\begin{revblock}
Each variant is retrained from Stage A with five seeds (Table~\ref{tab:input-ablation}). Removing the references and text statistics, or taking the references from a different paper, costs about 1.6 points. Both variants remain well above \roberta trained on the same data.

\begin{table}[h]
\caption{Pooled three-class Macro-F1 (\%) of \peerreveal input variants.}
\label{tab:input-ablation}
\centering\small
\begin{tabular}{lr}
\toprule
Model & Macro-F1\\
\midrule
\peerreveal, all inputs (5 seeds) & \textbf{81.46 $\pm$ 1.10}\\
Target review only: no references, no text statistics (5 seeds) & 79.76 $\pm$ 1.39\\
References from a different paper (5 seeds) & 79.88 $\pm$ 0.46\\
\roberta (adv.), same data & 73.04\\
\peerrevealbase (5 seeds) & 71.03 $\pm$ 0.80\\
\bottomrule
\end{tabular}
\end{table}
\end{revblock}

\subsection{\texorpdfstring{\rev{Hybrid Authorship}}{Hybrid Authorship}}
\label{app:hybrid}
\begin{revblock}
For 558 papers (158 ML, 200 Nature Communications, 200 PeerJ), we start from the paper's AI-rewritten test review (0\%) and replace 25, 50, or 75\% of its claims with claims from the paper's AI-generated test review (100\%). Claude selects compatible claims and integrates them so that the review stays coherent.

\begin{table}[h]
\caption{Share of reviews labeled AI-generated (\%) as the AI-generated share of claims grows.}
\label{tab:hybrid}
\centering\small
\begin{tabular}{lrrrrr}
\toprule
AI-generated share & 0\% & 25\% & 50\% & 75\% & 100\%\\
\midrule
\peerreveal & 7.0 & 17.6 & 30.8 & 47.6 & 93.5\\
\bottomrule
\end{tabular}
\end{table}

\peerreveal labels more reviews as AI-generated as the share grows (Table~\ref{tab:hybrid}). The other detectors barely respond; \roberta, for example, labels only 0--7.5\% of the 0--75\% reviews as AI-generated.
\end{revblock}

\subsection{\texorpdfstring{\rev{Human Reviews after the Release of ChatGPT}}{Human Reviews after the Release of ChatGPT}}
\label{app:post-chatgpt}
\begin{revblock}
All ML human reviews come from ICLR and NeurIPS 2021--22, before ChatGPT's release on November 30, 2022. We dated the Nature Communications and PeerJ training reviews by their Crossref publication dates: 351 and 1,284 predate the release, while 572 and 775 come after it. Fewer public reviews exist before the release, and \citet{pmlr-v235-liang24b} find no significant increase of AI-modified content in Nature portfolio reviews. Retraining without any post-release human review gives 81.25 $\pm$ 0.60 (3 seeds), essentially identical to the full model with the same seeds (81.29). The retrained model reaches 81.50 on the pre-release part of the test set.
\end{revblock}

\section{Detector Implementation Details}
\label{app:detector-implementation}

This appendix specifies the preprocessing pipeline, architecture, and optimization settings for the three-domain \peerreveal models. Claim extraction, reference selection, and claim embedding are fixed across \peerrevealbase, \peerreveal, and \semdetect.

\subsection{Claim Extraction}
\label{app:claim-extraction}

Each review is decomposed into functionally labeled claims by an LLM extractor; as in \semdetect, this granularity follows retrieval work showing that the unit of text representation strongly affects downstream performance~\citep{chen-etal-2024-dense,duarte-etal-2024-lumberchunker}. Reviews taken from the \semdetect release, namely the ML regular corpus and the human-written reviews of PeerShift-ML, keep the claims released with it, which were extracted with Gemini-2.5-Flash; all other reviews are extracted by Gemini 2.5 Flash. A JSON schema requiring one string array per category is supplied in the prompt text; decoding is unconstrained. The output budget is $4{,}096$ tokens, with temperature $0.1$ and thinking disabled. The extractor sees only the review text; it never sees the manuscript, the writer model, or the class label. The system prompt is:
\begin{quote}\small
You are given the full text of a peer review. Your task is to extract and organize the reviewer's comments into bullet points under the following categories:\\
1. \textbf{Factual Restatement} -- Summaries or descriptions of what the paper does, its methods, contributions, or results.\\
2. \textbf{Evaluation} -- Judgments of quality, including both strengths (positive evaluations) and weaknesses/limitations (negative evaluations).\\
3. \textbf{Constructive Input} -- Actionable suggestions or recommendations for improvement.\\
4. \textbf{Clarification Dialogue} -- Questions directed to the authors or requests for clarification.\\
5. \textbf{Meta-Commentary} -- Remarks about the broader context, such as fit for the venue, clarity of writing, novelty, or overall recommendation.\\
Instructions:\\
-- For each category, extract bullet points from the review text that match its definition.\\
-- Keep each bullet point focused on a single coherent idea, but \textbf{do not break ideas into fragments}---preserve the full meaning so the point is self-contained and understandable.\\
-- Maintain the reviewer's original phrasing where possible, lightly compressing or restructuring only for clarity.\\
-- If the review contains no content for a category, leave it blank.\\
-- Do not add information that is not present in the review.
\end{quote}

Following \semdetect, we retain evaluation, constructive-input, and clarification-dialogue claims and discard factual restatements and meta-commentary. Claims follow the category order above and extractor order within each category.

Table~\ref{tab:claim-counts} reports mean claim counts per category on the 5,430 PeerShift reviews. AI-generated reviews carry the most evaluation claims (18.3 against 6.9 for human reviews), while human reviews carry the most constructive-input claims (7.2 against 4.5). Rewrites are the shortest in claim terms. Appendix~\ref{app:examples} reproduces one paper's three reviews in full.

\begin{table}[h]
\caption{Mean claims per review by category on PeerShift, with the three retained categories marked. Retained counts are computed before the 128-claim cap.}
\label{tab:claim-counts}
\centering\small
\setlength{\tabcolsep}{4pt}
\begin{tabular}{lrrrrrrr}
\toprule
 & Factual & Evaluation$^{\ast}$ & Constructive$^{\ast}$ & Clarification$^{\ast}$ & Meta & All & Retained\\
Class & restatement & & input & dialogue & commentary & & \\
\midrule
AI & 13.3 & 18.3 & 4.5 & 3.6 & 2.6 & 42.3 & 26.4\\
Rewrite & 4.5 & 8.5 & 6.0 & 2.3 & 1.6 & 22.9 & 16.8\\
Human & 6.3 & 6.9 & 7.2 & 4.1 & 1.5 & 26.0 & 18.2\\
\bottomrule
\end{tabular}
\end{table}

\subsection{Reference Reviews}
\label{app:reference-reviews}

Each target review is paired with three AI reviews of the same manuscript, generated in advance with conventional prompts and matched to its requested rating when available. References are selected once, before any training, and the selected triple is frozen: it does not change between \peerrevealbase and \peerreveal, between training stages, or between \peerreveal and \semdetect.

For PeerShift-ML, the reference pool comprises Gemini 2.5 Flash, Gemini 2.5 Pro, DeepSeek V3.1, and Qwen3-235B-A22B, with three references selected per target. PeerShift's Nature Communications and PeerJ subsets use one reference each from Qwen3-235B-A22B, DeepSeek V3.2, and Mistral Large 3. These PeerShift references use \semdetect's base prompt: given the manuscript and requested recommendation, it asks for a concise, critical, constructive review with a paper summary, strengths, weaknesses, questions for the authors, and a final judgement. No searched mechanism is appended. The new-domain training corpora select three references from their regular AI pool, which additionally includes GPT-oss-120B; these references do not condition on a requested rating.

The three references are concatenated into one pooled block in the frozen selection order, using the same retained categories as the target. The pooled block contains 61.8 claims per review on average (median 61, 10th percentile 50, 90th percentile 74, maximum 96), so the 128-position reference cap in Table~\ref{tab:detector-hparams} is never reached on any evaluated review.

\subsection{Claim Encoder and Textual Features}
\label{app:claim-encoder}

Retained claims are encoded by a frozen Qwen3-Embedding-0.6B sentence encoder in half precision with a batch size of 32 and left-side padding, producing one $1{,}024$-dimensional vector per claim. The encoder is never fine-tuned; the only learned transformation of claim content is the shared linear projection into the 64-dimensional model space described in Section~\ref{sec:unified-detector}. \semdetect uses the same frozen encoder, so the gap between the two systems isolates how claim relationships are modeled.

The four review-level textual features are computed on the target review only, with Mistral-7B-Instruct-v0.3~\citep{jiang2023mistral7b}\footnote{\url{https://huggingface.co/mistralai/Mistral-7B-Instruct-v0.3}} in half precision as the reference language model and a maximum length of $4{,}096$ tokens: mean token perplexity, the \fastdetectgpt sampling discrepancy estimated from $10{,}000$ samples on the first 800 tokens, the percentage of tokens inside the model's top-200 predictions, and mean predictive entropy. The four values are standardized with training-set statistics that are frozen into the checkpoint, with means $(7.83, -4.47, .978, 1.654)$ and standard deviations $(2.60, 2.21, .011, .270)$ in the order above, and are encoded as feature tokens (Section~\ref{sec:unified-detector}).

\subsection{Architecture and Optimization}
\label{app:detector-hparams}

Table~\ref{tab:detector-hparams} lists the architecture, and Table~\ref{tab:detector-training} the two training stages. The detector has $300{,}295$ trainable parameters, at least two orders of magnitude fewer than the transformer detectors it is compared against. Both stages use AdamW with a constant learning rate and no warmup, and the optimizer state is reset when Stage B begins. The loss is the sum of a binary cross-entropy on AI versus non-AI and a conditional rewrite-versus-human cross-entropy evaluated on non-AI examples, with unit weight on the conditional term, no class reweighting, and no distillation or parameter-anchoring terms. Model selection maximizes the unweighted mean of the three per-domain development Macro-F1 scores; this selected epoch 10 of Stage A as \peerrevealbase and epoch 16 of Stage B as \peerreveal. Predictions are taken by argmax of the three-class distribution in Equation~\ref{eq:classification-head}, with no threshold tuning and no ensembling.

\begin{table}[H]
\caption{\peerreveal detector architecture. Attention and MLP dropout are used in both training stages; textual-token and reference-block dropout are enabled only in Stage B.}
\label{tab:detector-hparams}
\centering\small
\begin{tabular}{lp{7.2cm}}
\toprule
Component & Setting\\
\midrule
Claim encoder & Qwen3-Embedding-0.6B, $1{,}024$-d, frozen\\
Model width & 64\\
Encoder layers & 4\\
Attention heads & 4\\
Feed-forward width & 256\\
Classification MLP width & 128\\
Attention mask & directed blockwise (Section~\ref{sec:unified-detector})\\
Attention dropout & $.1$\\
MLP dropout & $.2$\\
Textual-token dropout & $0$ (Stage A), $.3$ (Stage B)\\
Reference-block dropout & $0$ (Stage A), $.1$ (Stage B)\\
Max target claims & 128\\
Max reference claims & 128\\
Textual features & 4\\
Trainable parameters & $300{,}295$\\
\bottomrule
\end{tabular}
\end{table}

\begin{table}[H]
\caption{Detector training. Stage A produces \peerrevealbase; Stage B continues from the selected Stage A checkpoint and produces \peerreveal.}
\label{tab:detector-training}
\centering\small
\begin{tabular}{lll}
\toprule
 & Stage A (regular) & Stage B (adversarial)\\
\midrule
Initialization & random & Stage A, epoch 10\\
Optimizer & AdamW, weight decay $.01$ & AdamW, weight decay $.05$\\
Learning rate & $8\times10^{-5}$, constant & $1\times10^{-4}$, constant\\
Batch size & 64 & 64\\
Steps per epoch & 406 & 406\\
Epoch budget & 30 & 20\\
Domain sampling (ML:Nature:PeerJ) & $1{:}1{:}1$ & $2{:}1{:}1$\\
Class sampling (AI:rewrite:human) & $1{:}1{:}1$ & $1{:}1{:}1$\\
Adversarial fraction & 0 & $.75$ AI, $.5$ rewrite\\
Gradient clipping & $1.0$ & $1.0$\\
Selected epoch & 10 & 16\\
\bottomrule
\end{tabular}
\end{table}

\section{Qualitative Examples}
\label{app:examples}

This appendix shows one paper from the PeerShift ML evaluation set together with all three of its review classes, so that the same submission, the same subject matter, and the same recommendation are held fixed while only authorship varies. The paper is \emph{Continual Normalization: Rethinking Batch Normalization for Online Continual Learning}~\citep{DBLP:conf/iclr/PhamLH22} (ICLR 2022). All three reviews carry a Borderline recommendation. The AI-rewritten review is a reconstruction of the human review shown here. Each review is reproduced in full, exactly as the detectors received it.

Table~\ref{tab:example-scores} gives every detector's output on these three reviews. The six three-class detectors are decided by the argmax of their class posterior; the eight binary detectors are decided by the fixed thresholds of Table~\ref{tab:macro-f1-thresholds}, under which a review is flagged when its score reaches the threshold, and AI-rewritten reviews count as negatives together with human-written ones. Of the 39 decisions made by the thirteen detectors other than \peerreveal, twenty are wrong.

\begin{table}[H]
\caption{Detector outputs on the three reviews of \emph{Continual Normalization} (ICLR 2022). Three-class detectors report $P(\mathrm{AI})/P(\mathrm{rewritten})/P(\mathrm{human})$ and decide by argmax; binary detectors report their score and flag a review when it reaches the threshold, treating rewritten and human reviews alike as negatives. Values are rounded to three decimals; $^{\ast}$ marks a wrong decision.}
\label{tab:example-scores}
\centering\small
\setlength{\tabcolsep}{4pt}
\begin{tabular}{lrccc}
\toprule
& & Human & AI-rewritten & AI-generated\\
Detector & Threshold & 319 words & 254 words & 419 words\\
& & & Kimi K2.5 & Grok 4.3\\
\midrule
\multicolumn{5}{l}{\emph{Three-class detectors}}\\
\roberta (regular) & argmax & $.000/.000/1.000$ & $.000/1.000/.000$ & $.000/1.000/.000^{\ast}$\\
\roberta (adversarial) & argmax & $.000/.004/.996$ & $.000/.999/.001$ & $.000/.999/.001^{\ast}$\\
\semdetect (regular) & argmax & $.321/.351/.328^{\ast}$ & $.313/.319/.368^{\ast}$ & $.315/.319/.367^{\ast}$\\
\semdetect (adversarial) & argmax & $.038/.391/.571$ & $.033/.110/.857^{\ast}$ & $.062/.216/.723^{\ast}$\\
\peerrevealbase & argmax & $.000/.004/.996$ & $.000/.841/.158$ & $.000/.006/.994^{\ast}$\\
\peerreveal & argmax & $.000/.004/.996$ & $.001/.999/.000$ & $.989/.011/.000$\\
\midrule
\multicolumn{5}{l}{\emph{Binary detectors}}\\
\logrank & $-1.128$ & $-1.784$ & $-2.266$ & $-2.538^{\ast}$\\
\fastdetectgpt & $1.402$ & $0.721$ & $-1.779$ & $-2.686^{\ast}$\\
\binoculars & $-0.947$ & $-0.214^{\ast}$ & $-0.263^{\ast}$ & $-0.793$\\
\mage & $1.000$ & $0.948$ & $0.000$ & $0.098^{\ast}$\\
\radar & $0.634$ & $0.888^{\ast}$ & $0.884^{\ast}$ & $0.737$\\
\editlens & $0.429$ & $0.014$ & $0.475^{\ast}$ & $0.669$\\
\tf & $0.291$ & $0.033$ & $0.078$ & $0.025^{\ast}$\\
\anchor & $0.873$ & $0.895^{\ast}$ & $0.893^{\ast}$ & $0.798^{\ast}$\\
\bottomrule
\end{tabular}
\end{table}

\paragraph{Human-written review (319 words).}
\begin{quote}\small
This paper investigates the problem of cross-task normalization in a non-stationary online continual learning(CL) setting. It argues that batch normalization (BN) is important to CL; however, BN in the current form introduces a bias towards current task, leading to catastrophic forgetting. The paper presents a continual normalization(CN) layer to reduce this bias. In particular, it combines spatial and batch normalization into one layer that is suitable for CL. The authors conducted experiments to evaluate the performance of the proposed method.

Strong Points

The paper takes one of the most import issues in continual learning: non-stationary online CL setting. For me, the problem itself is real and practical.

The proposed approach is reasonable and addresses the problem of cross-task normalization in CL setting.

The proposed adaptive normalization layers can be a plug-in replacement of BN.

The authors made extensive comparison between CN and prior work such as BN, IN, and GN.

Overall, the paper is well written. In particular, the Related Work section has a nice flow and puts the proposed method into context. Despite the method having limited novelty, the method has been well motivated by pointing out the limitations in SOTA methods.

The authors provide code for reproducing the results in the paper.

Weak Points

The proposed adaptive normalization layer is a straight-forward combination of spatial and batch normalization. So the novelty is limited.

Table 2, in DER++ setting, CN's FM score (for catastrophic forgetting) is not as good as GN's. Please explain the reason for that.

Overall, I vote for marginally accepting. I like the idea of cross-task normalization and handling it by the proposed adaptive normalization layer. My major concern is about the limited novelty of the paper and the performance of CN on catastrophic forgetting (see weakness above). Hopefully the authors can address my concern in the rebuttal period.

[After rebuttal]\\
After reading the other reviewers' comments and the authors' rebuttal, I confirm my rating.
\end{quote}

\paragraph{AI-rewritten review (Kimi K2.5, 254 words).}
\begin{quote}\small
This paper investigates cross-task normalization in non-stationary online continual learning, arguing that batch normalization introduces a troubling bias toward the current task that ultimately drives catastrophic forgetting. The proposed continual normalization layer attempts to mitigate this by combining spatial and batch normalization, and I find the core idea genuinely appealing---the notion of an adaptive normalization mechanism tailored for continual learning makes practical sense.

The work delivers several strengths worth acknowledging. It addresses a real problem that practitioners face, and CN's plug-in compatibility with existing architectures is a genuine convenience. The comparisons with batch, instance, and group normalization are extensive, the writing---particularly the related work section---is clear, and the provided code supports reproducibility. The motivation underlying the method is sound even if the technical contribution itself feels somewhat familiar.

That said, the novelty remains a sticking point. CN essentially stitches together established normalization techniques in a straightforward manner, which limits its conceptual contribution. More troubling is the empirical result in Table 2 under the DER++ setting, where CN's forgetting measure actually underperforms group normalization. The paper owes readers an explanation for this pattern, as it cuts against the central claim that CN specifically addresses forgetting.

I land at a marginal accept. The cross-task normalization perspective and the proposed adaptive layer are ideas I genuinely like and want to see in the literature. However, the limited novelty combined with the concerning forgetting results temper my enthusiasm---the paper earns its place but not a stronger endorsement. After reviewing the rebuttal and other comments, I maintain this rating.
\end{quote}

\paragraph{AI-generated review (Grok 4.3, 419 words).}
\begin{quote}\small
The paper begins from a concrete and well-motivated observation: on the five-task pMNIST stream, replacing BN's running statistics with oracle moments computed from all training data (denoted BN*) raises final accuracy while leaving learning accuracy essentially unchanged and lowering the forgetting measure (Table 1). The gap is largest without a replay buffer and shrinks once ER supplies a modest amount of past-task data, consistent with the claim that recent-task moments bias normalization of older examples. This toy result directly supports the existence of a cross-task normalization effect under strictly online, single-epoch training.

That observation, however, does not automatically establish the same mechanism as the primary driver of performance differences once the authors move to Split CIFAR-100 and Mini-ImageNet. In Table 2, CN improves both LA and ACC relative to BN under ER and DER++, yet the reduction in FM is frequently comparable to (or smaller than) the reduction already obtained by plain GN. Because GN normalizes entirely within-sample and therefore cannot perform any cross-task moment sharing, the similar or better forgetting numbers under GN already achieve one of the desiderata listed in Section 4.3. The additional gain from CN could therefore be explained either by a successful balance between batch-wise transfer and per-sample adaptivity, or simply by the fact that a two-stage GN-then-BN cascade produces more stable gradients in the low-batch, single-epoch regime irrespective of task ordering. The experiments do not contain a direct measurement---such as the per-task divergence between test-time feature moments under CN versus BN---that would separate these accounts.

The same ambiguity appears in the class-incremental results (Table 4). CN(G=32) yields the highest average accuracy, but the standard deviations remain large at the smallest buffer size and the ranking between CN(G=8) and CN(G=32) is not uniform across memory regimes. Without an ablation that freezes the running statistics or measures how much each method's normalization statistics drift across task boundaries, it remains unclear whether the reported margin is attributable to reduced cross-task bias or to generic optimization advantages of the hybrid normalizer. The long-tailed experiments (Tables 5--6) exhibit still smaller and occasionally reversed gains on minority classes, further indicating that any benefit is sensitive to distribution skew and not uniformly explained by the proposed mechanism.

Consequently, while the initial pMNIST diagnostic is illuminating and CN is a lightweight, drop-in replacement that improves accuracy on the tested benchmarks, the evidence tying those improvements specifically to mitigation of cross-task normalization is indirect and admits plausible alternative interpretations. Stronger causal measurements would be required to move the contribution beyond the borderline.
\end{quote}

\paragraph{What the errors look like.}
The three reviews all discuss the same normalization results and arrive at the same recommendation. \peerrevealbase assigns the generated and human reviews similarly high human probabilities ($.994$ and $.996$): before adversarial training the detector reads a paper-grounded generated review as human with high confidence. \peerreveal moves that review to $P(\mathrm{AI})=.989$ while leaving the human review at $P(\mathrm{human})=.996$, which is the gain Section~\ref{sec:adversarial-training} trains for. The rewritten review separates the systems further still: \peerreveal places $.999$ on the rewritten class, whereas both \semdetect variants call it human, and the eight binary detectors have no way to express the distinction at all, since their only question is whether a review is AI-generated. Among those eight, the errors run in both directions on the same paper --- \anchor flags the human and rewritten reviews while missing the generated one, and \tf, \logrank, \fastdetectgpt, and \mage miss the generated review after passing the other two --- which is the pattern behind the low binary Macro-F1 values of Table~\ref{tab:binary-main}.

\section{Adversarial Generator Example}
\label{app:generator-example}

This appendix shows one review written by the ML generator after adversarial training (Section~\ref{sec:adversarial-training}). The generator was given \emph{Learning to Guide and to be Guided in the Architect-Builder Problem}~\citep{DBLP:conf/iclr/BardeKNMPO22} (ICLR 2022), a Reject recommendation, and the paper-only strategy condition. The quality judge of Appendix~\ref{app:quality-judge} scores the review 5 on all four dimensions, and RevUtil averages 5.0. \peerreveal never saw this review during training. The review is reproduced in full, exactly as the detectors received it.

Table~\ref{tab:generator-example-scores} gives every detector's output, decided as in Table~\ref{tab:example-scores}. Ten of the thirteen detectors other than \peerreveal miss the review.

\begin{table}[htbp]
\caption{Detector outputs on the generated review of the Architect-Builder paper (ICLR 2022). Three-class detectors report $P(\mathrm{AI})/P(\mathrm{rewritten})/P(\mathrm{human})$ and decide by argmax; binary detectors report their score and flag the review when it reaches the threshold of Table~\ref{tab:macro-f1-thresholds}. Values are rounded to three decimals; $^{\ast}$ marks a missed detection.}
\label{tab:generator-example-scores}
\centering\small
\setlength{\tabcolsep}{4pt}
\begin{tabular}{lrc}
\toprule
& & AI-generated\\
Detector & Threshold & 317 words\\
& & Qwen3.5-9B\\
\midrule
\multicolumn{3}{l}{\emph{Three-class detectors}}\\
\roberta (regular) & argmax & $1.000/.000/.000$\\
\roberta (adversarial) & argmax & $.999/.000/.000$\\
\semdetect (regular) & argmax & $.313/.319/.368^{\ast}$\\
\semdetect (adversarial) & argmax & $.156/.006/.839^{\ast}$\\
\peerrevealbase & argmax & $.001/.003/.997^{\ast}$\\
\peerreveal & argmax & $.999/.001/.000$\\
\midrule
\multicolumn{3}{l}{\emph{Binary detectors}}\\
\logrank & $-1.128$ & $-2.646^{\ast}$\\
\fastdetectgpt & $1.402$ & $-2.503^{\ast}$\\
\binoculars & $-0.947$ & $-1.069^{\ast}$\\
\mage & $1.000$ & $0.038^{\ast}$\\
\radar & $0.634$ & $0.992$\\
\editlens & $0.429$ & $0.368^{\ast}$\\
\tf & $0.291$ & $0.011^{\ast}$\\
\anchor & $0.873$ & $0.808^{\ast}$\\
\bottomrule
\end{tabular}
\end{table}

\paragraph{Generated review (Qwen3.5-9B, 317 words).}
\begin{quote}\small
The paper proposes the Architect--Builder Problem (ABP) and ABIG, an algorithm that learns guiding communication without rewards or demonstrations by alternating between modelling and guiding frames, using shared intent and self-imitation. On a tiny gridworld with up to six blocks, ABIG achieves decent task success and shows some transfer when trained on ``place''. The idea is intriguing and the analysis of forget/reinforce dynamics is thoughtful. However, the empirical story feels weak: environment is trivial, baselines are sparse (random actions; an oracle ABIG-no-intent that cheats at test), no ablation of priors beyond one switch, no comparison to standard MARL/teacher-student methods, and scalability/generalization is barely explored. Writing is dense and sometimes opaque; several claims rely on ``interesting'' analogies to naming games rather than solid evidence. Overall, interesting setting but not enough substance for acceptance now.

Strengths

\nopagebreak[4]
Clear formalisation of ABP; nice intuition about stationarity via frames; good analytic insight into how self-imitation reinforces message--action associations and forgets unused ones; clean experiments showing transfer from simple to harder tasks when trained on ``place''; clean plots and stats.

Weaknesses

Very narrow evaluation: tiny gridworld, few tasks, no vision/perception, no locomotion beyond navigation/grasp-drop. Baselines are weak and somewhat unfair; no strong MARL/IL/HRI comparators. Ablations lack depth (no testing of frame length, network size, vocab size effects beyond one sweep). Claims about general communication emergence are overstated given the toy domain. Writing is occasionally unclear; some derivations feel hand-waved. No discussion of failure modes or sample efficiency. The ``transfer'' gains may simply reflect task decomposition rather than genuine protocol emergence.

Questions for Authors

Why no MARL/teacher-student baselines? Can ABIG match them when rewards\slash demonstrations are available? What happens with larger grids, more agents, or continuous actions\slash perception? How sensitive are results to frame design\slash network\slash architecture? Does transfer generalize to truly new semantics or just similar geometries? Are claims about naming-game parallels supported empirically? Please clarify oracle baseline at test and justify its fairness.

Final Judgement

Reject
\end{quote}

\paragraph{What the detectors see.}
\peerrevealbase reads the review as human ($P(\mathrm{human})=.997$), and so do both \semdetect variants; \peerreveal, trained on the adversarial reviews of Section~\ref{sec:adversarial-training}, assigns $P(\mathrm{AI})=.999$. \anchor compares the review with a same-paper GPT-5.5 reference and misses it ($0.808$ against $0.873$), and \tf and the other binary detectors except \radar all score it on the human side of their thresholds. Both \roberta variants also detect the review. \radar flags it, but also flags both non-AI reviews in Appendix~\ref{app:examples} ($0.888$ and $0.884$ against a threshold of $0.634$).

\end{document}